\RequirePackage{fix-cm}
\documentclass[twocolumn]{svjour3}
\smartqed  

\pdfoutput=1

\usepackage{times}
\usepackage{microtype} 

\usepackage[numbers]{natbib}
\usepackage{multicol}
\usepackage[bookmarks=true]{hyperref}

\usepackage{amsmath,amssymb,amsfonts}
\usepackage{tikz, tikz-3dplot}
\usepackage{booktabs}
\usepackage{subfig}
\usepackage[binary-units=true]{siunitx}
\usepackage{multirow}
\usepackage{rotating} 
\usepackage{tabularx}
\usepackage{graphbox}
\usepackage[outline]{contour}
\usepackage{bm} 
\usepackage[nolist,nohyperlinks]{acronym}
\usepackage{pifont}
\usepackage{xcolor,colortbl}                                                                                                                                                                                   
\usepackage{hhline} 

\DeclareMathAlphabet{\mathsfit}{T1}{\sfdefault}{\mddefault}{\sldefault}
\SetMathAlphabet{\mathsfit}{bold}{T1}{\sfdefault}{\bfdefault}{\sldefault}

\newcommand{\PreserveBackslash}[1]{\let\temp=\\#1\let\\=\temp}
\newcolumntype{C}[1]{>{\PreserveBackslash\centering}p{#1}}
\newcolumntype{R}[1]{>{\PreserveBackslash\raggedleft}p{#1}}
\newcolumntype{L}[1]{>{\PreserveBackslash\raggedright}p{#1}}
\newcommand\ph{$\phantom{1}$} 

\newacro{TSDF}{truncated signed distance function}
\newacro{ICP}{iterative closest points}
\newacro{SLAM}{simultaneous localization and mapping}
\newacro{RBF}{radial basis functions}
\newacro{MLP}{multi-layer perceptron}
\newacro{CRF}{conditional random field}

\usetikzlibrary{calc,arrows,arrows.meta,backgrounds,tikzmark,shapes.geometric, decorations.pathreplacing,decorations.markings, spy}

\definecolor{ph-purple}{RGB}{129, 39, 232}
\definecolor{ph-blue}{RGB}{5, 131, 227}
\definecolor{ph-gray}{rgb}{0.5, 0.5, 0.5}
\definecolor{ph-orange}{RGB}{227, 127, 5}
\definecolor{ph-green}{RGB}{0, 135, 124}
\definecolor{ph-yellow}{RGB}{235, 201, 52}
\definecolor{ph-light-green}{RGB}{181, 209, 21}
\definecolor{ph-red}{RGB}{250, 101, 60}
\definecolor{Gainsboro}{RGB}{220, 220, 220}
\colorlet{ph-orange-light}{ph-orange!70}
\colorlet{ph-blue-light}{ph-blue!70}
\colorlet{ph-purple-light}{ph-purple!70}
\colorlet{ph-green-light}{ph-green!70}
\definecolor{ph-light-gray}{rgb}{0.75, 0.75, 0.75}

\newcommand{\norm}[1]{\left\lVert#1\right\rVert}
\newcommand{\abs}[1]{\lvert#1\rvert}

\newcommand{\reffig}[1]{Fig.~\ref{#1}}
\newcommand{\reftab}[1]{Tab.~\ref{#1}}
\newcommand{\refsec}[1]{Sec.~\ref{#1}}

\newcommand{\etal}{et al.~}
\newcommand{\wrt}{w.r.t.~}

\newcommand{\vast}{\bBigg@{4}}
\newcommand{\Vast}{\bBigg@{5}}

\tikzset{
	on each segment/.style={
		decorate,
		decoration={
			show path construction,
			moveto code={},
			lineto code={
				\path [#1]
				(\tikzinputsegmentfirst) -- (\tikzinputsegmentlast);
			},
			curveto code={
				\path [#1] (\tikzinputsegmentfirst)
				.. controls
				(\tikzinputsegmentsupporta) and (\tikzinputsegmentsupportb)
				..
				(\tikzinputsegmentlast);
			},
			closepath code={
				\path [#1]
				(\tikzinputsegmentfirst) -- (\tikzinputsegmentlast);
			},
		},
	},
	mid arrow/.style={postaction={decorate,decoration={
				markings,
				mark=at position .5 with {\arrow[#1]{stealth}}
	}}},
	pos arrow/.style 2 args={postaction={decorate,decoration={
				markings,
				mark=at position [#2] with {\arrow[#1]{stealth}}
	}}},
	custom arrow/.style={postaction={decorate,decoration={
			markings,
			mark=at position .528 with {\arrow[#1]{stealth}}
	}}},
	custom arrow2/.style={postaction={decorate,decoration={
			markings,
			mark=at position .475 with {\arrow[#1]{stealth}}
	}}},
}

\begin{document}\sloppy

\title{LatticeNet: Fast Spatio-Temporal Point Cloud Segmentation Using Permutohedral Lattices\\
}

\author{Radu Alexandru Rosu \and Peer Sch{\"u}tt \and Jan Quenzel \and  Sven Behnke
\thanks{This work has been funded by the Deutsche Forschungsgemeinschaft (DFG, German Research Foundation) under Germany's Excellence Strategy - EXC 2070 - 390732324 and by the German Federal Ministry of Education and Research (BMBF) in the project ”Kompetenzzentrum: Aufbau des Deutschen Rettungsrobotik-Zentrums” (A-DRZ).}%
} 


\institute{Radu Alexandru Rosu \at
	Friedrich-Hirzebruch-Allee 8 \\ 
	Tel.: +49 (0) 228 73-54159 \\
	Fax: +49 (0) 228 73-4425 \\
	\email{rosu@ais.uni-bonn.de}           
}

\date{}

\maketitle
\global\csname @topnum\endcsname 0
\global\csname @botnum\endcsname 0

\begin{abstract}
Deep convolutional neural networks (CNNs) have shown outstanding performance in the task of semantically segmenting images. Applying the same methods on 3D data still poses challenges due to the heavy memory requirements and the lack of structured data. Here, we propose LatticeNet, a novel approach for 3D semantic segmentation, which takes raw point clouds as input. A PointNet describes the local geometry which we embed into a sparse permutohedral lattice. The lattice allows for fast convolutions while keeping a low memory footprint. Further, we introduce DeformSlice, a novel learned data-dependent interpolation for projecting lattice features back onto the point cloud. We present results of 3D segmentation on multiple datasets where our method achieves state-of-the-art performance. We also extend and evaluate our network for instance and dynamic object segmentation.
\end{abstract}


\section{Introduction}

Environment understanding is a crucial ability for autonomous agents. Perceiving not only the geometrical structure of the scene but also distinguishing between different classes of objects therein enables tasks like manipulation and interaction that were previously not possible. Within this field, semantic segmentation of 2D images is a mature research area, showing outstanding success in dense per pixel categorization on images~\cite{long2015fully,chen2017rethinking,lin2017refinenet}. However, the task of semantically labelling 3D data is still an open area of research as it poses several challenges that need to be addressed.

First, 3D data is often represented in an unstructured manner --- unlike the grid-like structure of images. This raises difficulties for current approaches which assume a regular structure upon which convolutions are defined.

\begin{figure}[t]

	\centering
	\tdplotsetmaincoords{-35.3}{0}
	\tdplotsetrotatedcoords{0}{-45}{0}
	\begin{tikzpicture}[tdplot_main_coords, remember picture, >={Stealth[inset=1pt,length=8pt,angle'=30,round]} ]

	\begin{scope}
	\def\solidImg{./imgs/teaser/moto_9_solid.png} 
	\newlength{\WS}
	\newlength{\HS}
	\settowidth{\WS}{\includegraphics{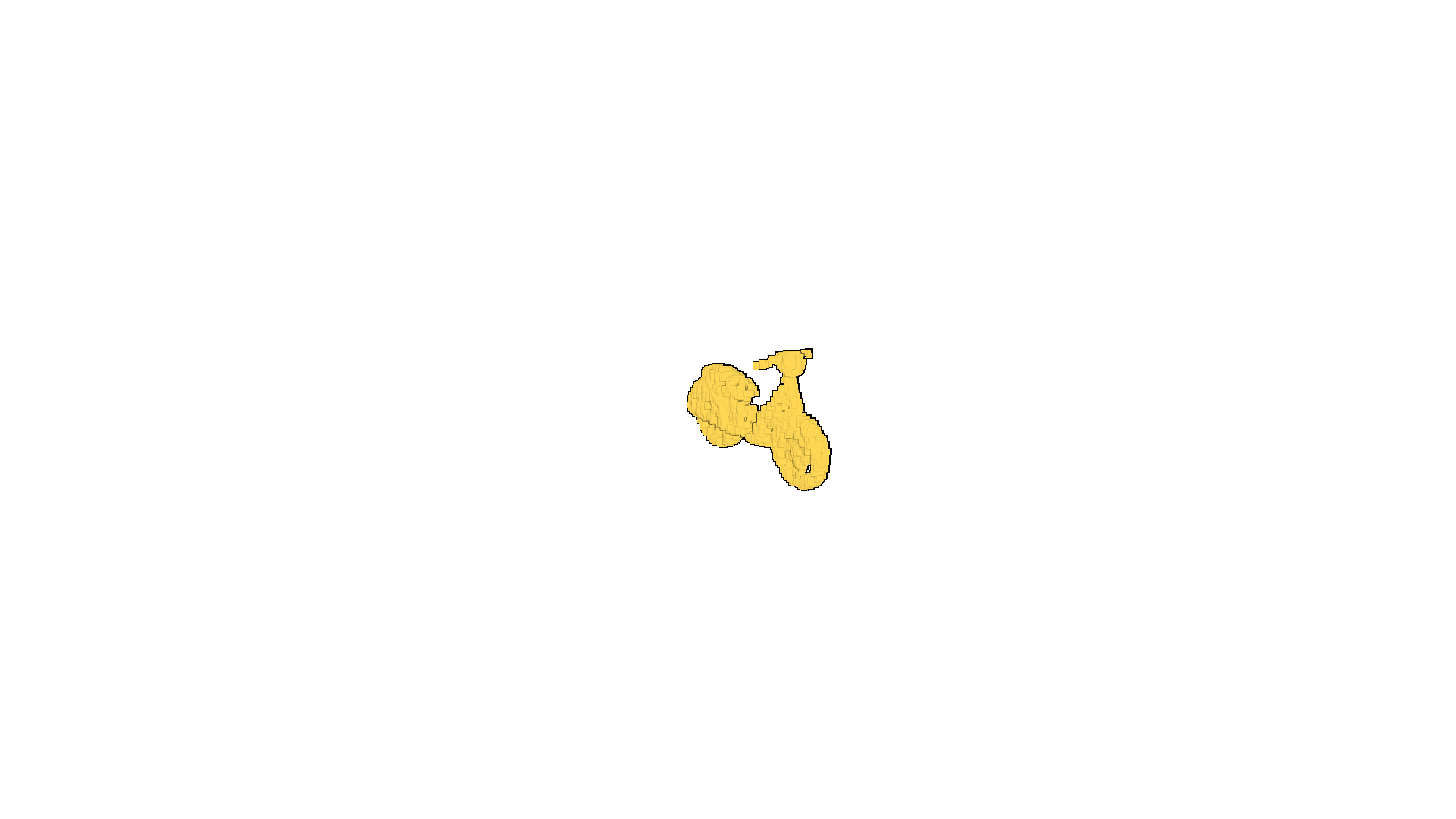}}
	\settoheight{\HS}{\includegraphics{\solidImg}}
	\node[inner sep=0pt] (russell) at (-3,0) {\includegraphics[trim=.46\WS{} .4\HS{} .425\WS{} .41\HS{},clip, width=.15\textwidth]{\solidImg}};
	\end{scope}

	\node[] (i10) at (-2.3,0.5){};
	\node[] (i11) at (-1.9,0.6){};
	\node[] (i12) at (-1.1,1.2){};
	\node[] (i13) at (-0.2,0.4){};
	\fill[ph-blue!70] (i13) circle (1.5pt);
	\begin{scope}[decoration={
		markings,
		mark=at position 0.5 with {\arrow[rotate=20]{>}}}
	] 
	\draw [gray, dashed, postaction={decorate}] plot [smooth, tension=0.8] coordinates { (i10) (i11) (i12) (i13) };
	\end{scope}

	\node[] (i20) at (-2.0,-0.5){};
	\node[] (i21) at (-1.5,-0.55){};
	\node[] (i22) at (-0.9,-0.3){};
	\node[] (i23) at (-0.45,-0.0){};
	\fill[ph-blue!70] (i23) circle (1.5pt);
	\begin{scope}[decoration={
		markings,
		mark=at position 0.4 with {\arrow{>}}}
	] 
	\draw [gray, dashed, postaction={decorate}] plot [smooth, tension=0.6] coordinates { (i20) (i21) (i22) (i23) };
	\end{scope}
	
	\node[] (o10) at (2.3,0.5){};
	\node[] (o11) at (1.9,0.6){};
	\node[] (o12) at (1.1,1.2){};
	\node[] (o13) at (0.2,0.4){};
	\fill[ph-blue!70] (o13) circle (1.5pt);
	\begin{scope}[decoration={
		markings,
		mark=at position 0.5 with {\arrow[rotate=-15]{<}}}
	] 
	\draw [gray, dashed, postaction={decorate}] plot [smooth, tension=0.6] coordinates { (o10) (o11) (o12) (o13) };
	\end{scope}
	
	\node[] (o20) at (2.0,-0.5){};
	\node[] (o21) at (1.5,-0.55){};
	\node[] (o22) at (0.9,-0.3){};
	\node[] (o23) at (0.45,-0.0){};
	\fill[ph-blue!70] (o23) circle (1.5pt);
	\begin{scope}[decoration={
		markings,
		mark=at position 0.35 with {\arrow{<}}}
	] 
	\draw [gray, dashed, postaction={decorate}] plot [smooth, tension=0.6] coordinates { (o20) (o21) (o22) (o23) };
	\end{scope}

	\begin{scope}[tdplot_rotated_coords]
	\newcommand{\LatticeHops}{2}
	\newcommand{\LatticeScale}{0.8}
	\foreach \x in {0,...,\LatticeHops}{%
		\foreach \y in {0,...,\LatticeHops}{%
			\foreach \z in {0,...,\LatticeHops}{%
				\coordinate [tdplot_rotated_coords] (\x;\y;\z) at (\x*\LatticeScale,\y*\LatticeScale,\z*\LatticeScale);
				\fill[gray] (\x;\y;\z) circle (1.5pt);
	} }}
	
	\pgfmathtruncatemacro\NrLines{\LatticeHops-1}
	\foreach \x in {0,...,\NrLines}{%
		\foreach \y in {0,...,\NrLines}{%
			\foreach \z in {0,...,\NrLines}{%
				\pgfmathtruncatemacro\nx{\x+1}
				\pgfmathtruncatemacro\ny{\y+1}
				\pgfmathtruncatemacro\nz{\z+1}
				\draw[thin,gray] (\x;\y;\z)--(\nx;\y;\z);
				\draw[thin,gray] (\x;\y;\z)--(\x;\ny;\z);
				\draw[thin,gray] (\x;\y;\z)--(\x;\y;\nz);
				
				\draw[thin,gray] (\nx;\ny;\nz)--(\x;\ny;\nz);
				\draw[thin,gray] (\nx;\ny;\nz)--(\nx;\y;\nz);
				\draw[thin,gray] (\nx;\ny;\nz)--(\nx;\ny;\z);
				
				\draw[thin,gray] (\x;\ny;\z)--(\nx;\ny;\z);
				\draw[thin,gray] (\nx;\y;\z)--(\nx;\ny;\z);
				\draw[thin,gray] (\x;\ny;\z)--(\x;\ny;\nz);
				\draw[thin,gray] (\x;\y;\nz)--(\x;\ny;\nz);
				\draw[thin,gray] (\x;\y;\nz)--(\nx;\y;\nz);
				\draw[thin,gray] (\nx;\y;\z)--(\nx;\y;\nz);
				
	} } }
	%
	\renewcommand{\LatticeHops}{1} 
	\pgfmathtruncatemacro\NrLines{\LatticeHops-1}
	\foreach \x in {0,...,\NrLines}{%
		\foreach \y in {0,...,\NrLines}{%
			\foreach \z in {0,...,\NrLines}{%
				\pgfmathtruncatemacro\nx{\x+1}
				\pgfmathtruncatemacro\ny{\y+1}
				\pgfmathtruncatemacro\nz{\z+1}
				\draw[thick,ph-orange] (\x;\y;\z)--(\nx;\y;\z);
				\draw[thick,ph-orange] (\x;\y;\z)--(\x;\ny;\z);
				\draw[thick,ph-orange] (\x;\y;\z)--(\x;\y;\nz);
				
				\draw[thick,ph-orange] (\nx;\ny;\nz)--(\x;\ny;\nz);
				\draw[thick,ph-orange] (\nx;\ny;\nz)--(\nx;\y;\nz);
				\draw[thick,ph-orange] (\nx;\ny;\nz)--(\nx;\ny;\z);
				
				\draw[thick,ph-orange] (\x;\ny;\z)--(\nx;\ny;\z);
				\draw[thick,ph-orange] (\nx;\y;\z)--(\nx;\ny;\z);
				\draw[thick,ph-orange] (\x;\ny;\z)--(\x;\ny;\nz);
				\draw[thick,ph-orange] (\x;\y;\nz)--(\x;\ny;\nz);
				\draw[thick,ph-orange] (\x;\y;\nz)--(\nx;\y;\nz);
				\draw[thick,ph-orange] (\nx;\y;\z)--(\nx;\y;\nz);
				
	} } }
	\foreach \x in {0,...,\LatticeHops}{%
		\foreach \y in {0,...,\LatticeHops}{%
			\foreach \z in {0,...,\LatticeHops}{%
				\coordinate (\x;\y;\z) at (\x*\LatticeScale,\y*\LatticeScale,\z*\LatticeScale);
				\fill[ph-orange] (\x;\y;\z) circle (2.5pt);
	} }}

	\end{scope}

	\begin{scope}
	\def\segImg{./imgs/teaser/moto_9_seg.png} 
	\newlength{\WG}
	\newlength{\HG}
	\settowidth{\WG}{\includegraphics{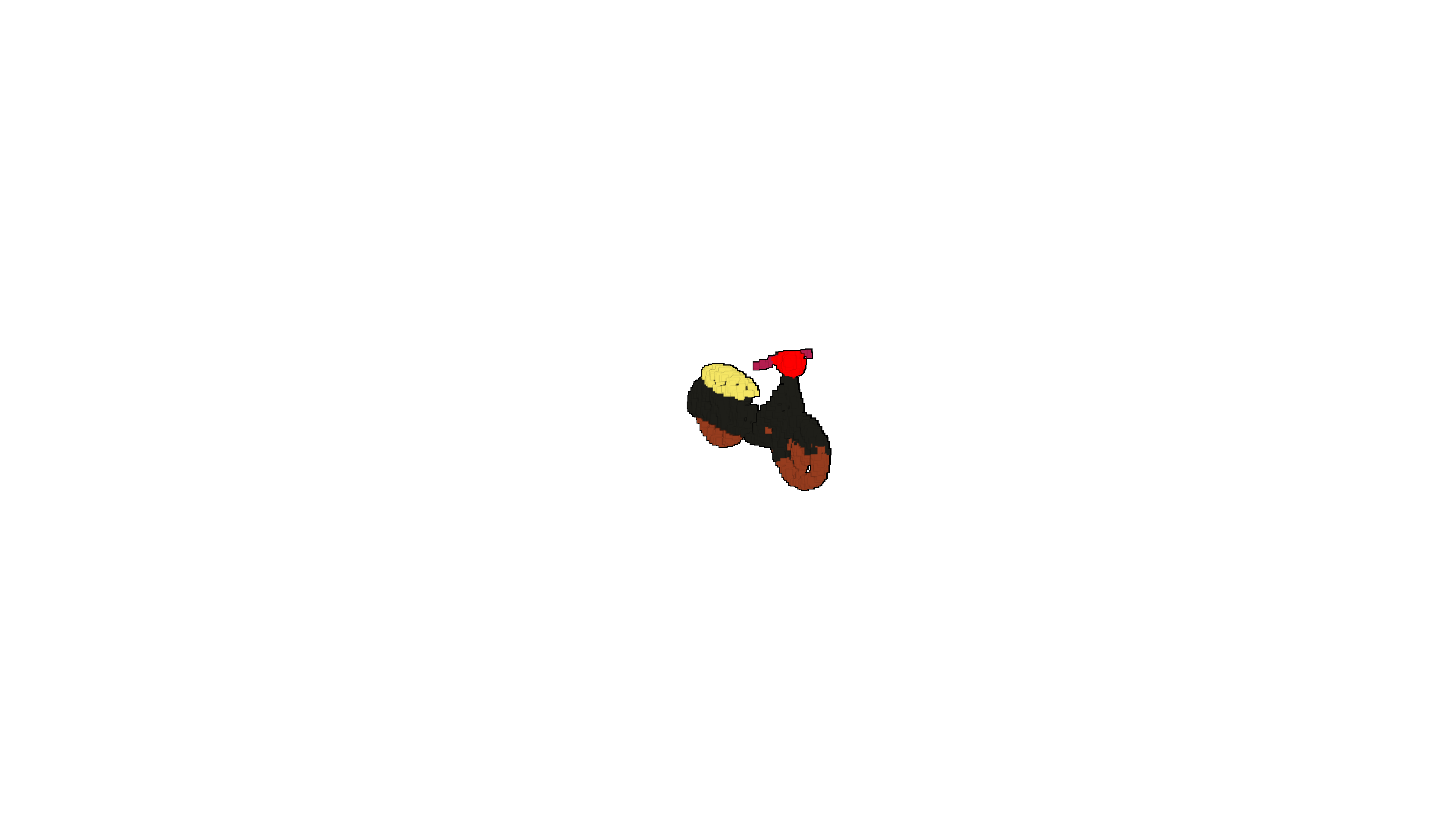}}
	\settoheight{\HG}{\includegraphics{\segImg}}
	\node[inner sep=0pt] (russell) at (2.9,0) {\includegraphics[trim=.46\WG{} .4\HG{} .425\WG{} .41\HG{},clip, width=.15\textwidth]{\segImg}};
	\end{scope}
	
	\end{tikzpicture}
	
	\caption{Semantic segmentation: LatticeNet takes raw point clouds as input and embeds them into a sparse lattice where convolutions are applied. Features on the lattice are projected back onto the point cloud to yield a final segmentation.}
	\label{fig:teaser}
\end{figure}

Second, the performance of current 3D networks is limited by their memory requirements. Storing 3D information in a dense structure is prohibitive for even high-end GPUs, clearly indicating the need for a sparse structure.

Third, discretization issues caused by imposing a regular grid onto point clouds can negatively affect the network's performance and interpolation is necessary to cope with quantization artifacts~\cite{tchapmi2017segcloud}.

In this work, we propose LatticeNet, a novel approach for point cloud segmentation which alleviates the previously mentioned problems.
Hence, our contributions are:
\begin{itemize}
	\item a hybrid architecture which leverages the strength of PointNet to obtain low-level features and sparse 3D convolutions to aggregate global context,
	\item a framework suitable for sparse data onto which all common CNN operators are defined, and 
	\item a novel slicing operator that is end-to-end trainable for mapping features of a regular lattice grid back onto an unstructured point cloud.
\end{itemize}

In addition to our \textit{Robotics: Science and System} conference paper~\cite{rosu2019latticenet} we make the following additional contributions:
\begin{itemize}
	\item an extension with discriminative loss that allows LatticeNet to perform instance segmentation, and
	\item a network architecture capable of processing temporal information in order to improve semantic segmentation and to distinguish between dynamic and static objects within the scene.
\end{itemize}

\section{Related Work}

\subsection{Semantic Segmentation}

3D Semantic segmentation approaches can be categorized depending on data representation upon which they operate.

\noindent\textbf{Point cloud networks:}
The first category of networks operates directly on the raw point cloud. 

From this area, PointNet~\cite{qi2017pointnet} is one of the pioneering works. The method processes raw point clouds by individually embedding the points into a higher-dimensional space and applying max-pooling for permutation-invariance to obtain a global scene descriptor. The descriptor can be used for both classification and semantic segmentation.
However, PointNet does not take local information into account which is essential for the segmentation of highly-detailed objects. This has been partially solved in the subsequent work of PointNet++~\cite{qi2017pointnet++} which applies PointNet hierarchically, capturing both local and global contextual information.

Chen~\etal\cite{chen2018deep} use a similar approach but they input the point responses \wrt a sparse set of \ac{RBF} scattered in 3D space. Optimizing jointly for the extent and center of the \ac{RBF} kernels allows to obtain a more explicit modelling of the spatial distribution.

PointCNN~\cite{li2018pointcnn} deals with the permutation invariance not by using a symmetric aggregation function, but by learning a $K\times K$ matrix for the $K$ input points that permutes the cloud into a canonical form.

\noindent\textbf{Voxel networks:}
3D Convolutions in this category work on discretized cubic or tetrahedral volume elements.

SEGCloud~\cite{tchapmi2017segcloud} voxelizes the point cloud into a uniform 3D grid and applies 3D convolutions to obtain per-voxel class probabilities. A \ac{CRF} is used to smooth the labels and enforce global consistency. The class scores are transferred back to the points using trilinear interpolation. The usage of a dense grid results in high memory consumption while our approach uses a permutohedral lattice stored sparsely. Additionally, their voxelization results in a loss of information due to the discretization of the space. We avoid quantization issues by using a PointNet architecture to summarize the local neighborhood.

Rethage~\etal\cite{rethage2018fully} perform semantic segmentation on a voxelized point cloud and employ a PointNet architecture as a low-level feature extractor. The usage of a dense grid, however, leads to high memory usage and slow inference, requiring various seconds for medium-sized point clouds.

SplatNet~\cite{su2018splatnet} is the work most closely related to ours. It alleviates the computational burden of 3D convolutions by using a sparse permutohedral lattice, performing convolutions only around the surfaces.
It discretizes the space in uniform simplices and accumulates the features of the raw point cloud onto the vertices of the lattice using a splatting operation. Convolutions are applied on the lattice vertices and a slicing operation barycentrically interpolates the features of the vertices back onto the point cloud. A series of splat-conv-slice operations are applied to obtain contextual information. 
The main disadvantage is that splat and slice operations are not learned and repeated application slowly degrades the point clouds features as they act as Gaussian filters~\cite{baek2009some}. Furthermore, storing high-dimensional features for each point in the cloud is memory intensive which limits the maximum number of points that can be processed. In contrast, our approach has learned operations for splatting and slicing which brings more representational power to the network. We also restrict their usage to only the beginning and the end of the network, leaving the rest of the architecture fully convolutional.

\noindent\textbf{Mesh networks:}
The connectivity of triangular or quadrilateral mesh faces enables easy computation of normal vectors and establishes local tangent planes.

GCNN~\cite{masci2015geodesic} operates on small local patches which are convolved using a series of rotated filters, followed by max-pooling to deal with the ambiguity in the patch orientation. However, the max-pooling disregards the orientation. MoNet~\cite{monti2017geometric} deals with the orientation ambiguity by aligning the kernels to the principal curvature of the surface. Yet, this does not solve cases in which the local curvature is not informative, e.g. for walls or ceilings. TextureNet~\cite{huang2019texturenet} further improves on the idea by using a global 4-RoSy orientations field. This provides a smooth orientation field at any point on the surface which is aligned to the edges of the mesh and has only a 4-direction ambiguity. Defining convolution on patches oriented according to the 4-RoSy field yields significantly improved results.

\noindent\textbf{Graph networks:} 
These methods allow arbitrary topologies to connect vertices and lift the restriction of triangular or quadrilateral meshes.

Wang~\etal\cite{wang2018deep} and Wu~\etal\cite{wu2019pointconv} define a convolution operator over non-grid structured data by having continuous values over the full vector space. The weights of these continuous filters are parametrized by an \ac{MLP}.

Defferrard~\etal\cite{defferrard2016convolutional} formulate CNNs in the context of spectral graph theory. They define the convolution in the Fourier domain with Chebyshev polynomials to obtain fast localized filters. However, spectral approaches are not directly transferable to a new graph as the Fourier basis changes. Additionally, the learned filters are rotation invariant which can be seen as a limitation to the representational power of the network.

\noindent\textbf{Multi-view networks:}
The convolution operation is well defined in 2D and hence, there is an interest in casting 3D segmentation as a series of single-view segmentations which are fused together.

Pham~\etal\cite{pham2019real} simultaneously reconstruct the scene geometry and recover the semantics by segmenting sequences of RGB-D frames. The segmentation is transferred from 2D images to the 3D world and fused with previous segmentations. A \ac{CRF} finally resolves noisy predictions.

TangentConv~\cite{tatarchenko2018tangent} assumes that the data is sampled from locally Euclidean surfaces and project the local surface geometry onto a tangent plane to which 2D convolutions can be applied. This requires a heavy preprocessing for normal calculation. In contrast, our approach can deal with raw point clouds without requiring normals.

\subsection{Motion Segmentation}

For the task of motion segmentation two approaches have been widely used: Networks either incorporate multiple point clouds directly or accumulate a sequence of individually segmented point clouds. 

Shi~\etal\cite{shi2020spsequencenet} present their U-Net based architecture SpSequenceNet for semantic segmentation on 4D point clouds. They input two point clouds and generate the output for the later one with a voxel-based method.
They designed two modules, the Cross-frame Global Attention~(CGA) and the Cross-frame Local Interpolation~(CLI) module. The CGA acts as a teacher that uses the data from $P_{t-1}$ to focus the network on the important features of $P_t$.
The CLI module fuses information between both point clouds by combining the spatial and temporal information.

Kernel Point Convolution (KPConv)~\cite{thomas2019kpconv} operates directly on the point clouds by facilitating convolution weights that are located in Euclidean space. Points in the vicinity of these kernels are weighted and summed together to feature vectors.
KPConv~\cite{thomas2019kpconv}, DarkNet53Seg~\cite{behley2019semantickitti} and TangentConv~\cite{tatarchenko2018tangent} were previously used for the segmentation of 4D point clouds by accumulating multiple clouds of a sequence.

\subsection{Instance Segmentation} 
Researchers extended principles from 2D to obtain instances in 3D which can 
be roughly categorized in proposal-based and proposal-free methods. 
	
	\noindent\textbf{Proposal-based:}
		This type solves the problem in two stages.		
The first network stage generates proposals of bounding boxes for the objects in the scene. A second stage performs foreground-background segmentation on the points within the bounding boxes in order to get valid instances.
		
		Yang~\etal\cite{yang2019learning} present a single-stage method for instance segmentation that can train both the proposal and the point-mask prediction network in an end-to-end manner.
		Yi~\etal\cite{yi2019gspn} alleviate some of the issues associated with wrong bounding box predictions by using an analysis-by-synthesis strategy.
		
	\noindent\textbf{Proposal-free:}
		Proposal-free methods tackle instance segmentation without the need of generating object proposals. They usually rely on predicting point embedding and apply clustering to recover the instances. 

		Many proposal-free approaches base their work on the 2D instance segmentation of Brabandere~\etal\cite{de2017semantic} in which pixel embeddings are predicted. There, a discriminative loss encourages the embeddings that belong to the same instance to be clustered together while embeddings from different instances should be further apart.
		
		SPGN~\cite{wang2018sgpn} learns a similarity matrix for all point pairs, based on which, similar points are merged to instances.
		VoteNet~\cite{qi2019deep} uses a Hough voting mechanism where the points predict the offset towards the object center. A clustering algorithm finally recovers the object instances.
	
		Neven~\etal\cite{neven2019instance} alleviate some of the issues associated with proposal-free methods by allowing also the clustering algorithm to be part of the training by jointly optimizing the spatial embeddings and the clustering bandwidth.
		
		Wang~\etal\cite{wang2019associatively} proposed a framework that allows for semantic and instances to be predicted simultaneously and for the two tasks to mutually benefit from each other. Similarly, Pham~\etal\cite{pham2019jsis3d} recover both instances and semantics and apply a \ac{CRF} to improve the predictions accuracy.
		
		Most of these works utilize a PointNet~\cite{qi2017pointnet} or PointNet++~\cite{qi2017pointnet++} network to predict the point embeddings. In our case, we extend LatticeNet in a similar manner to other proposal-free methods but predict the embeddings using the lattice convolutions.

\section{Notation}
Throughout this paper, we use bold upper-case characters to denote matrices and bold lower-case characters to denote vectors.

The vertices of the $d$-dimensional permutohedral lattice are defined as a tuple $v=\left( \mathbf{c}_v, \mathbf{x}_v \right)$, with $\mathbf{c}_v\in\mathbb{Z}^{ (d+1) }$ denoting the coordinates of the vertex and $\mathbf{x}_v \in \mathbb{R}^{ v_d }$ representing the values stored at vertex $v$. The full lattice containing $n$ vertices is denoted with $V=\left( \mathbf{C}, \mathbf{X} \right)$, with $\mathbf{C}\in\mathbb{Z}^{ n \times (d+1) }$ representing the coordinate matrix and $\mathbf{X}\in\mathbb{R}^{ n \times v_d }$ the value matrix.

The points in a cloud are defined as a tuple $p=\left( \mathbf{g}_p, \mathbf{f}_p \right)$, with $\mathbf{g}_p\in\mathbb{R}^{ d }$ denoting the coordinates of the point and $\mathbf{f}_p \in \mathbb{R}^{ f_d }$ representing the features stored at point $p$ (color, normals, etc.). The full point cloud containing $m$ points is denoted by $P=\left( \mathbf{G}, \mathbf{F} \right)$ with $\mathbf{G}\in\mathbb{R}^{ m \times d }$ being the positions matrix and $\mathbf{F}\in\mathbb{R}^{ m \times f_d }$ the feature matrix. The feature matrix $\mathbf{F}$ can also be empty in which case $f_d$ is set to zero. 

For motion segmentation we define a sequence of point clouds as $P_{seq} = \left(P_0, P_1, \ldots, P_n\right)$ with $P_n=\left( \mathbf{G}, \mathbf{F} \right)$. We define a timestep as processing one cloud of this sequence. 

We denote with $I_p$ the set of lattice vertices of the simplex that contains point $p$. The set $I_p$ always contains $d+1$ vertices as the lattice tessellates the space in uniform simplices with $d+1$ vertices each. Furthermore, we denote with $J_v$ the set of points $p$ for which 
vertex $v$ is one of the vertices of the containing simplices. Hence, these are the points that contribute to vertex $v$ through the splat operation.

We denote with $\mathcal{S}$ the splatting operation, with $\mathcal{Y}$ the slicing operation, with $\mathcal{\tilde{Y}}$ the deformable slicing, with $\mathcal{P}$ the PointNet module, with $\mathcal{D}_G$ and $\mathcal{D}_F$ the distribution of the point positions and the points features, respectively, and with $\mathcal{G}$ the gathering operation.

\section{Permutohedral Lattice}

The $d$-dimensional permutohedral lattice is formed by projecting the scaled regular grid $(d+1)\mathbb{Z}^{d+1}$ along the vector $\mathbf{1}=\left[1,\ldots,1\right]$ onto the hyperplane $H_d$: $\mathbf{p}\cdot\mathbf{1}=0$. 

The lattice tessellates the space into uniform $d$-dimensional simplices. Hence, for $d=2$ the space is tessellated with triangles and for $d=3$ into tetrahedra. The enclosing simplex of any point can be found by a simple rounding algorithm~\cite{baek2009some}.

Due to the scaling and projection of the regular grid, the coordinates $\mathbf{c}_v$ of each lattice vertex sum up to zero. Each vertex has $2(d+1)$ immediate neighboring vertices. 
The coordinates of these neighbors are separated by a vector of form $\pm \left[ -1,\ldots,-1,d,-1,\ldots,-1 \right]\in\mathbb{Z}^{d+1}$.

The vertices of the permutohedral lattice are stored in a sparse manner using a hash map in which the key is the coordinate $\mathbf{c}_v$ and the value is $\mathbf{x}_v$. Hence, we only allocate the simplices that contain the 3D surface of interest. This sparse allocation allows for efficient implementation of all typical operations in CNNs (convolution, pooling, transposed convolution, etc.).


The permutohedral lattice has several advantages \wrt standard cubic voxels. The number of vertices for each simplex is given by $d+1$ which scales linearly with increasing dimension, in contrast to the $2^d$ for standard voxels. This small number of vertices per simplex allows for fast splatting and slicing operations.
Furthermore, splatting and slicing create piece-wise linear outputs as they use barycentric interpolation. In contrast, standard quantization in cubic voxels create piece-wise constant outputs, leading to discretization artefacts.

Spatial correspondences between lattice vertices are given by design and the hashmap: If the hashmap stays the same for the whole sequence, spatially identical lattice vertices of different point clouds are always mapped to the same entries. This is visualized in \reffig{fig:temporal_fuse} where features from two different time-steps are fused together.

\section{Method}
The input to our method is a point cloud $P=\left( \mathbf{G}, \mathbf{F} \right)$ containing coordinates and per-point features.

We define the scale of the lattice by scaling the positions $\mathbf{G}$ as $\mathbf{G}_s=\mathbf{G}/\pmb{\sigma}$, where $\pmb{\sigma} \in\mathbb{R}^{d}$ is the scaling factor. The higher the sigma the less number of vertices will be needed to cover the point cloud and the coarser the lattice will be. For ease of notation, unless otherwise specified, we refer to $\mathbf{G}_s$ as $\mathbf{G}$ as we usually only need the scaled version.
 
\subsection{Common Operations on Permutohedral Lattice} \label{operations_original}
In this section, we will explain in detail the standard operations on a permutohedral lattice that are used in previous works~\cite{su2018splatnet,gu2019hplflownet}.

\noindent\textbf{Splatting} refers to the interpolation of point features onto the values of the lattice $V$ using barycentric weighting~(\reffig{fig:splat}). Each point splats onto $d+1$ lattice vertices and their weighted features are summed onto the vertices.

\noindent\textbf{Convolving} operates analogously to standard spatial convolutions in 2D or 3D, i.e. a weighted sum of the vertex values together with its neighbors is computed. We use convolutions that span over the 1-hop ring around a vertex and hence convolve the values of $2(d+1)+1$ vertices~(\reffig{fig:convolve}).

\bgroup
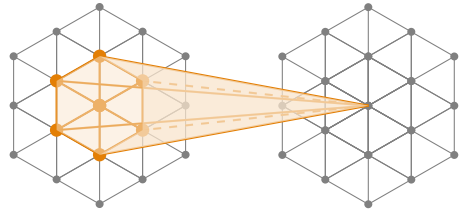
\begin{figure}[]
	\centering
	\tdplotsetmaincoords{-35.3}{0}
	\tdplotsetrotatedcoords{0}{-45}{0}
	\begin{tikzpicture}[tdplot_main_coords, remember picture, >={Stealth[inset=10pt,length=8pt,angle'=28,round]} ]
	\begin{scope}[tdplot_rotated_coords]
	\newcommand{\LatticeHops}{2}
	\newcommand{\LatticeScale}{0.8}
	\foreach \x in {0,...,\LatticeHops}{%
		\foreach \y in {0,...,\LatticeHops}{%
			\foreach \z in {0,...,\LatticeHops}{%
				\coordinate [tdplot_rotated_coords] (\x;\y;\z) at (\x*\LatticeScale,\y*\LatticeScale,\z*\LatticeScale);
				\fill[gray] (\x;\y;\z) circle (1.5pt);
	} }}
	
	\pgfmathtruncatemacro\NrLines{\LatticeHops-1}
	\foreach \x in {0,...,\NrLines}{%
		\foreach \y in {0,...,\NrLines}{%
			\foreach \z in {0,...,\NrLines}{%
				\pgfmathtruncatemacro\nx{\x+1}
				\pgfmathtruncatemacro\ny{\y+1}
				\pgfmathtruncatemacro\nz{\z+1}
				\draw[thin,gray] (\x;\y;\z)--(\nx;\y;\z);
				\draw[thin,gray] (\x;\y;\z)--(\x;\ny;\z);
				\draw[thin,gray] (\x;\y;\z)--(\x;\y;\nz);
				
				\draw[thin,gray] (\nx;\ny;\nz)--(\x;\ny;\nz);
				\draw[thin,gray] (\nx;\ny;\nz)--(\nx;\y;\nz);
				\draw[thin,gray] (\nx;\ny;\nz)--(\nx;\ny;\z);
				
				\draw[thin,gray] (\x;\ny;\z)--(\nx;\ny;\z);
				\draw[thin,gray] (\nx;\y;\z)--(\nx;\ny;\z);
				\draw[thin,gray] (\x;\ny;\z)--(\x;\ny;\nz);
				\draw[thin,gray] (\x;\y;\nz)--(\x;\ny;\nz);
				\draw[thin,gray] (\x;\y;\nz)--(\nx;\y;\nz);
				\draw[thin,gray] (\nx;\y;\z)--(\nx;\y;\nz);
				
	} } }
	%
	\renewcommand{\LatticeHops}{1} 
	\pgfmathtruncatemacro\NrLines{\LatticeHops-1}
	\foreach \x in {0,...,\NrLines}{%
		\foreach \y in {0,...,\NrLines}{%
			\foreach \z in {0,...,\NrLines}{%
				\pgfmathtruncatemacro\nx{\x+1}
				\pgfmathtruncatemacro\ny{\y+1}
				\pgfmathtruncatemacro\nz{\z+1}
				\draw[thick,ph-orange] (\x;\y;\z)--(\nx;\y;\z);
				\draw[thick,ph-orange] (\x;\y;\z)--(\x;\ny;\z);
				\draw[thick,ph-orange] (\x;\y;\z)--(\x;\y;\nz);
				
				\draw[thick,ph-orange] (\nx;\ny;\nz)--(\x;\ny;\nz);
				\draw[thick,ph-orange] (\nx;\ny;\nz)--(\nx;\y;\nz);
				\draw[thick,ph-orange] (\nx;\ny;\nz)--(\nx;\ny;\z);
				
				\draw[thick,ph-orange] (\x;\ny;\z)--(\nx;\ny;\z);
				\draw[thick,ph-orange] (\nx;\y;\z)--(\nx;\ny;\z);
				\draw[thick,ph-orange] (\x;\ny;\z)--(\x;\ny;\nz);
				\draw[thick,ph-orange] (\x;\y;\nz)--(\x;\ny;\nz);
				\draw[thick,ph-orange] (\x;\y;\nz)--(\nx;\y;\nz);
				\draw[thick,ph-orange] (\nx;\y;\z)--(\nx;\y;\nz);
				
	} } }
	\foreach \x in {0,...,\LatticeHops}{%
		\foreach \y in {0,...,\LatticeHops}{%
			\foreach \z in {0,...,\LatticeHops}{%
				\coordinate (\x;\y;\z) at (\x*\LatticeScale,\y*\LatticeScale,\z*\LatticeScale);
				\fill[ph-orange] (\x;\y;\z) circle (2.5pt);
	} }}

	\end{scope}
	\begin{scope}[tdplot_rotated_coords, shift={(5,2.5)}]
	\newcommand{\LatticeHops}{2}
	\newcommand{\LatticeScale}{0.8}
	\foreach \x in {0,...,\LatticeHops}{%
		\foreach \y in {0,...,\LatticeHops}{%
			\foreach \z in {0,...,\LatticeHops}{%
				\coordinate (s\x;\y;\z) at (\x*\LatticeScale,\y*\LatticeScale,\z*\LatticeScale);
				\fill[gray] (s\x;\y;\z) circle (1.5pt);
	} } }

	\pgfmathtruncatemacro\NrLines{\LatticeHops-1}
	\foreach \x in {0,...,\NrLines}{%
		\foreach \y in {0,...,\NrLines}{%
			\foreach \z in {0,...,\NrLines}{%
				\pgfmathtruncatemacro\nx{\x+1}
				\pgfmathtruncatemacro\ny{\y+1}
				\pgfmathtruncatemacro\nz{\z+1}
				\draw[thin,gray] (s\x;\y;\z)--(s\nx;\y;\z);
				\draw[thin,gray] (s\x;\y;\z)--(s\x;\ny;\z);
				\draw[thin,gray] (s\x;\y;\z)--(s\x;\y;\nz);
				
				\draw[thin,gray] (s\nx;\ny;\nz)--(s\x;\ny;\nz);
				\draw[thin,gray] (s\nx;\ny;\nz)--(s\nx;\y;\nz);
				\draw[thin,gray] (s\nx;\ny;\nz)--(s\nx;\ny;\z);
				
				rest of lines
				\draw[thin,gray] (s\x;\ny;\z)--(s\nx;\ny;\z);
				\draw[thin,gray] (s\nx;\y;\z)--(s\nx;\ny;\z);
				\draw[thin,gray] (s\x;\ny;\z)--(s\x;\ny;\nz);
				\draw[thin,gray] (s\x;\y;\nz)--(s\x;\ny;\nz);
				\draw[thin,gray] (s\x;\y;\nz)--(s\nx;\y;\nz);
				\draw[thin,gray] (s\nx;\y;\z)--(s\nx;\y;\nz);
				
	} } }
	
	\draw[thick,ph-orange] (s0;0;0)--(0;1;0); 
	\draw[thick,ph-orange] (s0;0;0)--(1;0;1); 
	\draw[thick,ph-orange] (s0;0;0)--(0;0;1);
	\draw[thick,ph-orange,dashed] (s0;0;0)--(1;0;0);
	\draw[thick,ph-orange,dashed] (s0;0;0)--(1;1;0);
	\draw[thick,ph-orange] (s0;0;0)--(0;1;1);
	
	\fill [ph-orange!20, fill opacity = .5]   (s0;0;0)--(1;1;0)--(0;1;0)--cycle;
	\fill [ph-orange!20, fill opacity = .5]   (s0;0;0)--(0;1;0)--(0;1;1)--cycle;
	\fill [ph-orange!20, fill opacity = .5]   (s0;0;0)--(0;1;1)--(0;0;1)--cycle;
	\fill [ph-orange!20, fill opacity = .5]   (s0;0;0)--(0;0;1)--(1;0;1)--cycle;
	\fill [ph-orange!20, fill opacity = .5]   (s0;0;0)--(1;0;1)--(1;0;0)--cycle;
	\fill [ph-orange!20, fill opacity = .5]   (s0;0;0)--(1;0;0)--(1;1;0)--cycle;
	
	\end{scope}
	\end{tikzpicture}
	\caption{Convolution: The neighboring vertices of a lattice are convolved similarly to standard 2D convolutions. If a neighbor is not allocated in the sparse structure, we assume that it has a value of zero.} \label{fig:convolve}
\end{figure}
\egroup

\noindent\textbf{Slicing} is the inverse operation to splatting. The vertex values of the lattice are interpolated back for each position with the same weights used during splatting. The weighted contributions from the simplexes $d+1$ vertices are summed up~(\reffig{fig:slice}).

\subsection{Proposed Operations on Permutohedral Lattice}
The operations defined in Section \refsec{operations_original} are typically used in a cascade of splat-conv-slice to obtain dense predictions~\cite{su2018splatnet}. However, splatting and slicing act as Gaussian kernel low-pass filtering on encoded information~\cite{baek2009some}. Their repeated usage at every layer is detrimental to the accuracy of the network. Additionally, splatting acts as a weighted average on the feature vectors where the weights are only determined through barycentric interpolation. Including the weights as trainable parameter allows the network to decide on a better interpolation scheme. Furthermore, as the network grows deeper and feature vectors become higher-dimensional, slicing consumes increasingly more memory, as it assigns the features to the points. Since in most cases $\abs{P}\gg\abs{V}$, it is more efficient to store the features only in the lattice vertices.

To address these limitations, we propose four new operators on the permutohedral lattice which are more suitable for CNNs and dense prediction tasks.

\noindent\textbf{Distribute} is defined as the list of features that each lattice vertex receives. However, they are not summed as done by splatting:  
\begin{align} \label{eq:splat}
	\mathbf{x}_v &= \mathcal{S}(P,V) = \sum_{p\in J_v} b_{pv} \mathbf{f}_p,
\end{align}
where $\mathbf{x}_v$ is the value of lattice vertex $v$ and $b_{pv}$ is the barycentric weight between point $p$ and lattice vertex $v$.

	 Instead, our distribute operators $\mathcal{D}_G$ and $\mathcal{D}_F$ concatenate coordinates and features of the contributing points:
	 \begin{align} \label{eq:distribute}
		\mathbf{x}_v &= \mathcal{P} ( \mathbf{D}_{v_g} ; \mathbf{D}_{v_f} ),   \\
		\mathbf{D}_{v_g} &= \mathcal{D}_G(P,V) = \{\, \mathbf{g}_p -\bm{\mu}_v  \mid p\in J_v \,\},\\
		\mathbf{D}_{v_f} &= \mathcal{D}_F(P,V) = \{\, \mathbf{f}_p \mid p\in J_v \,\}, \\
		\bm{\mu}_v &= \frac{1}{ \abs{ J_v  } } \sum_{p\in J_v} \mathbf{g}_p,
	\end{align}
where $\mathbf{D}_{v_g} \in\mathbb{R}^{ \abs{ J_v  } \times d }    $ and $\mathbf{D}_{v_f}  \in\mathbb{R}^{ \abs{ J_v  } \times f_d } $ are matrices containing the distributed coordinates and features, respectively, for the contributing points into a vertex $v$. The matrices are concatenated and processed by a PointNet $\mathcal{P}$ to obtain the final vertex value $\mathbf{x}_v$. \reffig{fig:splat_distribute} illustrates the difference between splatting and distributing.
	
	Note that we use a different distribute function for coordinates then for point features. For coordinates, we subtract the mean of the contributing coordinates. The intuition behind this is that coordinates by themselves are not very informative \wrt the potential semantic class. However, the local distribution is more informative as it gives a notion of the geometry.
		
\bgroup
\begin{figure}[]
	\centering
	\captionsetup[subfloat]{farskip=0pt,captionskip=2pt}
	\tdplotsetmaincoords{-35.3}{0}
	\tdplotsetrotatedcoords{0}{-45}{0}
	\subfloat[Splat]{\label{fig:splat}
		\begin{tikzpicture}[tdplot_main_coords,>={Stealth[inset=2pt,length=8pt,angle'=30,round]}] 
		\begin{scope}[tdplot_rotated_coords]
		\newcommand{\LatticeHops}{1}
		\newcommand{\LatticeScale}{1.8}
		\foreach \x in {0,...,\LatticeHops}{%
			\foreach \y in {0,...,\LatticeHops}{%
				\foreach \z in {0,...,\LatticeHops}{%
					\coordinate (\x;\y;\z) at (\x*\LatticeScale,\y*\LatticeScale,\z*\LatticeScale);
					\fill[gray] (\x;\y;\z) circle (1.5pt);
			} }
		}
		
		\pgfmathtruncatemacro\NrLines{\LatticeHops-1}
		\foreach \x in {0,...,\NrLines}{%
			\foreach \y in {0,...,\NrLines}{%
				\foreach \z in {0,...,\NrLines}{%
					\pgfmathtruncatemacro\nx{\x+1}
					\pgfmathtruncatemacro\ny{\y+1}
					\pgfmathtruncatemacro\nz{\z+1}
					\draw[thin,gray] (\x;\y;\z)--(\nx;\y;\z);
					\draw[thin,gray] (\x;\y;\z)--(\x;\ny;\z);
					\draw[thin,gray] (\x;\y;\z)--(\x;\y;\nz);
					
					\draw[thin,gray] (\nx;\ny;\nz)--(\x;\ny;\nz);
					\draw[thin,gray] (\nx;\ny;\nz)--(\nx;\y;\nz);
					\draw[thin,gray] (\nx;\ny;\nz)--(\nx;\ny;\z);
					
					\draw[thin,gray] (\x;\ny;\z)--(\nx;\ny;\z);
					\draw[thin,gray] (\nx;\y;\z)--(\nx;\ny;\z);
					\draw[thin,gray] (\x;\ny;\z)--(\x;\ny;\nz);
					\draw[thin,gray] (\x;\y;\nz)--(\x;\ny;\nz);
					\draw[thin,gray] (\x;\y;\nz)--(\nx;\y;\nz);
					\draw[thin,gray] (\nx;\y;\z)--(\nx;\y;\nz);
					
		} } }
		
		\node[rectangle,fill=ph-orange-light,inner sep=0pt, minimum size=3.5pt] (p1) at (-0.75*\LatticeScale, -0.2*\LatticeScale, 0.0*\LatticeScale) {};
		\node[rectangle,fill=ph-blue-light,inner sep=0pt, minimum size=3.5pt] (p2) at (0.2*\LatticeScale, -0.3*\LatticeScale, 0.0*\LatticeScale) {};
		
		\draw[->,ph-light-gray,shorten >=6pt,shorten <=1pt](p1)--(0;0;0);
		\draw[->,ph-light-gray,shorten >=3pt,shorten <=1pt](p1)--(0;0;1);
		\draw[->,ph-light-gray,shorten >=4pt,shorten <=1pt](p1)--(0;1;1);
		\draw[->,ph-light-gray,shorten >=6pt,shorten <=1pt](p2)--(0;0;0);
		\draw[->,ph-light-gray,shorten >=3pt,shorten <=1pt](p2)--(1;0;0);
		\draw[->,ph-light-gray,shorten >=3pt,shorten <=1pt](p2)--(1;0;1);
		
		\node[circle,fill=ph-purple-light,inner sep=0pt, minimum size=10pt] (L1V1) at (0;0;0) {};
		\node[circle,fill=ph-orange-light,inner sep=0pt, minimum size=5pt] (L2V1) at (0;0;1) {};
		\node[circle,fill=ph-orange-light,inner sep=0pt, minimum size=9pt] (L3V1) at (0;1;1) {};
		\node[circle,fill=ph-blue-light,inner sep=0pt, minimum size=5pt] (L4V1) at (1;0;0) {};
		\node[circle,fill=ph-blue-light,inner sep=0pt, minimum size=6pt] (L5V1) at (1;0;1) {};
		
		\end{scope}
		\end{tikzpicture}
	}
	\hspace{10pt}
	\subfloat[Distribute]{\label{fig:distribute}
		\begin{tikzpicture}[tdplot_main_coords, >={Stealth[inset=2pt,length=8pt,angle'=30,round]}]
		\begin{scope}[tdplot_rotated_coords]
		\newcommand{\LatticeHops}{1}
		\newcommand{\LatticeScale}{1.8}
		\foreach \x in {0,...,\LatticeHops}{%
			\foreach \y in {0,...,\LatticeHops}{%
				\foreach \z in {0,...,\LatticeHops}{%
					\coordinate (\x;\y;\z) at (\x*\LatticeScale,\y*\LatticeScale,\z*\LatticeScale);
					\fill[gray] (\x;\y;\z) circle (1.5pt);
			} }
		}
		
		\pgfmathtruncatemacro\NrLines{\LatticeHops-1}
		\foreach \x in {0,...,\NrLines}{%
			\foreach \y in {0,...,\NrLines}{%
				\foreach \z in {0,...,\NrLines}{%
					\pgfmathtruncatemacro\nx{\x+1}
					\pgfmathtruncatemacro\ny{\y+1}
					\pgfmathtruncatemacro\nz{\z+1}
					\draw[thin,gray] (\x;\y;\z)--(\nx;\y;\z);
					\draw[thin,gray] (\x;\y;\z)--(\x;\ny;\z);
					\draw[thin,gray] (\x;\y;\z)--(\x;\y;\nz);
					
					\draw[thin,gray] (\nx;\ny;\nz)--(\x;\ny;\nz);
					\draw[thin,gray] (\nx;\ny;\nz)--(\nx;\y;\nz);
					\draw[thin,gray] (\nx;\ny;\nz)--(\nx;\ny;\z);
					
					\draw[thin,gray] (\x;\ny;\z)--(\nx;\ny;\z);
					\draw[thin,gray] (\nx;\y;\z)--(\nx;\ny;\z);
					\draw[thin,gray] (\x;\ny;\z)--(\x;\ny;\nz);
					\draw[thin,gray] (\x;\y;\nz)--(\x;\ny;\nz);
					\draw[thin,gray] (\x;\y;\nz)--(\nx;\y;\nz);
					\draw[thin,gray] (\nx;\y;\z)--(\nx;\y;\nz);
					
		} } }
		
		\node[rectangle,fill=ph-orange-light,inner sep=0pt, minimum size=3.5pt] (p1) at (-0.75*\LatticeScale, -0.2*\LatticeScale, 0.0*\LatticeScale) {};
		\node[rectangle,fill=ph-blue-light,inner sep=0pt, minimum size=3.5pt] (p2) at (0.2*\LatticeScale, -0.3*\LatticeScale, 0.0*\LatticeScale) {};
		
		\draw[->,ph-light-gray,shorten >=9pt,shorten <=1pt](p1)--(0;0;0);
		\draw[->,ph-light-gray,shorten >=3pt,shorten <=1pt](p1)--(0;0;1);
		\draw[->,ph-light-gray,shorten >=4pt,shorten <=1pt](p1)--(0;1;1);
		\draw[->,ph-light-gray,shorten >=8pt,shorten <=1pt](p2)--(0;0;0);
		\draw[->,ph-light-gray,shorten >=3pt,shorten <=1pt](p2)--(1;0;0);
		\draw[->,ph-light-gray,shorten >=3pt,shorten <=1pt](p2)--(1;0;1);
		
		\def\R{8pt}
		\draw (0,0) circle[radius=\R];
		\fill [ph-orange-light, fill opacity = 1.0]   (180:\R) arc (180:0:\R) -- cycle; 
		\fill [ph-blue-light, fill opacity = 1.0]   (180:\R) arc (180:360:\R) -- cycle;
		
		\node[circle,fill=ph-orange-light,inner sep=0pt, minimum size=5pt] (L2V1) at (0;0;1) {};
		\node[circle,fill=ph-orange-light,inner sep=0pt, minimum size=9pt] (L3V1) at (0;1;1) {};
		\node[circle,fill=ph-blue-light,inner sep=0pt, minimum size=5pt] (L4V1) at (1;0;0) {};
		\node[circle,fill=ph-blue-light,inner sep=0pt, minimum size=6pt] (L5V1) at (1;0;1) {};
		
		\end{scope}
		\end{tikzpicture}
	}
	\caption{Splat and Distribute operations: Splatting uses barycentric weighting to add the features of points onto neighboring vertices. The na\"{\i}ve summation can be detrimental to the network as splatting acts as a Gaussian filter. Distributing stores all the features of the contributing points, causing no loss of information and allows further processing by the network.} \label{fig:splat_distribute}
\end{figure}
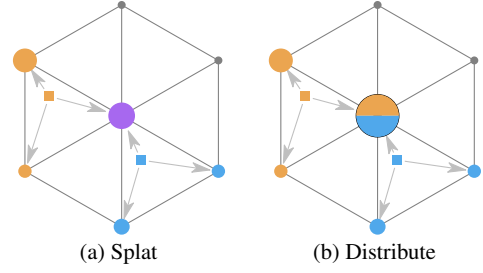

\egroup
	
\noindent\textbf{Downsampling} refers to a coarsening of the lattice, by reducing the number of vertices. This allows the network to capture more contextual information. Downsampling consists of two steps: creation of a coarse lattice and obtaining its values.
	Coarse lattices are created by repeatedly dividing the point cloud positions by 2 and using them to create new lattice vertices~\cite{barron2015fast}.
	The values of the coarse lattice are obtained by convolving over the finer lattice from the previous level~(\reffig{fig:coarsen}).
	Hence, we must embed the coarse lattice inside the finer one by scaling the coarse vertices by 2. Afterwards, the neighbors vertices over which we convolve are separated by a vector of form $\pm \left[ -1,\ldots,-1,d,-1,\ldots,-1  \right]\in\mathbb{Z}^{d+1}$. The downsampling operation effectively performs a strided convolution.

	\bgroup
	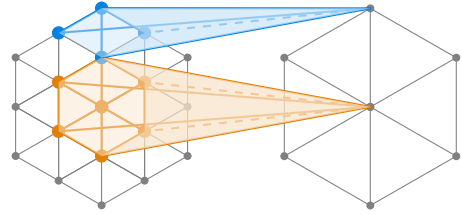
\begin{figure}[]
		\centering
		\tdplotsetmaincoords{-35.3}{0}
		\tdplotsetrotatedcoords{0}{-45}{0}
		\begin{tikzpicture}[tdplot_main_coords, remember picture, >={Stealth[inset=10pt,length=8pt,angle'=28,round]} ]
		\begin{scope}[tdplot_rotated_coords]
		\newcommand{\LatticeHops}{2}
		\newcommand{\LatticeScale}{0.8}
		\foreach \x in {0,...,\LatticeHops}{%
			\foreach \y in {0,...,\LatticeHops}{%
				\foreach \z in {0,...,\LatticeHops}{%
					\coordinate [tdplot_rotated_coords] (\x;\y;\z) at (\x*\LatticeScale,\y*\LatticeScale,\z*\LatticeScale);
					\fill[gray] (\x;\y;\z) circle (1.5pt);
		} }}
		
		\pgfmathtruncatemacro\NrLines{\LatticeHops-1}
		\foreach \x in {0,...,\NrLines}{%
			\foreach \y in {0,...,\NrLines}{%
				\foreach \z in {0,...,\NrLines}{%
					\pgfmathtruncatemacro\nx{\x+1}
					\pgfmathtruncatemacro\ny{\y+1}
					\pgfmathtruncatemacro\nz{\z+1}
					\draw[thin,gray] (\x;\y;\z)--(\nx;\y;\z);
					\draw[thin,gray] (\x;\y;\z)--(\x;\ny;\z);
					\draw[thin,gray] (\x;\y;\z)--(\x;\y;\nz);
					
					\draw[thin,gray] (\nx;\ny;\nz)--(\x;\ny;\nz);
					\draw[thin,gray] (\nx;\ny;\nz)--(\nx;\y;\nz);
					\draw[thin,gray] (\nx;\ny;\nz)--(\nx;\ny;\z);
					
					\draw[thin,gray] (\x;\ny;\z)--(\nx;\ny;\z);
					\draw[thin,gray] (\nx;\y;\z)--(\nx;\ny;\z);
					\draw[thin,gray] (\x;\ny;\z)--(\x;\ny;\nz);
					\draw[thin,gray] (\x;\y;\nz)--(\x;\ny;\nz);
					\draw[thin,gray] (\x;\y;\nz)--(\nx;\y;\nz);
					\draw[thin,gray] (\nx;\y;\z)--(\nx;\y;\nz);
					
		} } }
		%
		\renewcommand{\LatticeHops}{1} 
		\pgfmathtruncatemacro\NrLines{\LatticeHops-1}
		\foreach \x in {0,...,\NrLines}{%
			\foreach \y in {0,...,\NrLines}{%
				\foreach \z in {0,...,\NrLines}{%
					\pgfmathtruncatemacro\nx{\x+1}
					\pgfmathtruncatemacro\ny{\y+1}
					\pgfmathtruncatemacro\nz{\z+1}
					\draw[thick,ph-orange] (\x;\y;\z)--(\nx;\y;\z);
					\draw[thick,ph-orange] (\x;\y;\z)--(\x;\ny;\z);
					\draw[thick,ph-orange] (\x;\y;\z)--(\x;\y;\nz);
					
					\draw[thick,ph-orange] (\nx;\ny;\nz)--(\x;\ny;\nz);
					\draw[thick,ph-orange] (\nx;\ny;\nz)--(\nx;\y;\nz);
					\draw[thick,ph-orange] (\nx;\ny;\nz)--(\nx;\ny;\z);
					
					\draw[thick,ph-orange] (\x;\ny;\z)--(\nx;\ny;\z);
					\draw[thick,ph-orange] (\nx;\y;\z)--(\nx;\ny;\z);
					\draw[thick,ph-orange] (\x;\ny;\z)--(\x;\ny;\nz);
					\draw[thick,ph-orange] (\x;\y;\nz)--(\x;\ny;\nz);
					\draw[thick,ph-orange] (\x;\y;\nz)--(\nx;\y;\nz);
					\draw[thick,ph-orange] (\nx;\y;\z)--(\nx;\y;\nz);
					
		} } }
		\foreach \x in {0,...,\LatticeHops}{%
			\foreach \y in {0,...,\LatticeHops}{%
				\foreach \z in {0,...,\LatticeHops}{%
					\coordinate (\x;\y;\z) at (\x*\LatticeScale,\y*\LatticeScale,\z*\LatticeScale);
					\fill[ph-orange] (\x;\y;\z) circle (2.5pt);
		} }}
		
		\fill[ph-blue] (0;2;0) circle (2.5pt);
		\fill[ph-blue] (1;2;0) circle (2.5pt);
		\fill[ph-blue] (0;2;1) circle (2.5pt);
		\fill[ph-blue] (0;1;0) circle (2.5pt);
		\draw[thick,ph-blue] (0;2;0)--(1;2;0);
		\draw[thick,ph-blue] (0;2;0)--(0;1;0);
		\draw[thick,ph-blue] (0;2;0)--(0;2;1);

		\end{scope}
		\begin{scope}[tdplot_rotated_coords, shift={(5,2.5)}]
		\newcommand{\LatticeHops}{1}
		\newcommand{\LatticeScale}{1.6}
		\foreach \x in {0,...,\LatticeHops}{%
			\foreach \y in {0,...,\LatticeHops}{%
				\foreach \z in {0,...,\LatticeHops}{%
					\coordinate (s\x;\y;\z) at (\x*\LatticeScale,\y*\LatticeScale,\z*\LatticeScale);
					\fill[gray] (s\x;\y;\z) circle (1.5pt);
		} } }

		\pgfmathtruncatemacro\NrLines{\LatticeHops-1}
		\foreach \x in {0,...,\NrLines}{%
			\foreach \y in {0,...,\NrLines}{%
				\foreach \z in {0,...,\NrLines}{%
					\pgfmathtruncatemacro\nx{\x+1}
					\pgfmathtruncatemacro\ny{\y+1}
					\pgfmathtruncatemacro\nz{\z+1}
					\draw[thin,gray] (s\x;\y;\z)--(s\nx;\y;\z);
					\draw[thin,gray] (s\x;\y;\z)--(s\x;\ny;\z);
					\draw[thin,gray] (s\x;\y;\z)--(s\x;\y;\nz);
					
					\draw[thin,gray] (s\nx;\ny;\nz)--(s\x;\ny;\nz);
					\draw[thin,gray] (s\nx;\ny;\nz)--(s\nx;\y;\nz);
					\draw[thin,gray] (s\nx;\ny;\nz)--(s\nx;\ny;\z);
					
					rest of lines
					\draw[thin,gray] (s\x;\ny;\z)--(s\nx;\ny;\z);
					\draw[thin,gray] (s\nx;\y;\z)--(s\nx;\ny;\z);
					\draw[thin,gray] (s\x;\ny;\z)--(s\x;\ny;\nz);
					\draw[thin,gray] (s\x;\y;\nz)--(s\x;\ny;\nz);
					\draw[thin,gray] (s\x;\y;\nz)--(s\nx;\y;\nz);
					\draw[thin,gray] (s\nx;\y;\z)--(s\nx;\y;\nz);
					
		} } }
		
		\draw[thick,ph-orange] (s0;0;0)--(0;1;0); 
		\draw[thick,ph-orange] (s0;0;0)--(1;0;1); 
		\draw[thick,ph-orange] (s0;0;0)--(0;0;1);
		\draw[thick,ph-orange,dashed] (s0;0;0)--(1;0;0);
		\draw[thick,ph-orange,dashed] (s0;0;0)--(1;1;0);
		\draw[thick,ph-orange] (s0;0;0)--(0;1;1);
		
		\fill [ph-orange!20, fill opacity = .5]   (s0;0;0)--(1;1;0)--(0;1;0)--cycle;
		\fill [ph-orange!20, fill opacity = .5]   (s0;0;0)--(0;1;0)--(0;1;1)--cycle;
		\fill [ph-orange!20, fill opacity = .5]   (s0;0;0)--(0;1;1)--(0;0;1)--cycle;
		\fill [ph-orange!20, fill opacity = .5]   (s0;0;0)--(0;0;1)--(1;0;1)--cycle;
		\fill [ph-orange!20, fill opacity = .5]   (s0;0;0)--(1;0;1)--(1;0;0)--cycle;
		\fill [ph-orange!20, fill opacity = .5]   (s0;0;0)--(1;0;0)--(1;1;0)--cycle;
		
		\draw[thick,ph-blue] (s0;1;0)--(0;1;0); 
		\draw[thick,ph-blue] (s0;1;0)--(0;2;1);
		\draw[thick,ph-blue,dashed] (s0;1;0)--(1;2;0);
		\draw[thick,ph-blue] (s0;1;0)--(0;2;0);	
		
		\fill [ph-blue!20, fill opacity = .5]   (s0;1;0)--(0;2;1)--(0;1;0)--cycle;
		\fill [ph-blue!20, fill opacity = .5]   (s0;1;0)--(0;1;0)--(1;2;0)--cycle;
		\fill [ph-blue!20, fill opacity = .5]   (s0;1;0)--(0;2;0)--(1;2;0)--cycle;
		\fill [ph-blue!20, fill opacity = .5]   (s0;1;0)--(0;2;0)--(0;2;1)--cycle;
		
		\end{scope}
		\end{tikzpicture}
		\caption{Coarsen: Downsampling of the lattice is performed by embedding the coarse lattice in the finer one and convolving over the neighbors. This effectively performs a strided convolution. Transposed convolution is performed in an analogous manner by embedding a fine lattice into a coarse one.} \label{fig:coarsen}
	\end{figure}
	\egroup

\noindent\textbf{Upsampling} follows a similar reasoning. The fine vertices need first to be embedded in the coarse lattice using a division by 2. Afterwards, the neighboring vertices over which we convolve are separated by a vector of form $\pm \left[ -0.5,\ldots,-0.5,d/2,-0.5,\ldots,-0.5  \right]$. The careful reader will notice that in this case, the coordinates of the neighboring vertices may not be integer anymore; they may have a fractional part and will, therefore, lie in the middle of a coarser simplex. In this case we ignore the contribution of this neighboring vertices and only take the contribution of the center vertex.\\
The upsampling operation effectively performs a transposed convolution.

\noindent\textbf{DeformSlicing:}
	While the slicing operation $\mathcal{Y}$ barycentrically interpolates the values back to the points by using barycentric coordinates:
	\begin{align} \label{eq:slice}
	f_p &= \mathcal{Y}(P,V) = \sum_{v\in I_p} b_{pv} \mathbf{x}_v,
	\end{align}
	we propose the DeformSlicing $\mathcal{\tilde{Y}}$ which allows the network to directly modify the barycentric coordinates and shift the position within the simplex for data-dependent interpolation:
	\begin{align} \label{eq:deformSlice}
		f_p &= \mathcal{\tilde{Y}}(P,V) = \sum_{v\in I_p} (b_{pv} + \Delta b_{pv}) \mathbf{x}_v.
	\end{align}
	Here, $\Delta b_{pv}$ are offsets that are applied to the original barycentric coordinates. A parallel branch within our network first gathers the values from all the vertices in a simplex and regresses the $\Delta b_{pv}$:
	\begin{align}
		\mathbf{q}_p &= \mathcal{G}(P,V) = \{\, b_{pv} \mathbf{x}_v \mid v\in I_p \,\}, \\
		\Delta \mathbf{b}_p &= \mathcal{F}( \mathbf{q}_p ),
	\end{align}
	

	where $\mathbf{q}_p$ is a set containing the weighted values of all the vertices of the simplex containing $p$ and the prediction $\Delta \mathbf{b}_p=\{\, \Delta b_{pv} \mid v\in I_p \,\}$ is a set of offsets to the barycentric coordinates towards the $d+1$ vertices. \\
	With a slight abuse of notation --- due to the fact that the vertices of a simplex are always enumerated in a consistent manner, we can regard $\mathbf{b}_p$ and $\mathbf{q}_p$ as vectors in $\mathbb{R}^{ (d+1) }$  and  $\mathbb{R}^{ (d+1)v_d }$, respectively, and cast the prediction of offsets as a fully connected layer followed by a non-linearity:
	\begin{align}
	\Delta \mathbf{b}_p &= \mathcal{F}( \mathbf{q}_p ) = \sigma( \mathbf{q}_p \cdot \mathbf{W} + b ).
	\end{align}
	
	However, this prediction has the disadvantage of not being permutation equivariant; therefore, permutation of the vertices would not imply the same permutation in the barycentric offsets:
	\begin{align}
 	\mathcal{F}( \pi \mathbf{q}_p ) \neq \pi \mathcal{F}( \mathbf{q}_p ),
	\end{align}
	where $\pi$ is the set of all permutations of the $d+1$ vertices.
	
	
	It is important for our prediction to be permutation equivariant because the vertices may be arranged in any order and the barycentric offsets need to keep a consistent preference towards a certain vertexes’ features, regardless of its position within a simplex. 
	
	In order for the prediction of the offsets to be consistent with permutations of the vertices, we take inspiration from the work of \cite{Ravanbakhsh2016DeepLW} and \cite{zaheer2017deep} of equivariant layers and design $\mathcal{F}$ as:
		\begin{align}
		\Delta b_{pv} &= \sigma( b+  (b_{pv} \mathbf{x}_v - \max\limits_{d\in I_p}\{b_{pd} \mathbf{x}_d\} )  \cdot  \mathbf{W} ),\\
		\Delta \mathbf{b}_p &= \mathcal{F}( \mathbf{q}_p ) = \{\, \Delta b_{pv} \mid v\in I_p \,\},
	\end{align} 
	
	where $\mathbf{W} \in\mathbb{R}^{ v_d \times 1 }  $ is a weight matrix and $b \in\mathbb{R}  $ corresponds to a scalar bias. \\
	In other words, we subtract from each weighted vertex the maximum of the weighted values of all the other vertices in the simplex. Since the max operation is invariant to permutations of the input, the regression of the offsets is \textit{equivariant} to permutations of the vertices.

	The difference between the slicing and our DeformSlicing is visualized in~\reffig{fig:slice_deformSlice}
	
	\bgroup
	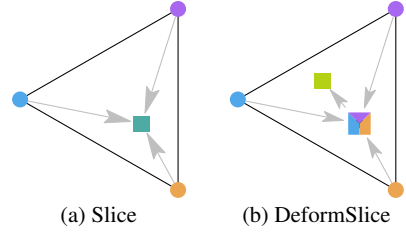
\begin{figure}[]
		\captionsetup[subfloat]{farskip=0pt,captionskip=2pt}
		\centering
		\subfloat[Slice]{\label{fig:slice}
			
			\begin{tikzpicture}[scale=0.8, >={Stealth[inset=2pt,length=11pt,angle'=28,round]} ]
			\coordinate (L1) at (0,0);
			\coordinate (L2) at (2.598,-1.5); 
			\coordinate (L3) at (2.598,1.5);
			
			\draw (L1) -- (L2) -- (L3) -- cycle;
			\fill[gray] (L1) circle (1.5pt);
			\fill[gray] (L2) circle (1.5pt);
			\fill[gray] (L3) circle (1.5pt);
			
			\node[circle,fill=ph-blue-light,inner sep=0pt, minimum size=6pt] (V1) at (L1) {};
			\node[circle,fill=ph-orange-light,inner sep=0pt, minimum size=6pt] (V2) at (L2) {};
			\node[circle,fill=ph-purple-light,inner sep=0pt, minimum size=6pt] (V3) at (L3) {};
			
			\node[circle,fill=black,inner sep=0pt, minimum size=3.5pt] (p1) at (2.0,-0.4) {};
			
			\draw[->,ph-light-gray,shorten >=2pt,shorten <=3pt](L1)--(p1);
			\draw[->,ph-light-gray,shorten >=2pt,shorten <=3pt](L2)--(p1);
			\draw[->,ph-light-gray,shorten >=2pt,shorten <=3pt](L3)--(p1);
			
			\node[rectangle,fill=ph-green-light,inner sep=0pt, minimum size=6pt] (SV1) at (p1) {};
			\end{tikzpicture}
		}
		\hspace{10pt}
		\subfloat[DeformSlice]{\label{fig:deformSlice}
			\begin{tikzpicture}[scale=0.8, >={Stealth[inset=2pt,length=9pt,angle'=28,round]} ]
			\coordinate (L1) at (0,0);
			\coordinate (L2) at (2.598,-1.5); 
			\coordinate (L3) at (2.598,1.5);
			
			\draw (L1) -- (L2) -- (L3) -- cycle;
			\fill[gray] (L1) circle (1.5pt);
			\fill[gray] (L2) circle (1.5pt);
			\fill[gray] (L3) circle (1.5pt);
			
			\node[circle,fill=ph-blue-light,inner sep=0pt, minimum size=6pt] (V1) at (L1) {};
			\node[circle,fill=ph-orange-light,inner sep=0pt, minimum size=6pt] (V2) at (L2) {};
			\node[circle,fill=ph-purple-light,inner sep=0pt, minimum size=6pt] (V3) at (L3) {};
			
			\node[circle,fill=black,inner sep=0pt, minimum size=3.5pt] (p1) at (2.0,-0.4) {};
			
			\draw[->,ph-light-gray,shorten >=4pt,shorten <=3pt](L1)--(p1);
			\draw[->,ph-light-gray,shorten >=4pt,shorten <=3pt](L2)--(p1);
			\draw[->,ph-light-gray,shorten >=4pt,shorten <=3pt](L3)--(p1);
			
			\node[draw=none,minimum size=3pt,regular polygon,regular polygon sides=4] (a) at (p1) {};
			\fill [ph-blue-light, fill opacity = 1.0]   (a.south)--(a.center)--(a.corner 2)--(a.corner 3)--cycle;
			\fill [ph-orange-light, fill opacity = 1.0]   (a.center)--(a.south)--(a.corner 4)--(a.corner 1)--cycle;
			\fill [ph-purple-light, fill opacity = 1.0]   (a.center)--(a.corner 1)--(a.corner 2)--cycle;

			\node[circle,fill=black,inner sep=0pt, minimum size=3.5pt] (p2) at (1.4,0.3) {};
			
			\draw[->,ph-light-gray,shorten >=2pt,shorten <=6pt](p1)--(p2);
			
			\node[rectangle,fill=ph-light-green,inner sep=0pt, minimum size=6pt] (Vfinal) at (p2) {};
			\end{tikzpicture}
		}
		\caption{Slice and DeformSlice: Slicing barycentrically interpolates the vertex values back onto a point. DeformSlice allows for the network to directly affect the interpolated value by learning offsets of the barycentric coordinates.} \label{fig:slice_deformSlice}
	\end{figure}
	\egroup

\section{Segmentation Methods} 	

Due to the flexibility of LatticeNet various segmentation methods can be implemented. In this section, we detail the methods used for each one.
\subsection{Semantic Segmentation}
	
	Semantic segmentation uses the default U-Net architecture described in the~\nameref{section:arch} section.
	It is trained with an equal part combination of cross entropy loss and Lov\'asz loss~\cite{berman2018lovasz}. The Lov\'asz loss acts as a surrogate for the intersection-over-union score and is especially useful for dealing with class imbalance. 


\subsection{Instance Segmentation}
	
	Our instance segmentation network follows the work of other proposal-free methods like~\cite{de2017semantic}. 
	We use LatticeNet to predict for each 3D point $p_i$ in the point cloud an embedding $x_i$. A discriminative loss encourages closeness in embeddings space for points of the same instance while promoting distance between different instances. Finally, we apply mean-shift clustering on the points in embeddings space. Points belonging to the same cluster are defined as an Instances.
	
	This discriminative loss can be expressed with three terms: 
	\begin{itemize}
		\item Variance term: The intra-cluster pull force that draws the embeddings towards the mean embedding.
		\item Distance term: An inter-cluster push force that forces the clusters to be far apart from each other in embedding space.
		\item Regularization term: A small force that pulls the cluster centers towards the origin in order to keep the activations bounded.
	\end{itemize}
	
	The full loss is then defined as:
	\begin{align} \label{eq:var}
	L_{var} &= \frac{1}{C} \sum_{c=1}^{C} \frac{1}{N_c} \sum_{i=1}^{N_c} \left[  \norm{\mu_c -x_i}  -\delta_v \right] ^2\\
	\label{eq:dist}
	L_{dist} &= \frac{1}{C (C-1)} \mathop{\sum_{c_A = 1}^{C} \sum_{c_B = 1}^{C}}_{c_A \neq c_B} \left[ 2 \delta_{\textrm{d}} - \lVert \mu_{c_A} - \mu_{c_B} \rVert \right]_{+}^2\\
	L_{reg} &= \frac{1}{C} \sum_{c=1}^{C} \lVert \mu_{c} \rVert\\
	L &= \alpha \cdot L_{var} + \beta \cdot L_{dist} + \gamma \cdot L_{reg}
	\end{align}
	
	We define $C$ as the number of clusters in the ground truth, $N_c$ as the number of elements in cluster $c$, $x_i$ as the embedding vector for point $p_i$ and $\mu_c$ as the mean or cluster center for cluster $c$. 
	The $\delta_{\textrm{v}}$ and $\delta_{\textrm{d}}$ are the margins for the variance and distance loss respectively.
	We set $\alpha = \beta = 1$ and $\gamma = 0.001$
	
\begin{figure*}[t]
		
		\centering
		\tdplotsetmaincoords{-35.3}{0}
		\tdplotsetrotatedcoords{0}{-45}{0}
		\begin{tikzpicture}[tdplot_main_coords, remember picture, >={Stealth[inset=1pt,length=8pt,angle'=30,round]} ]

		\begin{scope}
		\def\solidImg{./imgs/instance_seg/cloud_solid.png}
		\newlength{\WSS}
		\newlength{\HSS}
		\settowidth{\WSS}{\includegraphics{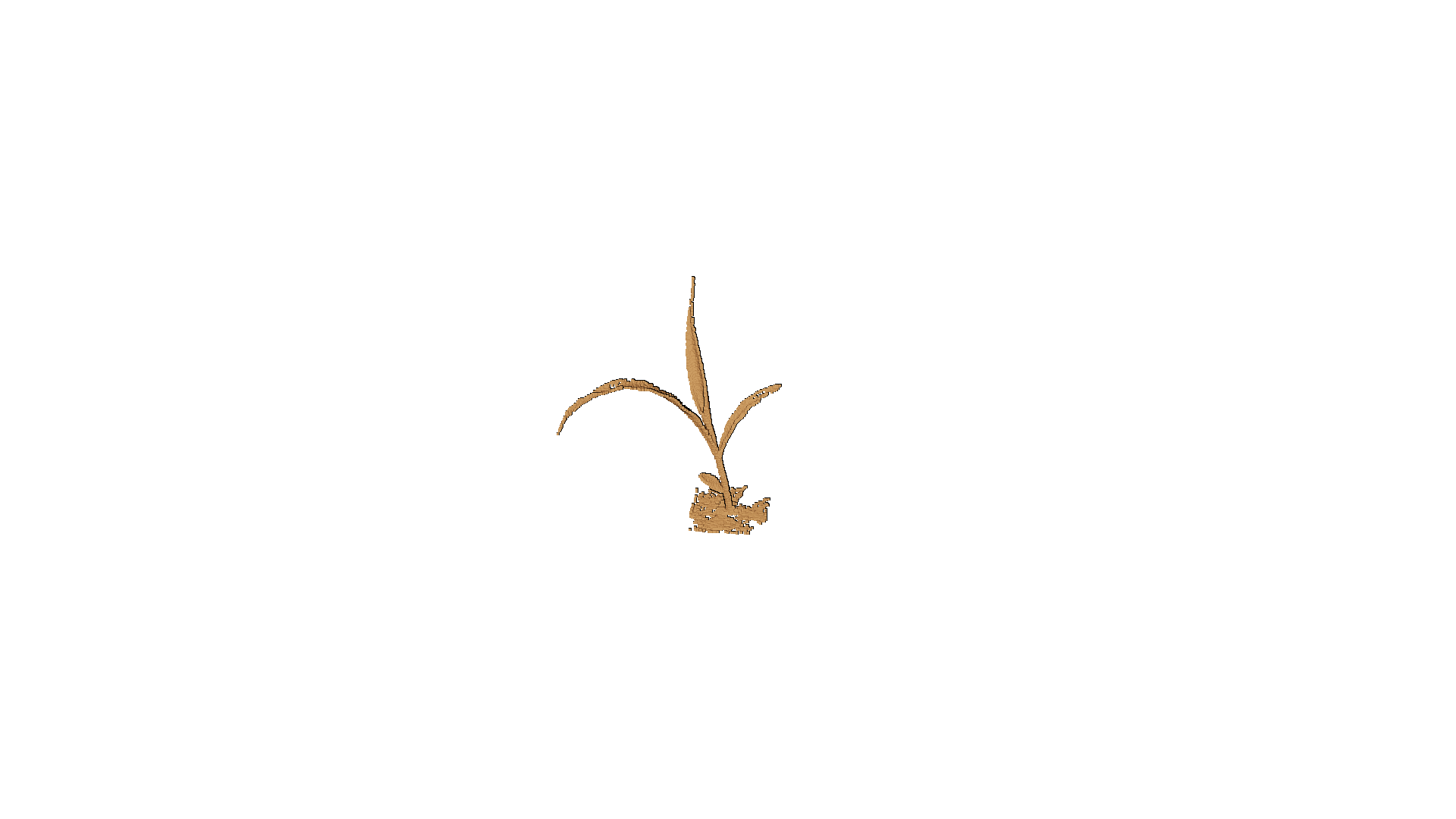}}
		\settoheight{\HSS}{\includegraphics{\solidImg}}
		
		\node[inner sep=0pt] (russell) at (-3.2, 0.5 ) {\includegraphics[trim=.33\WSS{} .33\HSS{} .425\WSS{} .3\HSS{},clip, width=.25\textwidth]{\solidImg}};
		
		\end{scope}

		\node[] (i10) at (-2.5,0.5){};
		\node[] (i11) at (-1.9,0.8){};
		\node[] (i12) at (-1.1,1.2){};
		\node[] (i13) at (-0.2,0.4){};
		\fill[ph-blue!70] (i13) circle (1.5pt);
		\begin{scope}[decoration={
			markings,
			mark=at position 0.5 with {\arrow[rotate=20]{>}}}
		] 
		\draw [gray, dashed, postaction={decorate}] plot [smooth, tension=0.8] coordinates { (i10) (i11) (i12) (i13) };
		\end{scope}

		\node[] (i20) at (-2.2,-0.5){};
		\node[] (i21) at (-1.5,-0.55){};
		\node[] (i22) at (-0.9,-0.3){};
		\node[] (i23) at (-0.45,-0.0){};
		\fill[ph-blue!70] (i23) circle (1.5pt);
		\begin{scope}[decoration={
			markings,
			mark=at position 0.4 with {\arrow{>}}}
		] 
		\draw [gray, dashed, postaction={decorate}] plot [smooth, tension=0.6] coordinates { (i20) (i21) (i22) (i23) };
		\end{scope}
		
		\node[] (o10) at (2.7,0.5){};
		\node[] (o11) at (1.9,0.8){};
		\node[] (o12) at (1.1,1.2){};
		\node[] (o13) at (0.2,0.4){};
		\fill[ph-blue!70] (o13) circle (1.5pt);
		\begin{scope}[decoration={
			markings,
			mark=at position 0.5 with {\arrow[rotate=-15]{<}}}
		] 
		\draw [gray, dashed, postaction={decorate}] plot [smooth, tension=0.6] coordinates { (o10) (o11) (o12) (o13) };
		\end{scope}
		
		\node[] (o20) at (2.0,-0.65){};
		\node[] (o21) at (1.5,-0.55){};
		\node[] (o22) at (0.9,-0.3){};
		\node[] (o23) at (0.45,-0.0){};
		\fill[ph-blue!70] (o23) circle (1.5pt);
		\begin{scope}[decoration={
			markings,
			mark=at position 0.35 with {\arrow{<}}}
		] 
		\draw [gray, dashed, postaction={decorate}] plot [smooth, tension=0.6] coordinates { (o20) (o21) (o22) (o23) };
		\end{scope}

		\begin{scope}[tdplot_rotated_coords]
		\newcommand{\LatticeHops}{2}
		\newcommand{\LatticeScale}{0.8}
		\foreach \x in {0,...,\LatticeHops}{%
			\foreach \y in {0,...,\LatticeHops}{%
				\foreach \z in {0,...,\LatticeHops}{%
					\coordinate [tdplot_rotated_coords] (\x;\y;\z) at (\x*\LatticeScale,\y*\LatticeScale,\z*\LatticeScale);
					\fill[gray] (\x;\y;\z) circle (1.5pt);
		} }}
		
		\pgfmathtruncatemacro\NrLines{\LatticeHops-1}
		\foreach \x in {0,...,\NrLines}{%
			\foreach \y in {0,...,\NrLines}{%
				\foreach \z in {0,...,\NrLines}{%
					\pgfmathtruncatemacro\nx{\x+1}
					\pgfmathtruncatemacro\ny{\y+1}
					\pgfmathtruncatemacro\nz{\z+1}
					\draw[thin,gray] (\x;\y;\z)--(\nx;\y;\z);
					\draw[thin,gray] (\x;\y;\z)--(\x;\ny;\z);
					\draw[thin,gray] (\x;\y;\z)--(\x;\y;\nz);
					
					\draw[thin,gray] (\nx;\ny;\nz)--(\x;\ny;\nz);
					\draw[thin,gray] (\nx;\ny;\nz)--(\nx;\y;\nz);
					\draw[thin,gray] (\nx;\ny;\nz)--(\nx;\ny;\z);
					
					\draw[thin,gray] (\x;\ny;\z)--(\nx;\ny;\z);
					\draw[thin,gray] (\nx;\y;\z)--(\nx;\ny;\z);
					\draw[thin,gray] (\x;\ny;\z)--(\x;\ny;\nz);
					\draw[thin,gray] (\x;\y;\nz)--(\x;\ny;\nz);
					\draw[thin,gray] (\x;\y;\nz)--(\nx;\y;\nz);
					\draw[thin,gray] (\nx;\y;\z)--(\nx;\y;\nz);
					
		} } }
		%
		\renewcommand{\LatticeHops}{1} 
		\pgfmathtruncatemacro\NrLines{\LatticeHops-1}
		\foreach \x in {0,...,\NrLines}{%
			\foreach \y in {0,...,\NrLines}{%
				\foreach \z in {0,...,\NrLines}{%
					\pgfmathtruncatemacro\nx{\x+1}
					\pgfmathtruncatemacro\ny{\y+1}
					\pgfmathtruncatemacro\nz{\z+1}
					\draw[thick,ph-orange] (\x;\y;\z)--(\nx;\y;\z);
					\draw[thick,ph-orange] (\x;\y;\z)--(\x;\ny;\z);
					\draw[thick,ph-orange] (\x;\y;\z)--(\x;\y;\nz);
					
					\draw[thick,ph-orange] (\nx;\ny;\nz)--(\x;\ny;\nz);
					\draw[thick,ph-orange] (\nx;\ny;\nz)--(\nx;\y;\nz);
					\draw[thick,ph-orange] (\nx;\ny;\nz)--(\nx;\ny;\z);
					
					\draw[thick,ph-orange] (\x;\ny;\z)--(\nx;\ny;\z);
					\draw[thick,ph-orange] (\nx;\y;\z)--(\nx;\ny;\z);
					\draw[thick,ph-orange] (\x;\ny;\z)--(\x;\ny;\nz);
					\draw[thick,ph-orange] (\x;\y;\nz)--(\x;\ny;\nz);
					\draw[thick,ph-orange] (\x;\y;\nz)--(\nx;\y;\nz);
					\draw[thick,ph-orange] (\nx;\y;\z)--(\nx;\y;\nz);
					
		} } }
		\foreach \x in {0,...,\LatticeHops}{%
			\foreach \y in {0,...,\LatticeHops}{%
				\foreach \z in {0,...,\LatticeHops}{%
					\coordinate (\x;\y;\z) at (\x*\LatticeScale,\y*\LatticeScale,\z*\LatticeScale);
					\fill[ph-orange] (\x;\y;\z) circle (2.5pt);
		} }}

		\end{scope}

		\begin{scope}
		\def\segImg{./imgs/instance_seg/emb.png} 
		\newlength{\WGS}
		\newlength{\HGS}
		\settowidth{\WGS}{\includegraphics{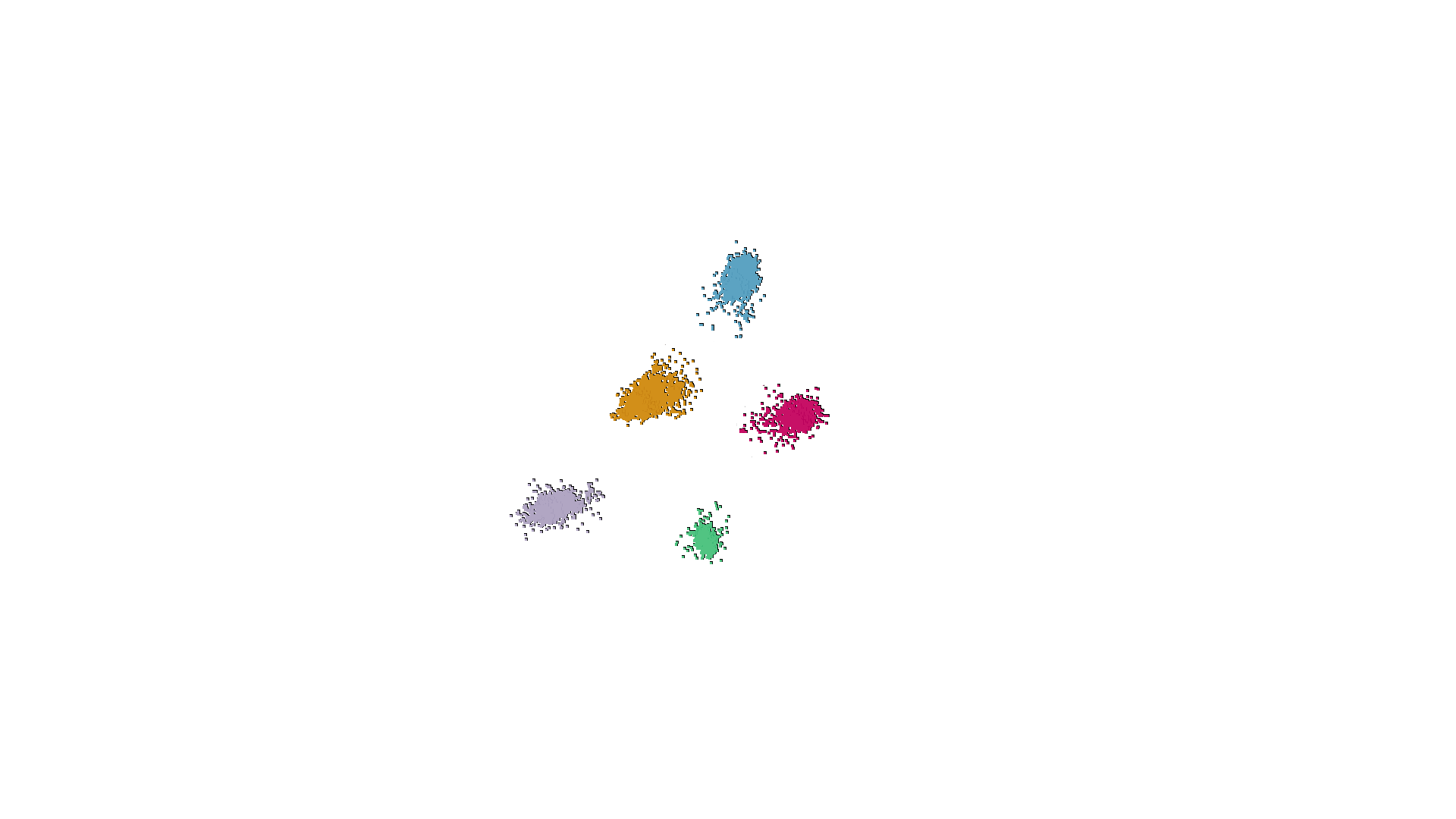}}
		\settoheight{\HGS}{\includegraphics{\segImg}}
		\node[inner sep=0pt] (russell) at (3.8,0) {\includegraphics[trim=.3\WGS{} .25\HGS{} .3\WGS{} .3\HGS{},clip, width=.3\textwidth]{\segImg}};
		\end{scope}
		
		\begin{scope}
		\def\segImg{./imgs/instance_seg/cloud_pred.png} 
		\newlength{\WC}
		\newlength{\HC}
		\settowidth{\WC}{\includegraphics{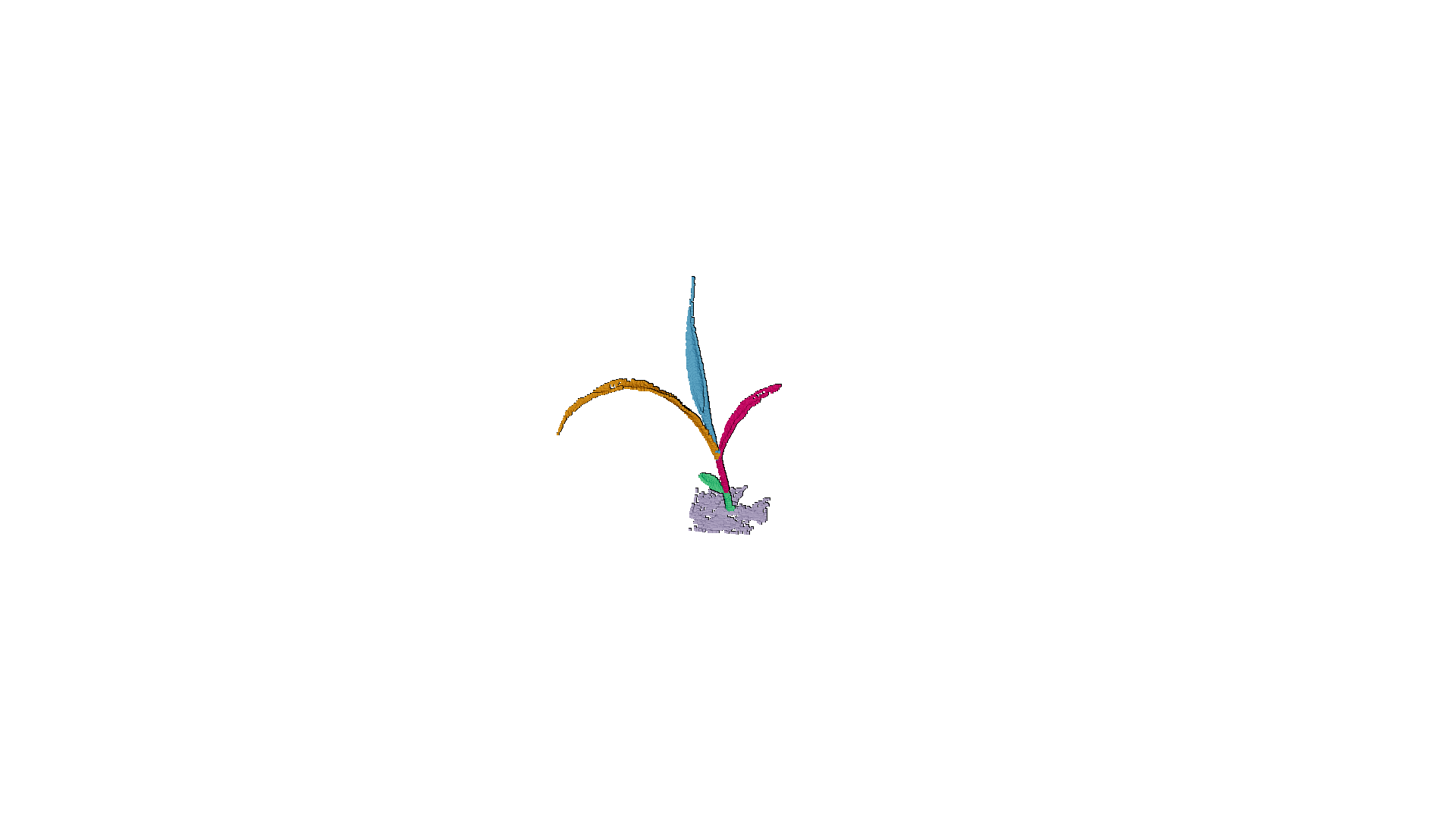}}
		\settoheight{\HC}{\includegraphics{\segImg}}
		\node[inner sep=0pt] (russell) at (7.2, 0.5) {\includegraphics[trim=.33\WC{} .33\HC{} .425\WC{} .3\HC{},clip, width=.25\textwidth]{\segImg}};
		\end{scope}
		
		\node (A) at (5.0,0) {};
		\node (B) at (5.9,0) {};
		\draw [->, gray] (A) -- (B);
		
		\node at(4.4, 0.1)[overlay, draw, black, dashed, line width=1pt, minimum width=1.0cm,circle]{}; 
		\node at(3.15, 0.35)[overlay, draw, black, dashed, line width=1pt, minimum width=1.0cm,circle]{}; 
		\node at(3.9, 1.6)[overlay, draw, black, dashed, line width=1pt, minimum width=1.0cm,circle]{}; 
		\node at(3.62, -1.22)[overlay, draw, black, dashed, line width=1pt, minimum width=1.0cm,circle]{}; 
		\node at(2.28, -0.88)[overlay, draw, black, dashed, line width=1pt, minimum width=1.0cm,circle]{}; 
		
		\end{tikzpicture}
		
		\caption{Instance segmentation: LatticeNet takes raw point clouds as input and embeds them into a sparse lattice where convolutions are applied. Features on the lattice are projected onto a 2D space where clustering is performed. The clusters define the instances of each object type in the original cloud.}
		\label{fig:instance_seg_pipeline}
	\end{figure*}

	A visualization of the pipeline for instance segmentation can be seen in ~\reffig{fig:instance_seg_pipeline}.

\subsection{Motion Segmentation} 

	Motion segmentation distinguishes between dynamic and static objects within a point cloud. For this, the network needs temporal information. We extend the original LatticeNet U-Net architecture with a recursive architecture that can process a sequence of point clouds $P_{seq}$ at times $t, t-1,\ldots, t-n$ and learn to distinguish for example between a moving car and a parked car. 
	
	The dynamic objects are considered as additional classes. Hence, we use the same loss as in the case of semantic segmentation. We also explore multiple ways to perform the fusion of temporal information which we detail in the~\nameref{section:arch} section.
	
\section{Network Architecture} \label{section:arch}
	Input to our network is a point cloud $P$ which may contain per-point features stored in $\mathbf{F}$. The output is class probabilities for each point $p$. In the recurrent network the input is an ordered set of point clouds $P_{seq}$ and the output are class probabilities for the last point cloud of the sequence. Moving and static objects are considered as different semantic classes.
	
	Our network architecture has a U-Net structure~\cite{ronneberger2015u} and is visualized in~\reffig{fig:architecture} together with the used individual blocks.
	
	The first layers distribute the point features onto the lattice and use a PointNet to obtain local features. Afterwards, a series of ResNet blocks~\cite{he2016deep}, followed by repeated downsampling, aggregates global context. The decoder branch mirrors the encoder architecture and upsamples through transposed convolutions. Finally, a DeformSlicing propagates lattice features onto the original point cloud. Skip connections are added by concatenating the encoder feature maps with matching decoder features.

\subsection{Temporal Fusion}

Incorporating temporal information for motion prediction over a sequence of point clouds relies on fusing information between multiple time-steps. For this purpose, the feature vectors of the timesteps $t-1$ and $t$ are passed through a Temporal Fusion block, as shown in \reffig{fig:rnn__architecture}. This fusion consists of a concatenation of both feature vectors and a linear layer followed by a non-linearity (\reffig{fig:temporal_fuse}).
Each new time-step allocates additional vertices in the lattice corresponding to newly explored areas in the map. For correct fusion, the features from the previous time-step need to be zero-padded so that the sizes match.

Additionally, we performed experiments with a single Temporal Fusion block in the network and max-pooling over both feature vectors instead of the linear layer, but found that three Temporal Fusion blocks achieved overall superior results.

It should be noted that our approach for temporal fusion relies on a sequence of clouds that are transformed into a common coordinate frame. The required scan poses for transformation can be obtained e.g. from GPS or SLAM.
	
	\bgroup
	\def\W{60pt}
	\def\H{10pt}
	\def\Sep{25pt}
	\def\ShiftLevel{20pt}
	\begin{figure}[]
		
		\begin{tikzpicture}
		
		\node[] at (4.5+0.19,-0.7) {
			\begin{tikzpicture}[scale=0.8, every node/.style={scale=0.8},>={Stealth[inset=3pt,length=6pt,angle'=28,round]} ]
			\node[rectangle, inner sep=0pt, minimum width=\W, minimum height=\H] (input) at (0,0) {};

			\node[rectangle,thin,dashed, inner sep=0pt, minimum width=\W, minimum height=\H] (gn-1) at (0,-\Sep*1) {\footnotesize GN};
			\node[rectangle,thin,dashed, inner sep=0pt, minimum width=\W, minimum height=\H] (relu-1) at (0,-\Sep*1-\H) {\footnotesize ReLU};
			\node[rectangle,thin,dashed, inner sep=0pt, minimum width=\W, minimum height=\H] (1x1-1) at (0,-\Sep*1-\H*2) {\footnotesize 1x1, $16,32,64$};
			\draw[] ($(gn-1.north west)$)  rectangle ($(1x1-1.south east)$) node[pos=0.0, right, xshift=\W, yshift=-\H*3/2] {$\times \textbf{3}$};
			\draw[thin,dashed] ($(gn-1.south west)$)  -- ($(gn-1.south east)$);
			\draw[thin,dashed] ($(relu-1.south west)$)  -- ($(relu-1.south east)$);
			
			\node[draw,rectangle, inner sep=0pt, minimum width=\W, minimum height=\H] (scatter-max) at (0,-\Sep*2-\H*2) {MaxPool};
			
			\draw [->]  (input) -> (gn-1) node[pos=0.5, right] {$64$-d};
			\draw [->]  (1x1-1) -> (scatter-max) node[pos=0.5, right] {};
			\draw [->]  (scatter-max.south) -> ($(scatter-max.south)-(0,12pt)$) node[pos=0.5, right] {};
			
			\node (circle) [draw, very thin, dashed, overlay, circle, minimum size = 36mm] at (0,-1.45) {};
			\end{tikzpicture}
			
		};

		\node[] at (4.5,-3.7) {
			\begin{tikzpicture}[scale=0.8, every node/.style={scale=0.8},>={Stealth[inset=3pt,length=6pt,angle'=28,round]} ]
			\node[rectangle, inner sep=0pt, minimum width=\W, minimum height=\H] (input) at (0,0) {};

			\node[rectangle,thin,dashed, inner sep=0pt, minimum width=\W, minimum height=\H] (gn-1) at (0,-\Sep*1) {\footnotesize GN};
			\node[rectangle,thin,dashed, inner sep=0pt, minimum width=\W, minimum height=\H] (relu-1) at (0,-\Sep*1-\H) {\footnotesize ReLU};
			\node[rectangle,thin,dashed, inner sep=0pt, minimum width=\W, minimum height=\H] (conv-1) at (0,-\Sep*1-\H*2) {\footnotesize conv, $256$};
			\draw[] ($(gn-1.north west)$)  rectangle ($(conv-1.south east)$) node[pos=0.0, right, xshift=\W, yshift=-\H*3/2] {$\times \textbf{2}$};
			\draw[thin,dashed] ($(gn-1.south west)$)  -- ($(gn-1.south east)$);
			\draw[thin,dashed] ($(relu-1.south west)$)  -- ($(relu-1.south east)$);
			
			\node[draw,circle, inner sep=0pt, minimum size=10pt] (add) at ($(conv-1.south)-(0,10pt)$) {+};
			
			\draw [->]  (input) -> (gn-1) node[pos=0.5, right] {$256$-d} node[midway,xshift=3.5pt] (first-line) {};
			\draw []  (conv-1.south) -- (add) {};
			\draw [->]  (add.south) -> ($(add.south)-(0,10pt)$) {};
			\draw [->] (first-line) to [out=180,in=90] ($(relu-1)-(\W/2+17pt,0)$) to [out=270, in=180 ] (add);
			
			\node (circle) [draw, very thin, dashed, overlay, circle, minimum size = 37mm] at (relu-1) {};
			\end{tikzpicture}
	
		};
	
		\node[] at (4.5,-7.1) {
			\begin{tikzpicture}[scale=0.8, every node/.style={scale=0.8},>={Stealth[inset=3pt,length=6pt,angle'=28,round]} ]
			\node[rectangle, inner sep=0pt, minimum width=\W, minimum height=\H] (input) at (0,0) {};
			
			\node[draw,rectangle, inner sep=0pt, minimum width=\W, minimum height=\H] (gather) at (0,-\Sep*1-10pt) {\footnotesize Gather};
			
			\node[rectangle,thin,dashed, inner sep=0pt, minimum width=\W, minimum height=\H] (gn-1) at (0,-\Sep*2-10pt) {\footnotesize MaxPool};
			\node[rectangle,thin,dashed, inner sep=0pt, minimum width=\W, minimum height=\H] (relu-1) at (0,-\Sep*2-\H-10pt) {\footnotesize SubtractMax};
			\node[rectangle,thin,dashed, inner sep=0pt, minimum width=\W, minimum height=\H] (1x1-1) at (0,-\Sep*2-\H*2-10pt) {\footnotesize Linear+Tanh};
			\draw[] ($(gn-1.north west)$)  rectangle ($(1x1-1.south east)$);
			\draw[thin,dashed] ($(gn-1.south west)$)  -- ($(gn-1.south east)$);
			\draw[thin,dashed] ($(relu-1.south west)$)  -- ($(relu-1.south east)$);
			
			\node[draw,rectangle, inner sep=0pt, minimum width=\W, minimum height=\H] (slice) at (0,-\Sep*3-\H*2-10pt) {slice};
			
			\draw [->,transform canvas={shift={(-10pt ,0)}}]  (input) -> (gather) node[pos=0.2, right] {P} node[pos=0.7,xshift=-6.5pt] (p-line) {};
			\draw [->,transform canvas={shift={(10pt ,0)}}]  (input) -> (gather) node[pos=0.2, right] {$64$-d} node[pos=0.7,xshift=6.5pt] (v-line) {};
			\draw [->]  (gather) -> (gn-1) node[pos=0.5, right] {$64\times p_d$-d};
			\draw [->]  (1x1-1) -> (slice) node[pos=0.5, right] {} node[pos=0.5, right] {delta};
			\draw [->]  (slice.south) -> ($(slice.south)-(0,12pt)$) node[pos=0.5, right] {};
			
			\draw[->] (p-line) --  ++(-1.5,0) |- (slice);
			\draw[->] (v-line) --  ++(+1.5,0) |- (slice);
			
			\node (circle) [draw, very thin, dashed, overlay, circle, minimum size = 48mm] at (0,-2.2) {};
			\end{tikzpicture}
		};
	
		\node[] at (0,-4) {
			\begin{tikzpicture}[>={Stealth[inset=2pt,length=4.5pt,angle'=28,round]}, scale=0.95, every node/.style={scale=0.95}]

			\scriptsize
			
			\def\W{60pt}
			\def\H{11pt}
			\def\Sep{15pt}
			\def\SepLvl{3pt}
			\def\SepUpLvl{5pt}
			\def\ShiftLevel{13pt}
			\def\BrightnessOrange{60}
			\def\BrightnessBlue{60}
			\def\BrightnessGreen{60}
			\def\Brightnessellow{60}
			\colorlet{inputColor}{ph-orange!80}
			\colorlet{distributeColor}{ph-blue!70}
			\colorlet{level1DownColor}{ph-blue!70}
			\colorlet{level2DownColor}{ph-green!70}
			\colorlet{bottleneckColor}{ph-yellow!80}
			\colorlet{level2UpColor}{ph-green!80}
			\colorlet{level1UpColor}{ph-blue!70}
			\colorlet{sliceColor}{ph-blue!70}
			\colorlet{linearColor}{ph-orange!80}
			\colorlet{concatColor}{ph-orange!80}
			\def\FSize{\footnotesize}
			\contourlength{0.05em} 
			\contournumber{20}  
			
			\node[rectangle,fill=inputColor, text=white, inner sep=0pt, minimum width=\W, minimum height=\H] (input) at (0,+5pt) {\FSize Input};
			\node[rectangle,fill=level1DownColor, text=white, inner sep=0pt, minimum width=\W, minimum height=\H] (distribute) at (0,-\Sep) {\FSize Distribute};
			\node[rectangle,fill=level1DownColor, text=white, inner sep=0pt, minimum width=\W, minimum height=\H] (pointnet) at (0,-\Sep*2) {\FSize PointNet};
			\node[rectangle,fill=level1DownColor, text=white, inner sep=0pt, minimum width=\W, minimum height=\H] (d1c1) at (0,-\Sep*3) {\FSize ResNet Block};
			\node[rectangle,fill=level2DownColor, text=white, inner sep=0pt, minimum width=\W, minimum height=\H] (downsample1) at (0+\ShiftLevel,-\Sep*4-\SepLvl) {\FSize Downsample};
			\node[rectangle,fill=level2DownColor, text=white, inner sep=0pt, minimum width=\W, minimum height=\H] (d2c1) at (0+\ShiftLevel,-\Sep*5-\SepLvl) {\FSize ResNet Block};
			\node[rectangle,fill=level2DownColor, text=white, inner sep=0pt, minimum width=\W, minimum height=\H] (d2c2) at (0+\ShiftLevel,-\Sep*6-\SepLvl) {\FSize ResNet Block};
			\node[rectangle,fill=bottleneckColor, text=white, inner sep=0pt, minimum width=\W, minimum height=\H] (downsample2) at (0+\ShiftLevel*2,-\Sep*7-\SepLvl*2) {\FSize Downsample};
			\node[rectangle,fill=bottleneckColor, text=white, inner sep=0pt, minimum width=\W, minimum height=\H] (bottleneck1) at (0+\ShiftLevel*2,-\Sep*8-\SepLvl*2) {\FSize ResNet Block};
			\node[rectangle,fill=bottleneckColor, text=white, inner sep=0pt, minimum width=\W, minimum height=\H] (bottleneck2) at (0+\ShiftLevel*2,-\Sep*9-\SepLvl*2) {\FSize ResNet Block};
			\node[rectangle,fill=bottleneckColor, text=white, inner sep=0pt, minimum width=\W, minimum height=\H] (upsample1) at (0+\ShiftLevel*2,-\Sep*10-\SepLvl*2) {\FSize Upsample};
			\node[rectangle,fill=level2UpColor, text=white, inner sep=0pt, minimum width=\W, minimum height=\H] (u1c1) at (0+\ShiftLevel,-\Sep*11-\SepLvl*3-\SepUpLvl) {\FSize ResNet Block};
			\node[rectangle,fill=level2UpColor, text=white, inner sep=0pt, minimum width=\W, minimum height=\H] (u1c2) at (0+\ShiftLevel,-\Sep*12-\SepLvl*3-\SepUpLvl) {\FSize ResNet Block};
			\node[rectangle,fill=level2UpColor, text=white, inner sep=0pt, minimum width=\W, minimum height=\H] (upsample2) at (0+\ShiftLevel,-\Sep*13-\SepLvl*3-\SepUpLvl) {\FSize Upsample};
			\node[rectangle,fill=level1UpColor, text=white, inner sep=0pt, minimum width=\W, minimum height=\H] (u2c1) at (0,-\Sep*14-\SepLvl*4-\SepUpLvl*2) {\FSize ResNet Block};
			\node[rectangle,fill=sliceColor, text=white, inner sep=0pt, minimum width=\W, minimum height=\H] (deform-slice) at (0,-\Sep*15-\SepLvl*4-\SepUpLvl*2) {\FSize DeformSlice};
			\node[rectangle,fill=linearColor, text=white, inner sep=0pt, minimum width=\W, minimum height=\H] (linear) at (0,-\Sep*16-\SepLvl*4-\SepUpLvl*2) {\FSize Linear};
			
			\def\ShiftXConcat{-\W/2 + \ShiftLevel/2} 
			\def\ShiftYConcat{\Sep/2-\H/2+\SepLvl/2+\SepUpLvl/2} 
			\path [draw=black,postaction={on each segment={custom arrow2=black}}, transform canvas={shift={(\ShiftXConcat ,0)}}] (d1c1) to (u2c1);
			\path [draw=black,postaction={on each segment={mid arrow=black}}, transform canvas={shift={(\ShiftXConcat ,0)}}] (d2c2) to (u1c1);
			
			\node[circle,fill=concatColor, text=white, inner sep=0pt, minimum size=10pt] (concat2) at ([xshift=\ShiftXConcat,yshift=\ShiftYConcat]u2c1.north) {\footnotesize $\bm{\mathsf{C}}$};
			\node[circle,fill=concatColor, text=white, inner sep=0pt, minimum size=10pt] (concat1) at ([xshift=\ShiftXConcat,yshift=\ShiftYConcat]u1c1.north) {\footnotesize $\bm{\mathsf{C}}$ };
			
			\draw[->] (upsample2) |- (concat2); 
			\draw[->] (upsample1) |- (concat1);
			
			\draw[->] ($(input.south) +(-10pt,0pt) $) --  ($(distribute.north) +(-10pt,0pt) $) node[midway,xshift=2.2pt] (p-middle) {} node[pos=0.5, right] {P} ;
			\draw[->] ($(input.south) +(10pt,0pt) $) -- ($(distribute.north) +(10pt,0pt) $) node[pos=0.5, right] {V};
			\draw[->] (distribute) -- (pointnet);
			\draw[->] (pointnet) -- (d1c1); 
			\draw[->] (downsample1) -- (d2c1);
			\draw[->] (d2c1) -- (d2c2);
			\draw[->] (downsample2) -- (bottleneck1); 
			\draw[->] (bottleneck1) -- (bottleneck2);
			\draw[->] (bottleneck2) -- (upsample1); 
			\draw[->] (u1c1) -- (u1c2);
			\draw[->] (u1c2) -- (upsample2);
			\draw[->] (u2c1) -- (deform-slice);
			\draw[->] (deform-slice) -- (linear);
			
			\path[draw=black,postaction={on each segment={custom arrow={black} }}
			] (p-middle) --  ++(-1.3,0) |- (deform-slice);
			
			\draw[->] ($ (d1c1.south) + (\ShiftLevel,0) $) -- (downsample1); 
			\draw[->] ($ (d2c2.south) + (\ShiftLevel,0) $) -- (downsample2); 
			
			\draw[dashed, overlay, very thin]  (pointnet.north east) -- (4.25, 0.35); 
			\draw[dashed, overlay, very thin]  (pointnet.south east) -- (4.35, -2.46);
			\draw[dashed, overlay, very thin]  (bottleneck1.north east) -- (4.1, -2.85); 
			\draw[dashed, overlay, very thin]  (bottleneck1.south east) -- (4.1, -5.55);
			\draw[dashed, overlay, very thin]  (deform-slice.north east) -- (3.6, -6.4); 
			\draw[dashed, overlay, very thin]  (deform-slice.south east) -- (4.25, -9.81);
					
			\end{tikzpicture}
		};
		\end{tikzpicture}
		
		\caption{Architecture: Our model follows a U-Net structure. For ease of representation, blocks which are repeated one after another are indicated with a multiplier on the right side of the operation.} \label{fig:architecture}
	\end{figure}
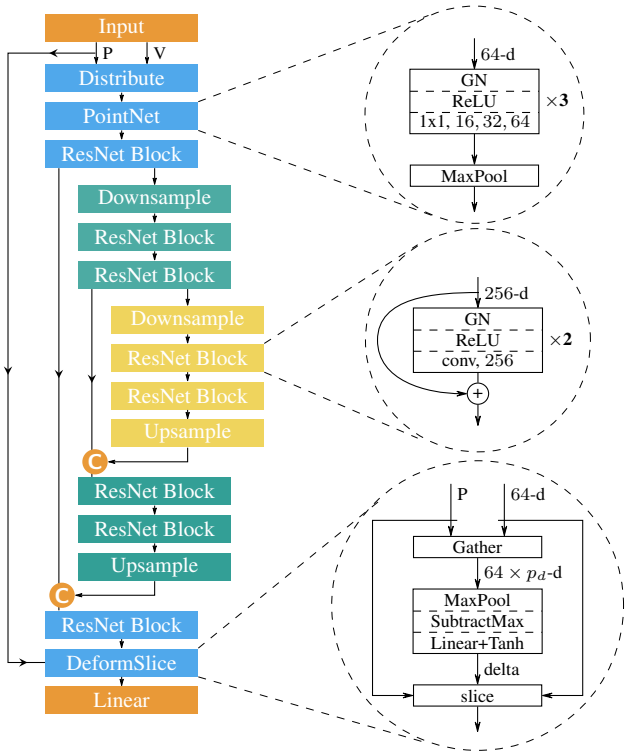
	\egroup

	
	\bgroup
	\def\W{60pt}
	\def\H{10pt}
	\def\Sep{25pt}
	\def\ShiftLevel{20pt}
	\begin{figure}[]
		
		\hspace*{0.07\linewidth}
		\begin{tikzpicture}

		
		\node[] at (4.5,-4) {
			\begin{tikzpicture}[>={Stealth[inset=2pt,length=4.5pt,angle'=28,round]}, scale=0.95, every node/.style={scale=0.95}]

			\scriptsize
			
			\def\W{60pt}
			\def\H{11pt}
			\def\Sep{15pt}
			\def\SepLvl{3pt}
			\def\SepUpLvl{5pt}
			\def\ShiftLevel{13pt}
			\def\BrightnessOrange{60}
			\def\BrightnessBlue{60}
			\def\BrightnessGreen{60}
			\def\Brightnessellow{60}
			\colorlet{inputColor}{ph-orange!80}
			\colorlet{distributeColor}{ph-blue!70}
			\colorlet{level1DownColor}{ph-blue!70}
			\colorlet{level2DownColor}{ph-green!70}
			\colorlet{bottleneckColor}{ph-yellow!80}
			\colorlet{level2UpColor}{ph-green!80}
			\colorlet{level1UpColor}{ph-blue!70}
			\colorlet{sliceColor}{ph-blue!70}
			\colorlet{linearColor}{ph-orange!80}
			\colorlet{concatColor}{ph-orange!80}
			\colorlet{temporalFusionColor}{ph-purple}
			\def\FSize{\footnotesize}
			\contourlength{0.05em} 
			\contournumber{20}  
			
			\node[rectangle,fill=inputColor, text=white, inner sep=0pt, minimum width=\W, minimum height=\H] (input) at (0,+5pt) {\FSize Input t-1};
			\node[rectangle,fill=level1DownColor, text=white, inner sep=0pt, minimum width=\W, minimum height=\H] (distribute) at (0,-\Sep) {\FSize Distribute};
			\node[rectangle,fill=level1DownColor, text=white, inner sep=0pt, minimum width=\W, minimum height=\H] (pointnet) at (0,-\Sep*2) {\FSize PointNet};
			\node[rectangle,fill=white, text=white, inner sep=0pt, minimum width=\W, minimum height=\H] (temporalFusionEarly) at (0,-\Sep*3) {\FSize Temporal Fusion};
			
			\node[rectangle,fill=level1DownColor, text=white, inner sep=0pt, minimum width=\W, minimum height=\H] (d1c1) at (0,-\Sep*4) {\FSize ResNet Block};
			\node[rectangle,fill=level1DownColor, text=white, inner sep=0pt, minimum width=\W, minimum height=\H] (d1c2) at (0,-\Sep*5) {\FSize ResNet Block};
			\node[rectangle,fill=white, text=white, inner sep=0pt, minimum width=\W, minimum height=\H] (temporalFusionMiddle) at (0,-\Sep*6) {\FSize Temporal Fusion};
			
			\node[rectangle,fill=level2DownColor, text=white, inner sep=0pt, minimum width=\W, minimum height=\H] (downsample1) at (0+\ShiftLevel,-\Sep*7-\SepLvl) {\FSize Downsample};
			\node[rectangle,fill=level2DownColor, text=white, inner sep=0pt, minimum width=\W, minimum height=\H] (d2c1) at (0+\ShiftLevel,-\Sep*8-\SepLvl) {\FSize ResNet Block};
			
			\node[rectangle,fill=bottleneckColor, text=white, inner sep=0pt, minimum width=\W, minimum height=\H] (downsample2) at (0+\ShiftLevel*2,-\Sep*9-\SepLvl*2) {\FSize Downsample};
			\node[rectangle,fill=bottleneckColor, text=white, inner sep=0pt, minimum width=\W, minimum height=\H] (bottleneck1) at (0+\ShiftLevel*2,-\Sep*10-\SepLvl*2) {\FSize ResNet Block};

			\node[rectangle,fill=bottleneckColor, text=white, inner sep=0pt, minimum width=\W, minimum height=\H] (upsample1) at (0+\ShiftLevel*2,-\Sep*11-\SepLvl*2) {\FSize Upsample};
			\node[rectangle,fill=level2UpColor, text=white, inner sep=0pt, minimum width=\W, minimum height=\H] (u1c1) at (0+\ShiftLevel,-\Sep*12-\SepLvl*3-\SepUpLvl) {\FSize ResNet Block};
			
			\node[rectangle,fill=level2UpColor, text=white, inner sep=0pt, minimum width=\W, minimum height=\H] (upsample2) at (0+\ShiftLevel,-\Sep*13-\SepLvl*3-\SepUpLvl) {\FSize Upsample};
			
			\node[rectangle,fill=white, text=white, inner sep=0pt, minimum width=\W, minimum height=\H] (temporalFusionLate) at (0,-\Sep*14 -\SepLvl*4-\SepUpLvl*2) {\FSize Temporal Fusion};
			
			\node[rectangle,fill=white, text=white, inner sep=0pt, minimum width=\W, minimum height=\H] (u2c1) at (0,-\Sep*15-\SepLvl*4-\SepUpLvl*2) {\FSize ResNet Block};
			\node[rectangle,fill=white, text=white, inner sep=0pt, minimum width=\W, minimum height=\H] (deform-slice) at (0,-\Sep*16-\SepLvl*4-\SepUpLvl*2) {\FSize DeformSlice};
			\node[rectangle,fill=white, text=white, inner sep=0pt, minimum width=\W, minimum height=\H] (linear) at (0,-\Sep*17-\SepLvl*4-\SepUpLvl*2) {\FSize Linear};
			
			\def\ShiftXConcat{-\W/2 + \ShiftLevel/2} 
			\def\ShiftYConcat{\Sep/2-\H/2+\SepLvl/2+\SepUpLvl/2} 
			\path [draw=black,postaction={on each segment={custom arrow2=black}}, transform canvas={shift={(\ShiftXConcat ,0)}}] (d1c2) to ($ (temporalFusionLate) + (0,0.25) $);
			\path [draw=black,postaction={on each segment={mid arrow=black}}, transform canvas={shift={(\ShiftXConcat ,0)}}] (d2c1) to (u1c1);
			
			\node[circle,fill=concatColor, text=white, inner sep=0pt, minimum size=10pt] (concat2) at ([xshift=\ShiftXConcat,yshift=\ShiftYConcat]temporalFusionLate.north) {\footnotesize $\bm{\mathsf{C}}$};
			\node[circle,fill=concatColor, text=white, inner sep=0pt, minimum size=10pt] (concat1) at ([xshift=\ShiftXConcat,yshift=\ShiftYConcat]u1c1.north) {\footnotesize $\bm{\mathsf{C}}$ };
			
			\draw[->] (upsample2) |- (concat2); 
			\draw[->] (upsample1) |- (concat1);
			
			\draw[->] ($(input.south) +(-10pt,0pt) $) --  ($(distribute.north) +(-10pt,0pt) $) node[midway,xshift=2.2pt] (p-middle) {} node[pos=0.5, right] {P} ;
			\draw[->] ($(input.south) +(10pt,0pt) $) -- ($(distribute.north) +(10pt,0pt) $) node[pos=0.5, right] {V};
			\draw[->] (distribute) -- (pointnet);
			\draw[->] (pointnet) -- (d1c1);
			\draw[->] (d1c1) -- (d1c2);
			\draw[->] (downsample1) -- (d2c1);
			\draw[->] (downsample2) -- (bottleneck1); 
			\draw[->] (bottleneck1) -- (upsample1); 
			\draw[->] (u1c1) -- (upsample2);
			
			
			\draw[->] ($ (d1c2.south) + (\ShiftLevel,0) $) -- (downsample1); 
			\draw[->] ($ (d2c1.south) + (\ShiftLevel,0) $) -- (downsample2);

			\draw[overlay,->, dashed, thick, -{Stealth[scale=1.0]} ] ($ (pointnet.south) + (0,-0.3) $) -- ($ (pointnet.south) + (-2.9,-0.3) $);
			\draw[overlay,->, dashed, thick, -{Stealth[scale=1.0]} ] ($ (d1c2.south) + (\ShiftXConcat-1.0,-0.3) $) -- ($ (d1c2.south) + (-2.9,-0.3) $); 
			\draw[overlay,->, dashed, thick, -{Stealth[scale=1.0]} ] ($ (concat2.south) + (0.0,-0.0) $) |- ($ (concat2.south) + (-2.1,-0.22) $); 
			
			\end{tikzpicture}
		};
		
		\node[] at (0.5,-4) {
			\begin{tikzpicture}[>={Stealth[inset=2pt,length=4.5pt,angle'=28,round]}, scale=0.95, every node/.style={scale=0.95}]

			\scriptsize
			
			\def\W{60pt}
			\def\H{11pt}
			\def\Sep{15pt}
			\def\SepLvl{3pt}
			\def\SepUpLvl{5pt}
			\def\ShiftLevel{13pt}
			\def\BrightnessOrange{60}
			\def\BrightnessBlue{60}
			\def\BrightnessGreen{60}
			\def\Brightnessellow{60}
			\colorlet{inputColor}{ph-orange!80}
			\colorlet{distributeColor}{ph-blue!70}
			\colorlet{level1DownColor}{ph-blue!70}
			\colorlet{level2DownColor}{ph-green!70}
			\colorlet{bottleneckColor}{ph-yellow!80}
			\colorlet{level2UpColor}{ph-green!80}
			\colorlet{level1UpColor}{ph-blue!70}
			\colorlet{sliceColor}{ph-blue!70}
			\colorlet{linearColor}{ph-orange!80}
			\colorlet{concatColor}{ph-orange!80}
			\colorlet{temporalFusionColor}{ph-purple}
			\def\FSize{\footnotesize}
			\contourlength{0.05em} 
			\contournumber{20}  
			
			\node[rectangle,fill=inputColor, text=white, inner sep=0pt, minimum width=\W, minimum height=\H] (input) at (0,+5pt) {\FSize Input t};
			\node[rectangle,fill=level1DownColor, text=white, inner sep=0pt, minimum width=\W, minimum height=\H] (distribute) at (0,-\Sep) {\FSize Distribute};
			\node[rectangle,fill=level1DownColor, text=white, inner sep=0pt, minimum width=\W, minimum height=\H] (pointnet) at (0,-\Sep*2) {\FSize PointNet};
			\node[rectangle,fill=temporalFusionColor, text=white, inner sep=0pt, minimum width=\W, minimum height=\H] (temporalFusionEarly) at (0,-\Sep*3) {\FSize Temporal Fusion};
			
			\node[rectangle,fill=level1DownColor, text=white, inner sep=0pt, minimum width=\W, minimum height=\H] (d1c1) at (0,-\Sep*4) {\FSize ResNet Block};
			\node[rectangle,fill=level1DownColor, text=white, inner sep=0pt, minimum width=\W, minimum height=\H] (d1c2) at (0,-\Sep*5) {\FSize ResNet Block};
			\node[rectangle,fill=temporalFusionColor, text=white, inner sep=0pt, minimum width=\W, minimum height=\H] (temporalFusionMiddle) at (0,-\Sep*6) {\FSize Temporal Fusion};
			
			\node[rectangle,fill=level2DownColor, text=white, inner sep=0pt, minimum width=\W, minimum height=\H] (downsample1) at (0+\ShiftLevel,-\Sep*7-\SepLvl) {\FSize Downsample};
			\node[rectangle,fill=level2DownColor, text=white, inner sep=0pt, minimum width=\W, minimum height=\H] (d2c1) at (0+\ShiftLevel,-\Sep*8-\SepLvl) {\FSize ResNet Block};
			
			\node[rectangle,fill=bottleneckColor, text=white, inner sep=0pt, minimum width=\W, minimum height=\H] (downsample2) at (0+\ShiftLevel*2,-\Sep*9-\SepLvl*2) {\FSize Downsample};
			\node[rectangle,fill=bottleneckColor, text=white, inner sep=0pt, minimum width=\W, minimum height=\H] (bottleneck1) at (0+\ShiftLevel*2,-\Sep*10-\SepLvl*2) {\FSize ResNet Block};

			\node[rectangle,fill=bottleneckColor, text=white, inner sep=0pt, minimum width=\W, minimum height=\H] (upsample1) at (0+\ShiftLevel*2,-\Sep*11-\SepLvl*2) {\FSize Upsample};
			\node[rectangle,fill=level2UpColor, text=white, inner sep=0pt, minimum width=\W, minimum height=\H] (u1c1) at (0+\ShiftLevel,-\Sep*12-\SepLvl*3-\SepUpLvl) {\FSize ResNet Block};

			\node[rectangle,fill=level2UpColor, text=white, inner sep=0pt, minimum width=\W, minimum height=\H] (upsample2) at (0+\ShiftLevel,-\Sep*13-\SepLvl*3-\SepUpLvl) {\FSize Upsample};
			
			\node[rectangle,fill=temporalFusionColor, text=white, inner sep=0pt, minimum width=\W, minimum height=\H] (temporalFusionLate) at (0,-\Sep*14 -\SepLvl*4-\SepUpLvl*2) {\FSize Temporal Fusion};
			\node[rectangle,fill=level1UpColor, text=white, inner sep=0pt, minimum width=\W, minimum height=\H] (u2c1) at (0,-\Sep*15-\SepLvl*4-\SepUpLvl*2) {\FSize ResNet Block};
			
			\node[rectangle,fill=sliceColor, text=white, inner sep=0pt, minimum width=\W, minimum height=\H] (deform-slice) at (0,-\Sep*16-\SepLvl*4-\SepUpLvl*2) {\FSize DeformSlice};
			\node[rectangle,fill=linearColor, text=white, inner sep=0pt, minimum width=\W, minimum height=\H] (linear) at (0,-\Sep*17-\SepLvl*4-\SepUpLvl*2) {\FSize Linear};
			
			\def\ShiftXConcat{-\W/2 + \ShiftLevel/2} 
			\def\ShiftYConcat{\Sep/2-\H/2+\SepLvl/2+\SepUpLvl/2} 
			\path [draw=black,postaction={on each segment={custom arrow2=black}}, transform canvas={shift={(\ShiftXConcat ,0)}}] (temporalFusionMiddle) to (temporalFusionLate);
			\path [draw=black,postaction={on each segment={mid arrow=black}}, transform canvas={shift={(\ShiftXConcat ,0)}}] (d2c1) to (u1c1);
			
			\node[circle,fill=concatColor, text=white, inner sep=0pt, minimum size=10pt] (concat2) at ([xshift=\ShiftXConcat,yshift=\ShiftYConcat]temporalFusionLate.north) {\footnotesize $\bm{\mathsf{C}}$};
			\node[circle,fill=concatColor, text=white, inner sep=0pt, minimum size=10pt] (concat1) at ([xshift=\ShiftXConcat,yshift=\ShiftYConcat]u1c1.north) {\footnotesize $\bm{\mathsf{C}}$ };
			
			\draw[->] (upsample2) |- (concat2); 
			\draw[->] (upsample1) |- (concat1);
			
			\draw[->] ($(input.south) +(-10pt,0pt) $) --  ($(distribute.north) +(-10pt,0pt) $) node[midway,xshift=2.2pt] (p-middle) {} node[pos=0.5, right] {P} ;
			\draw[->] ($(input.south) +(10pt,0pt) $) -- ($(distribute.north) +(10pt,0pt) $) node[pos=0.5, right] {V};
			\draw[->] (distribute) -- (pointnet);
			\draw[->] (pointnet) -- (temporalFusionEarly); 
			\draw[->] (temporalFusionEarly) -- (d1c1);
			\draw[->] (d1c1) -- (d1c2);
			\draw[->] (d1c2) -- (temporalFusionMiddle);
			\draw[->] (downsample1) -- (d2c1);
			\draw[->] (downsample2) -- (bottleneck1); 
			\draw[->] (bottleneck1) -- (upsample1); 
			\draw[->] (u1c1) -- (upsample2);
			\draw[->] (temporalFusionLate) -- (u2c1);
			\draw[->] (u2c1) -- (deform-slice);
			\draw[->] (deform-slice) -- (linear);
			
			\path[draw=black,postaction={on each segment={custom arrow={black} }}
			] (p-middle) --  ++(-1.3,0) |- (deform-slice);
			
			\draw[->] ($ (temporalFusionMiddle.south) + (\ShiftLevel,0) $) -- (downsample1); 
			\draw[->] ($ (d2c1.south) + (\ShiftLevel,0) $) -- (downsample2); 
			
			
			\end{tikzpicture}
		};
	\end{tikzpicture}
	
	\caption{Recurrent architecture: The features from previous time-steps are fused in the current time-step at multiple levels of the network. This allows the network to distinguish dynamic objects from static ones. } \label{fig:rnn__architecture}
	\end{figure}
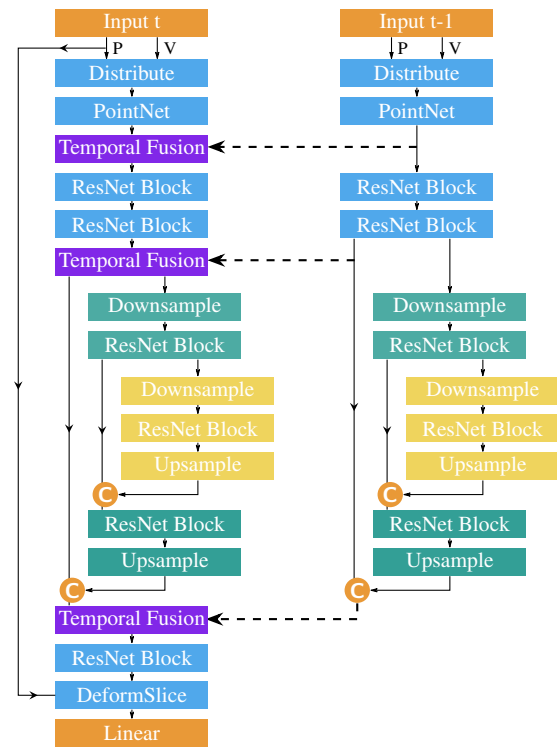
	\egroup

	
	\bgroup
	\begin{figure}[]
	
	\resizebox{0.48\textwidth}{!}{
	\begin{tikzpicture}
		\colorlet{featout}{ph-green!80}
		\colorlet{feat1}{ph-orange!80}
		\colorlet{feat2}{ph-blue!80}
		\colorlet{feat3}{ph-green!80}
		\colorlet{feat3}{ph-yellow}
		
		\newcolumntype{M}[1]{>{\centering\arraybackslash}m{#1}}
		\setlength\tabcolsep{0pt}
		\setlength\arrayrulewidth{1pt} 
		
		\node[] at (0,0) {
			\resizebox{1cm}{!}{ 
			\begin{tabular}{|@{\rule[-0.2cm]{0pt}{0.5cm}}*{2}{M{0.5cm} |}}
			\hline
			\cellcolor{featout} & \cellcolor{featout} \\ 
			\hline
			\cellcolor{featout} & \cellcolor{featout} \\ 
			\hline
			\cellcolor{featout} & \cellcolor{featout} \\ 
			\hline
			\cellcolor{featout} & \cellcolor{featout} \\
			\hline
			\cellcolor{featout} & \cellcolor{featout} \\ 
			\hline
			\cellcolor{featout} & \cellcolor{featout} \\  
			\hline
			\end{tabular}
			}
		};
	
		\node[] at (3.3,0) {
			\resizebox{1cm}{!}{ 
			\begin{tabular}{|@{\rule[-0.2cm]{0pt}{0.5cm}}*{2}{M{0.5cm} |}}
			\hline
			\cellcolor{feat1} & \cellcolor{feat1} \\ 
			\hline
			\cellcolor{feat1} & \cellcolor{feat1} \\ 
			\hline
			\cellcolor{feat1} & \cellcolor{feat1} \\ 
			\hline
			\cellcolor{feat1} & \cellcolor{feat1} \\
			\hline
			 0 & 0 \\ 
			\hline
			 0 & 0 \\ 
			\hline 
			\end{tabular}
			}
		}; 
	
		\node[] at (5.3,0) {
			\resizebox{1cm}{!}{  
			\begin{tabular}{|@{\rule[-0.2cm]{0pt}{0.5cm}}*{2}{M{0.5cm} |}}
			\hline
			\cellcolor{feat2} & \cellcolor{feat2} \\ 
			\hline
			0 & 0 \\ 
			\hline
			0 & 0 \\
			\hline
			\cellcolor{feat2} & \cellcolor{feat2} \\
			\hline
			\cellcolor{feat3} & \cellcolor{feat3} \\
			\hline
			\cellcolor{feat3} & \cellcolor{feat3} \\
			\hline  
			\end{tabular}
			}
		};
	
		\node[] at (0.0, -1.55) {
			$\mathbf{X}$
		};
		\node[] at (3.3, -1.55) {
			$\mathbf{X}_{t-1}$
		};
		\node[] at (5.3, -1.55) {
			$\mathbf{X}_{t}$
		};
	
		\node[] at (7.8, 1.5) {
			$\intercal$
		};
		\node[] at (8.1, 1.65) {
			$\intercal$
		};
	
		\node[overlay,anchor=west] at (0.4,0) {


			$= ReLU \left( W \left\{
			\rule{0cm}{1.5cm} \rule{5cm}{0cm} \right\}
			\rule{0cm}{1.6cm} \right) $ 
			


		};
	
		\node[overlay,anchor=west] at (3.7, -0.85) {
			\bigg\} pad
		};
		\node[overlay, anchor=west] at (5.7, -0.85) {
			\bigg\} new vertices
		};
		\node[overlay, text width=3cm, anchor=west] at (5.7, 0.45) {
			\bigg\} unoccupied 
		};

	\end{tikzpicture}
	}
	
	\caption{Temporal fusion: The features from the previous time-step are zero-padded in order to account for the new vertices that were allocated at the current time-step. The features are afterwards concatenated and passed through a linear layer followed by a non-linearity. } \label{fig:temporal_fuse}
	\end{figure}
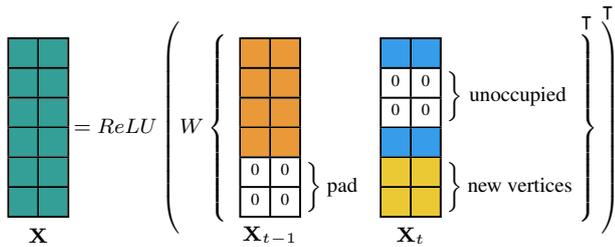
	\egroup

\section{Implementation}	
	Our lattice is stored sparsely on a hash map structure, which allows for fast access of neighboring vertices. Unlike \cite{su2018splatnet}, we construct the hash map directly on the GPU, saving us from incurring an expensive CPU to GPU memory copy.
	
	For memory savings, we implemented the DeformSlice and the last linear classification layer in one fused operation, avoiding the storage of high-dimensional feature vectors for each point in the point cloud.
	
	All of the lattice operators containing forwards and backwards passes are implemented on the GPU and exposed to PyTorch~\cite{paszke2017automatic}.
	
	Following recent works~\cite{he2016identity,huang2017densely}, all convolutions are pre-activated using Group Normalization~\cite{wu2018group} and a ReLU unit. We chose Group Normalization instead of the standard batch normalization due to greater stability for small batch sizes. We use the default of \num{32} groups.
	
	The models were trained using the Adam optimizer with a learning rate of $0.001$ and a weight decay of \num{e-4}. The learning rate was reduced by a factor of \num{10} when the loss plateaued.
	
	We share the PyTorch implementation of LatticeNet at \\ \url{https://github.com/AIS-Bonn/lattice_net}.
	
\section{Experiments}
	We evaluate our proposed lattice network on four different datasets: ShapeNet~\cite{yi2016scalable}, ScanNet~\cite{dai2017scannet}, SemanticKITTI~\cite{behley2019semantickitti} and Pheno4D~\cite{pheno4d}. For the task of semantic segmentation and motion segmentation we report the mean Intersection-over-Union~(mIoU). For the task of instance segmentation, we report the Symmetric Best Dice~(SBD)~\cite{de2017semantic}. SBD measures the accuracy of the instance segmentation by averaging for each input label the ground truth label yielding the maximum Dice score.
	
	We use a shallow model for ShapeNet and Pheno4D and a deeper model for ScanNet and SemanticKITTI as the datasets are larger. We augment all data using random mirroring and translations in space. For ScanNet, we also apply random color jitter. A video with additional footage of the experiments is available online~\footnote{\url{http://www.ais.uni-bonn.de/videos/RSS_2020_Rosu/}}. 
	
	\subsection{Evaluation of Segmentation Accuracy}
\noindent\textbf{ShapeNet part segmentation} is a subset of the ShapeNet dataset~\cite{yi2016scalable} which contains objects from \num{16} different categories each segmented into \num{2} - \num{6} parts. The dataset consists of points sampled from the surface of the objects, together with the ground truth label of the corresponding object part. The objects have an average of \num{2613} points.
We train and evaluate our network on each object individually. We use the official train/test splits as defined by the dataset containing a total of \num{12137} training objects and \num{2874} test objects. The results for our and five competing methods are gathered in \reftab{tab:shapenet} and visualized in~\reffig{fig:ShapeNetImages}.

We observe that for some classes, we obtain state-of-the-art performance and for other objects, the IoU is slightly lower than for other approaches. We ascribe this to the fact that training one fixed architecture size for each individual object is suboptimal as some objects like the "cap" have as few as \num{55} examples while others like the table have more than \num{5}K. This causes the network to be prone to overfitting on the easy object or underfitting on the difficult ones. A fair evaluation would require finding an architecture that performs well for all objects on average. However, due to various issues with mislabeled ground truths~\cite{su2018splatnet} we deem that experimentation with more architectures or with different regularization strengths for individual objects would overfit the dataset.

\noindent\textbf{ScanNet 3D segmentation}~\cite{dai2017scannet} consists of 3D reconstructions of real rooms. It contains $\approx 1500$ rooms segmented into \num{20} classes (bed, furniture, wall, etc.). The rooms have between \num{9}K and \num{537}K points --- on average \num{145}K. We segment an entire room at once without cropping.
We use the official train/test splits as defined by the dataset containing a total of \num{1201} training rooms and \num{100} test objects.
We obtain an IoU of 64.0 which is significantly higher than the most similar related work of SplatNet. It is to be noted that MinkowskiNet achieves a higher IoU but at the expense of an extremely high spatial resolution of \SI{2}{\centi\metre} per voxel. In contrast, our approach allocates lattice vertices so that each vertex covers approximately \num{30} points. On this dataset, this corresponds to a spatial extent of approximately \SI{10}{\centi\metre}.
	
\noindent\textbf{SemanticKITTI}~\cite{behley2019semantickitti} consists of semantically annotated LiDAR scans of real urban environments. The annotation covers a total of \num{19} classes for single scan evaluation and a total of \num{25} classes for multiple scan evaluation. Each scan contains between \num{82}K and \num{129}K points. We process each scan entirely without any cropping. We use the official train/validation splits as defined by the dataset. The test set is not publicly available and testing can only be done through the benchmark server.

The results for single scan are provided in~\reftab{tab:semanticKitti}. Our LatticeNet outperforms all other methods --- in case of the most similar SplatNet by more than a factor of two.
It is to be noted that DarkNet53Seg \cite{behley2019semantickitti}, DarkNet21Seg \cite{behley2019semantickitti} and SqueezeSegV2 \cite{wu2018squeezesegv2} are methods that operate on a 2D image by wrapping the LiDAR scans to 2D using spherical coordinates. In contrast, our method can operate on general point clouds, directly in 3D.

For motion segmentation we take as input three point clouds at consecutive time steps and output the segmentation for the final, most recent cloud. We overlap this time window so that every clouds gets to be segmented. For the first few clouds, the time window is reduced as there are no clouds from previous time-steps to give as input.
The results for the motion segmentation are provided in~\reftab{tab:results_multiscan}.

We observe that for motion segmentation we outperform other approaches except for KPConv\cite{thomas2019kpconv}, which has higher IoU. However, it is to be noted that KPconv cannot process a full point cloud at once due to memory constraints and rather processes sub-clouds centered around random spheres in the scene. The spheres are chosen randomly in the scene to ensure each point is tested multiple times by different sphere locations. Finally, a voting scheme gives the final prediction. In contrast, our approach can process a full point cloud without requiring neighborhood searching or partitioning in sub-clouds.

\noindent\textbf{Bonn Activity Maps}~\cite{tanke2019bonn} is a dataset for human tracking, activity recognition and anticipation of multiple persons. It contains annotations of persons, their trajectories and activities. The 3D reconstruction of the four kitchen scenarios is however of more interest to us. The environments are reconstructed as 3D colored meshes and have no ground truth semantic annotations. We trained our LatticeNet on the ScanNet dataset and evaluate it on the \num{4} kitchens in order to provide an annotation for each vertex of the mesh. The results are shown in \reffig{fig:SemPredBonnActivity}. We can observe that our network generalizes well to unseen datasets, recorded with different sensors and with different noise properties as the semantic segmentations look plausible and exhibit sharp borders between classes.

\noindent\textbf{Pheno4D}~\cite{pheno4d} is a spatio-temporal dataset of point clouds of maize and tomato plants with instance annotations of leaves. We use a shallow version of LatticeNet to compute per-point embeddings and cluster them using mean-shift to recover the instances. We compare with PointNet and PointNet++ as they are popular methods for computing per-point embeddings. Since the dataset contains 7 maize and 7 tomato plants, we train on the first 5 plants for each type and test on the remaining two. The results are gathered in~\reftab{tab:sbd}. We observe that our method is capable of computing more meaningful embeddings that create more distinctive clusters between each plant organ. 

\bgroup
\def\ElemWidth{1.7cm}
\def\Img{./imgs/predictions/activity_maps/gall_kitchen_rgb.png} 
\newlength{\WSA}
\newlength{\HSA}
\settowidth{\WSA}{\includegraphics{\Img}}
\settoheight{\HSA}{\includegraphics{\Img}}
\begin{figure}
	\centering
	\begin{tabular}{cccc}
		
		\includegraphics[trim=.29\WSA{} .14\HSA{} .34\WSA{} .16\HSA{},clip, width=\ElemWidth{}]{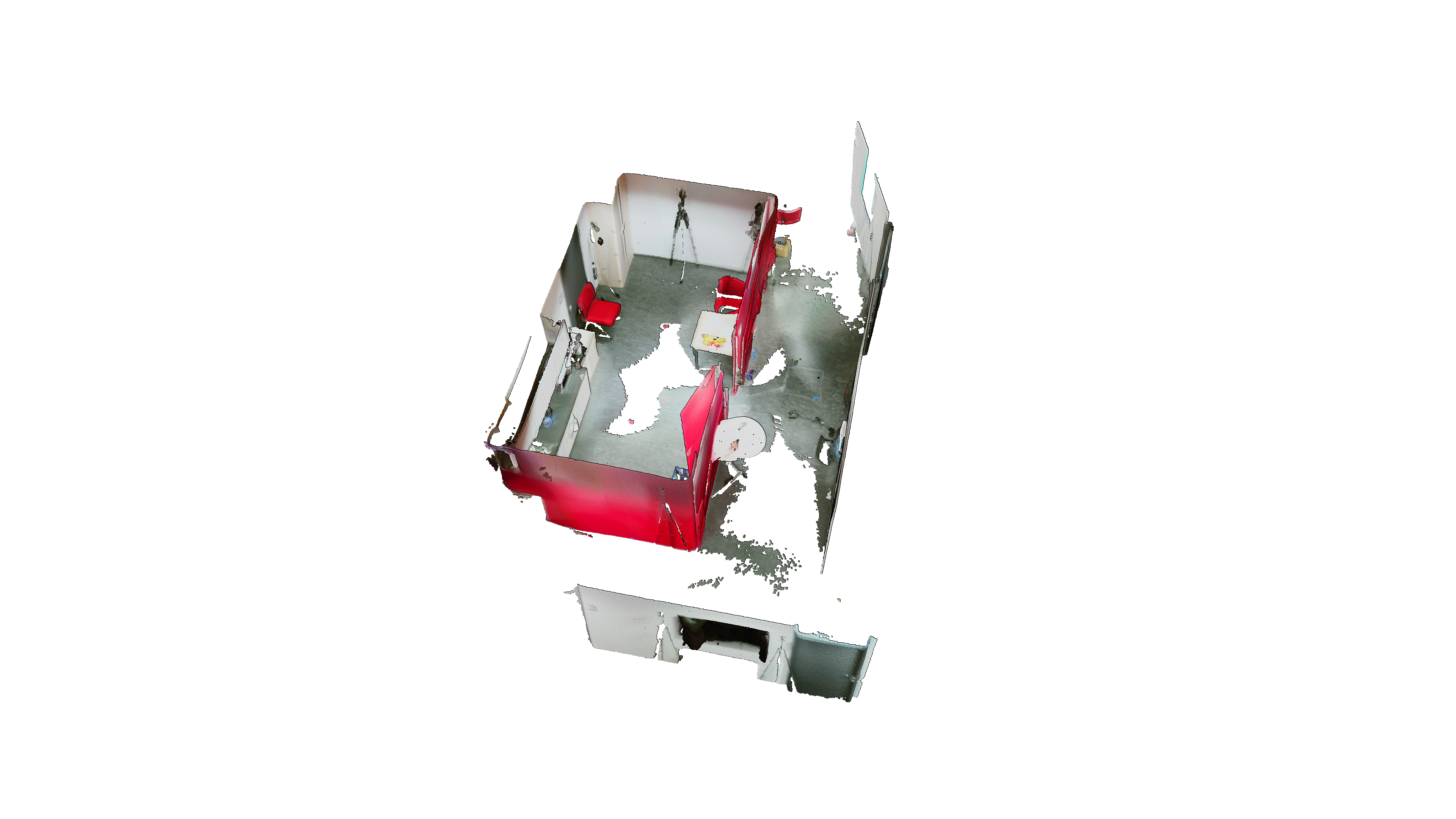}& 
		\includegraphics[trim=.33\WSA{} .21\HSA{} .37\WSA{} .16\HSA{},clip, width=\ElemWidth{}]{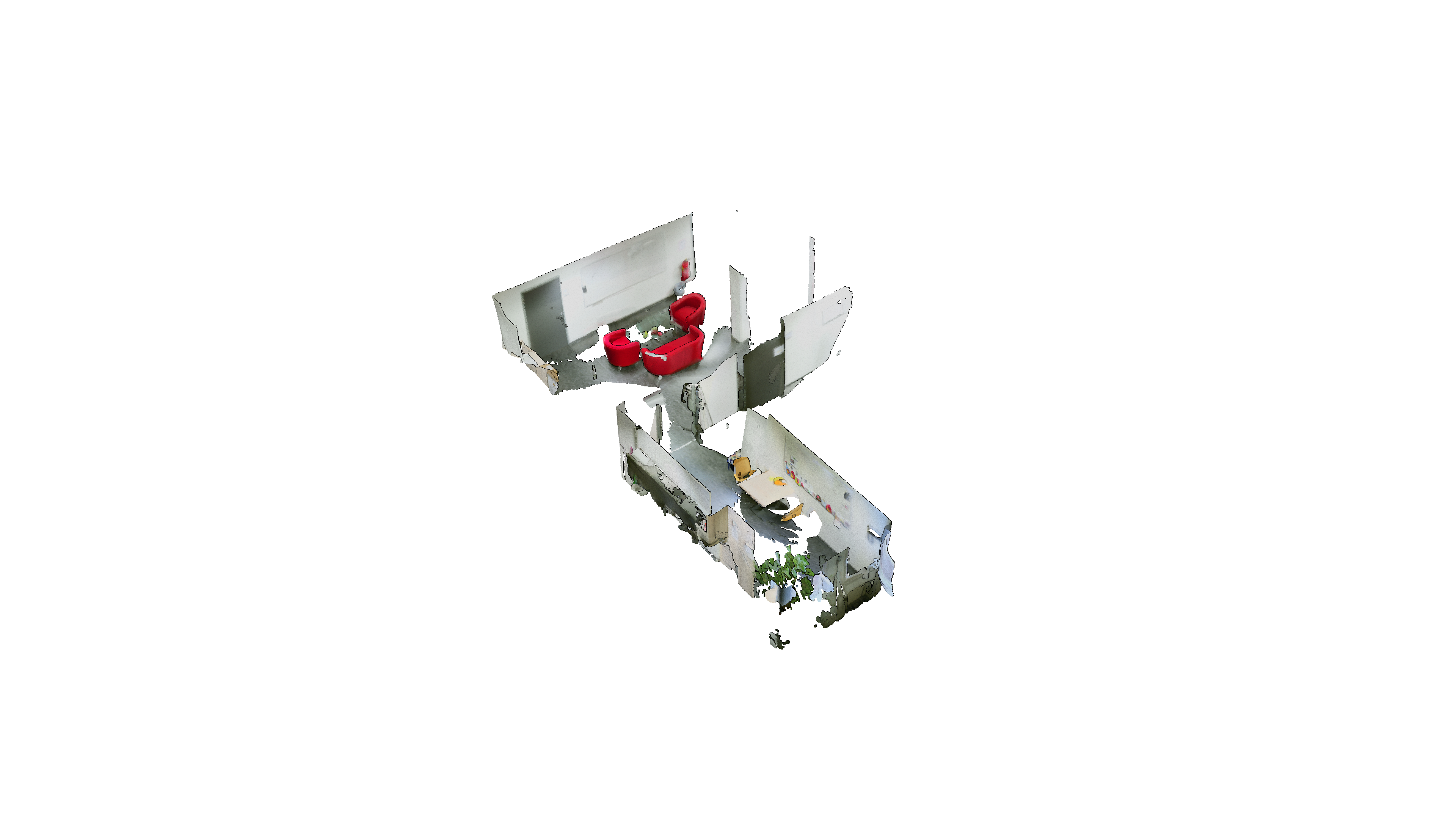}& 
		\includegraphics[trim=.29\WSA{} .15\HSA{} .32\WSA{} .16\HSA{},clip, width=\ElemWidth{}]{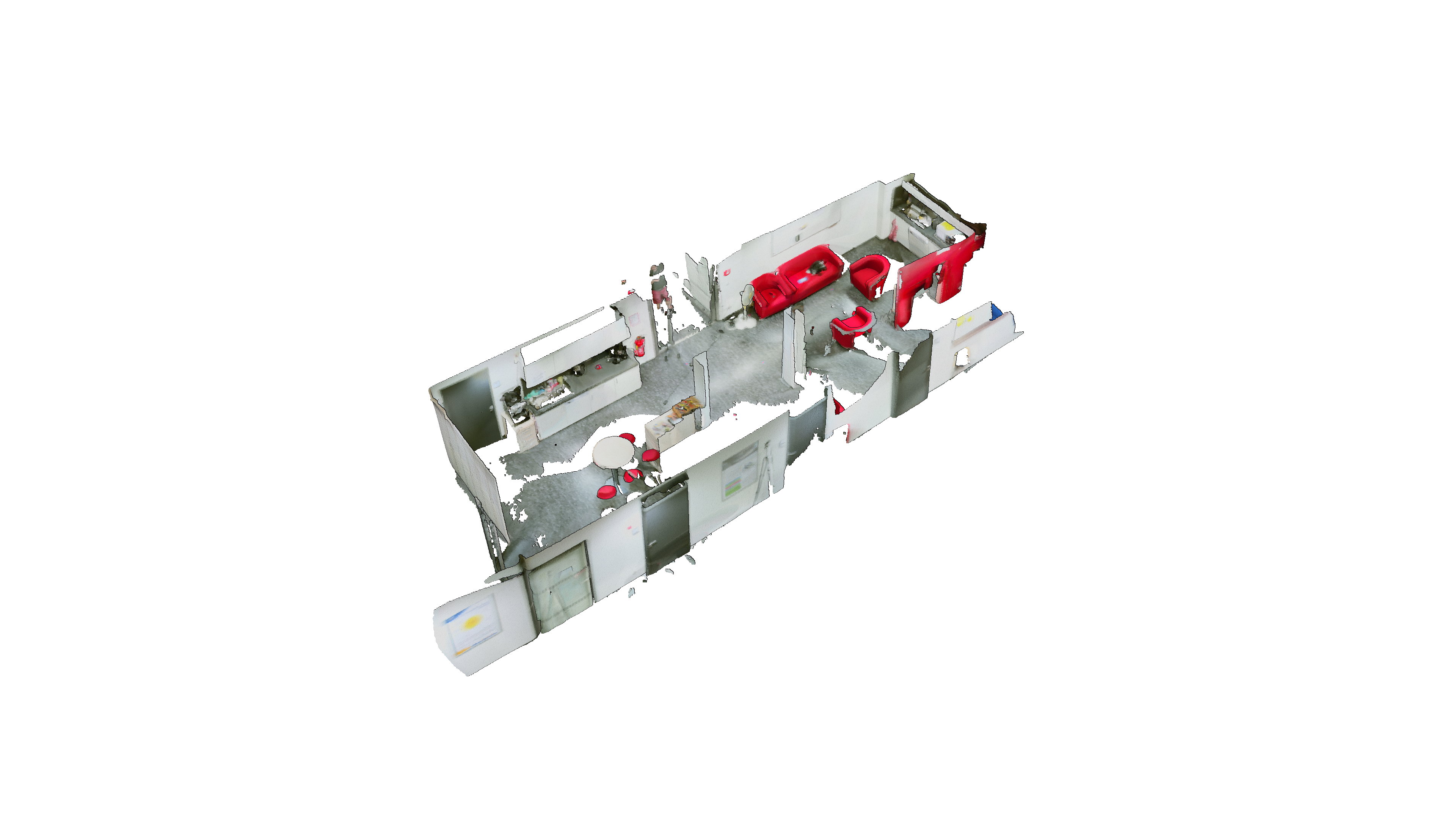}& 
		\includegraphics[trim=.33\WSA{} .2\HSA{} .30\WSA{} .16\HSA{},clip, width=\ElemWidth{}]{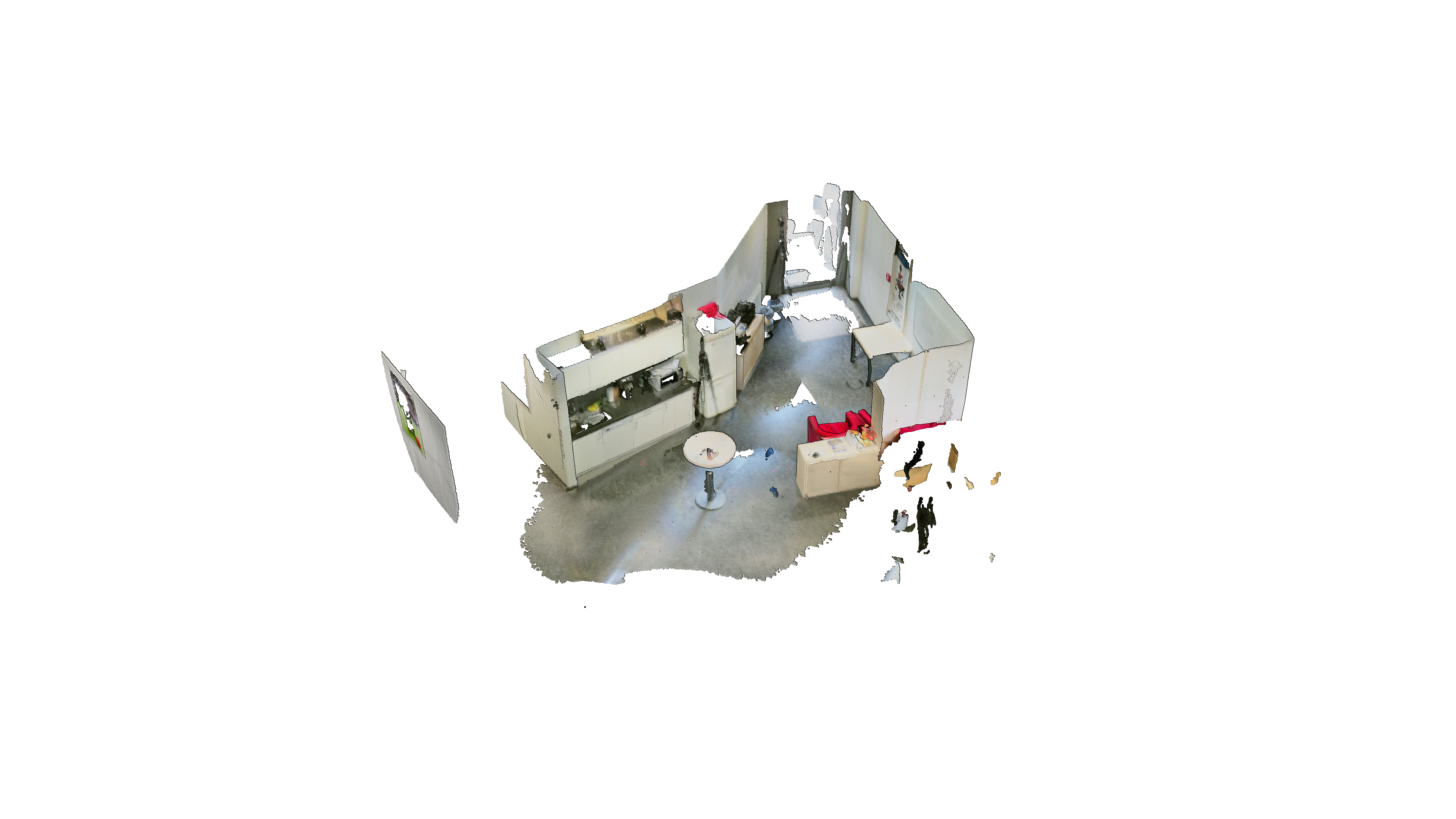} 
		\\ 
		
		\includegraphics[trim=.29\WSA{} .14\HSA{} .34\WSA{} .16\HSA{},clip, width=\ElemWidth{}]{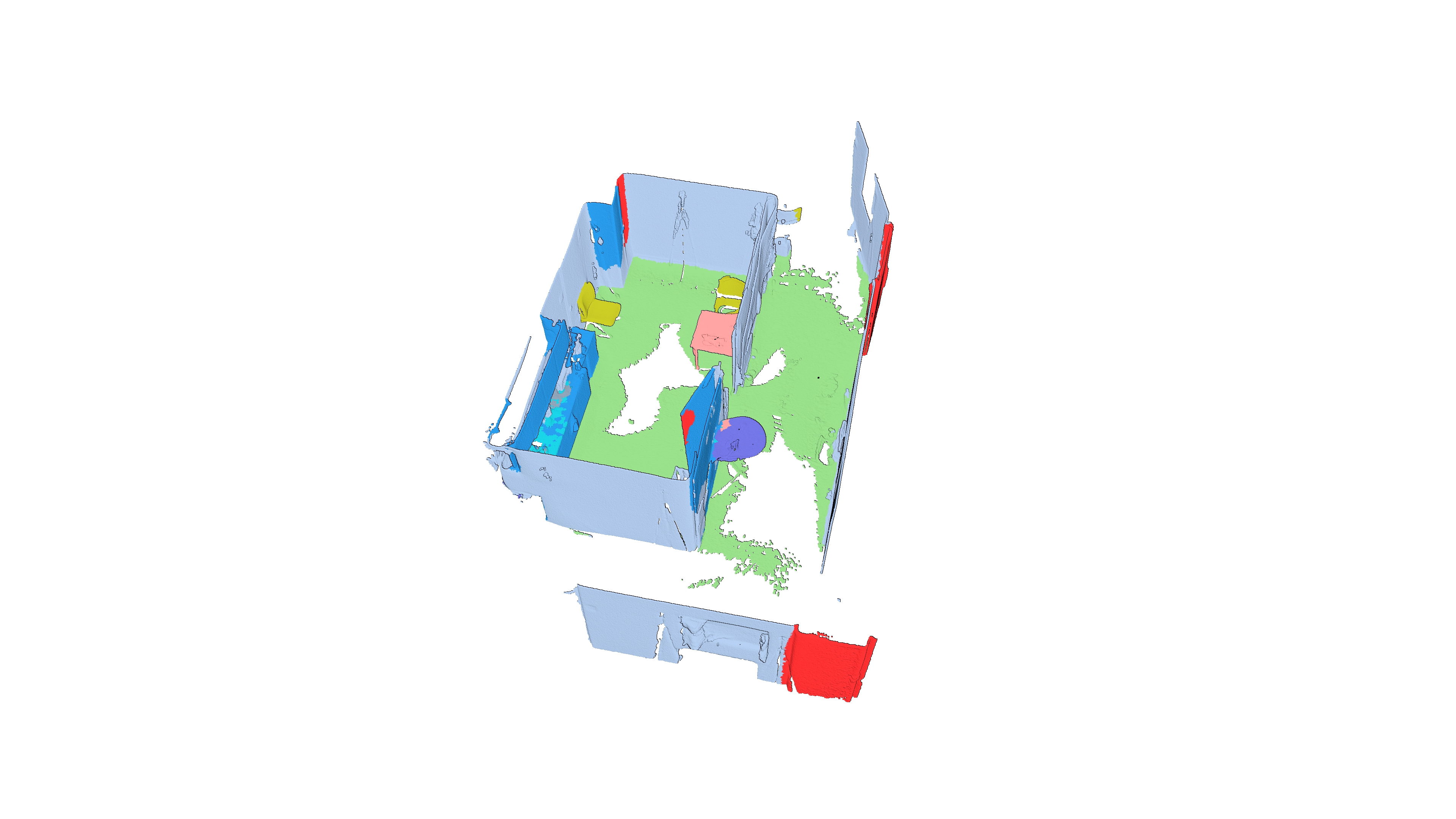}& 
		\includegraphics[trim=.33\WSA{} .21\HSA{} .37\WSA{} .16\HSA{},clip, width=\ElemWidth{}]{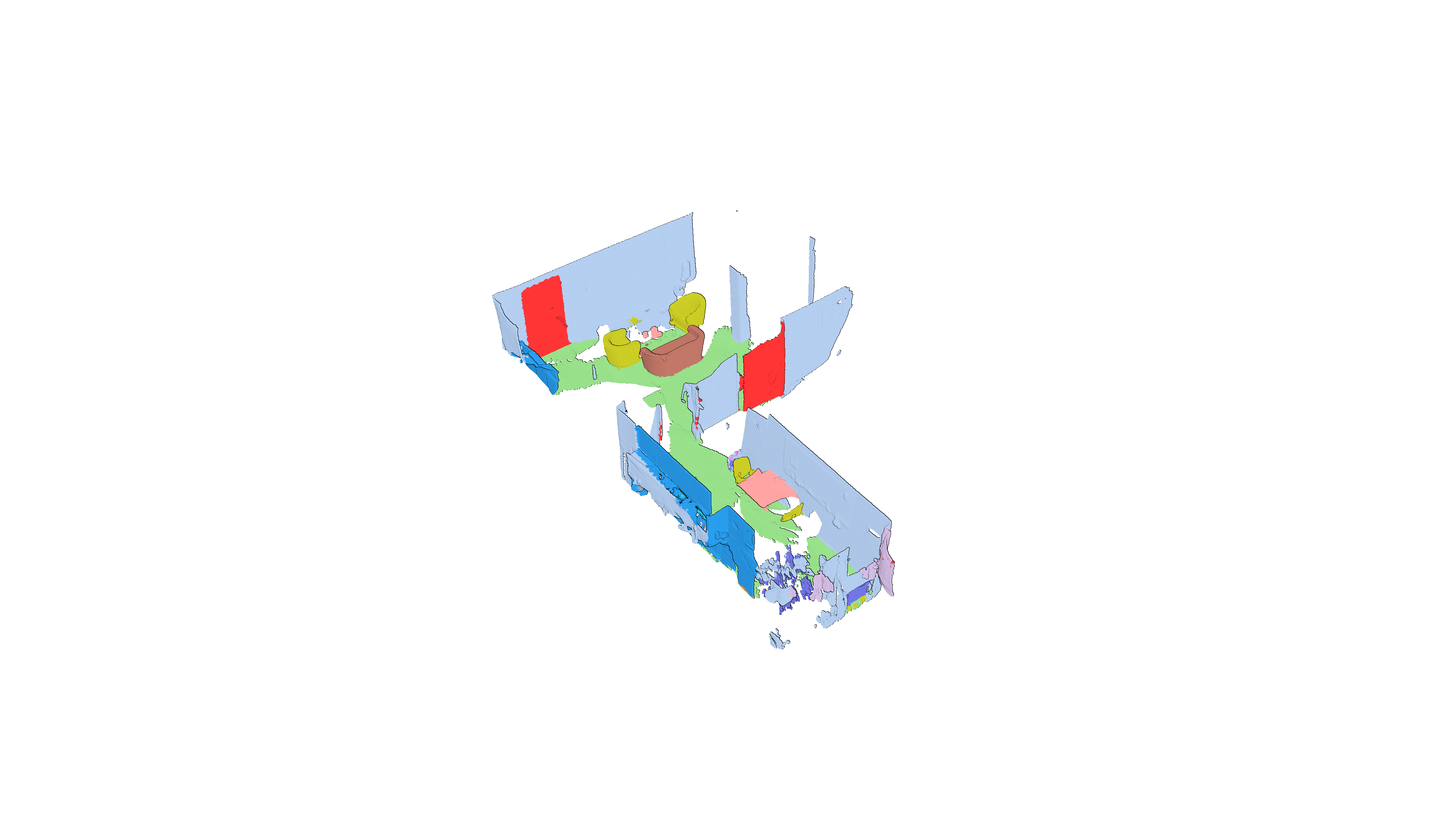}& 
		\includegraphics[trim=.29\WSA{} .15\HSA{} .32\WSA{} .16\HSA{},clip, width=\ElemWidth{}]{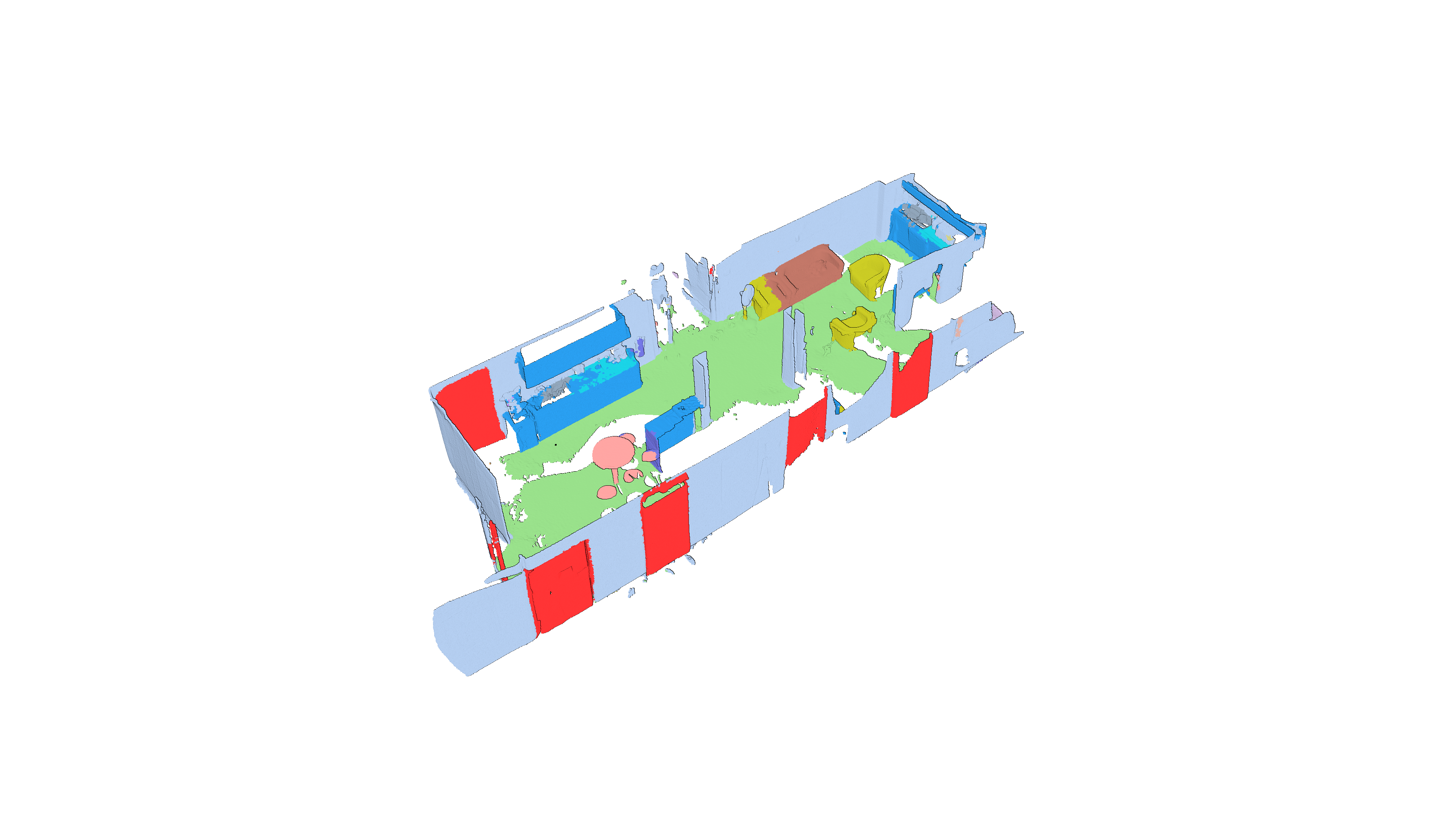}& 
		\includegraphics[trim=.33\WSA{} .2\HSA{} .30\WSA{} .16\HSA{},clip, width=\ElemWidth{}]{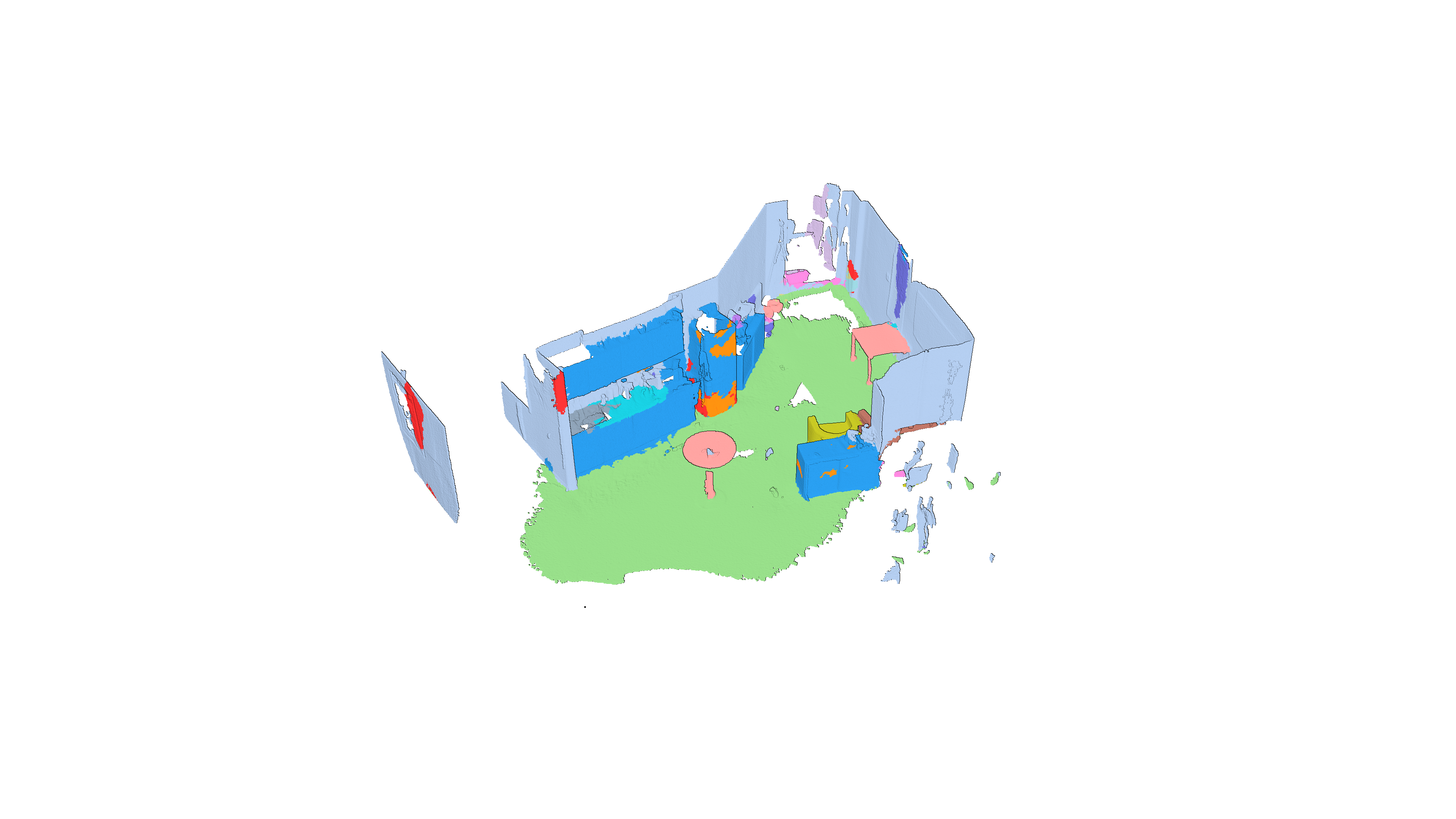}

	\end{tabular}
	\caption{ Bonn Activity Maps segmentations. Colored meshes are reconstructed from KinectV2 data using volumetric integration~\cite{niessner2013real,stotko2019efficient} and semantically segmented using LatticeNet. Color coding of semantic labels corresponds to the ScanNet dataset~\cite{dai2017scannet}.}
	\label{fig:SemPredBonnActivity}
\end{figure}
\egroup

	\bgroup
	\def\W{2.6cm}
	\setlength{\tabcolsep}{1pt}
	\def\ImgOne{./imgs/predictions/shapenet/lamp_65_gt.png}
	\def\ImgOnePred{./imgs/predictions/shapenet/lamp_65_pred.png}  
	\newlength{\WOne}
	\newlength{\HOne}
	\settowidth{\WOne}{\includegraphics{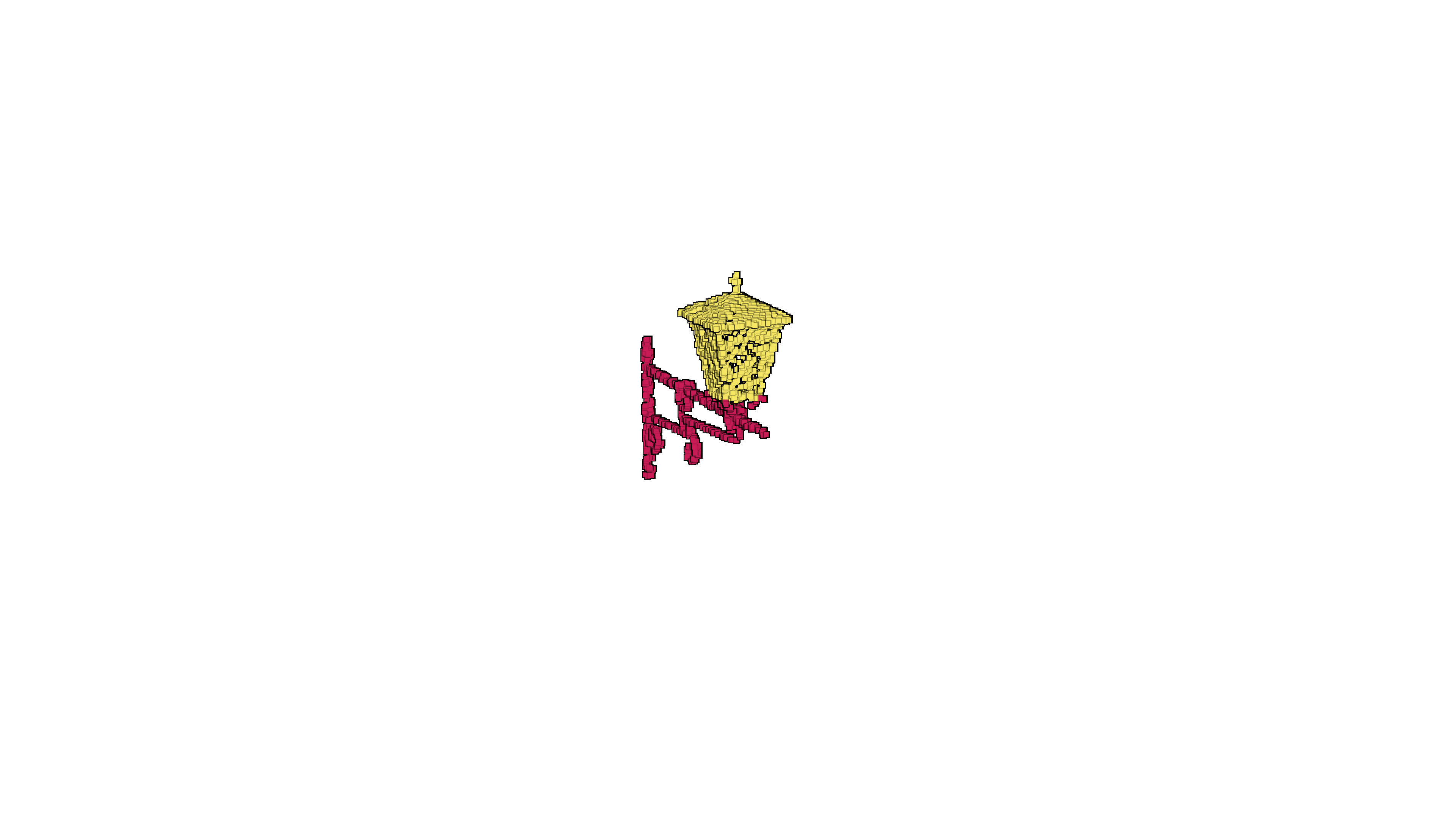}}
	\settoheight{\HOne}{\includegraphics{\ImgOne}}
	\def\ImgTwo{./imgs/predictions/shapenet/earphone_2_gt.png}
	\def\ImgTwoPred{./imgs/predictions/shapenet/earphone_2_pred.png}  
	\newlength{\WTwo}
	\newlength{\HTwo}
	\settowidth{\WTwo}{\includegraphics{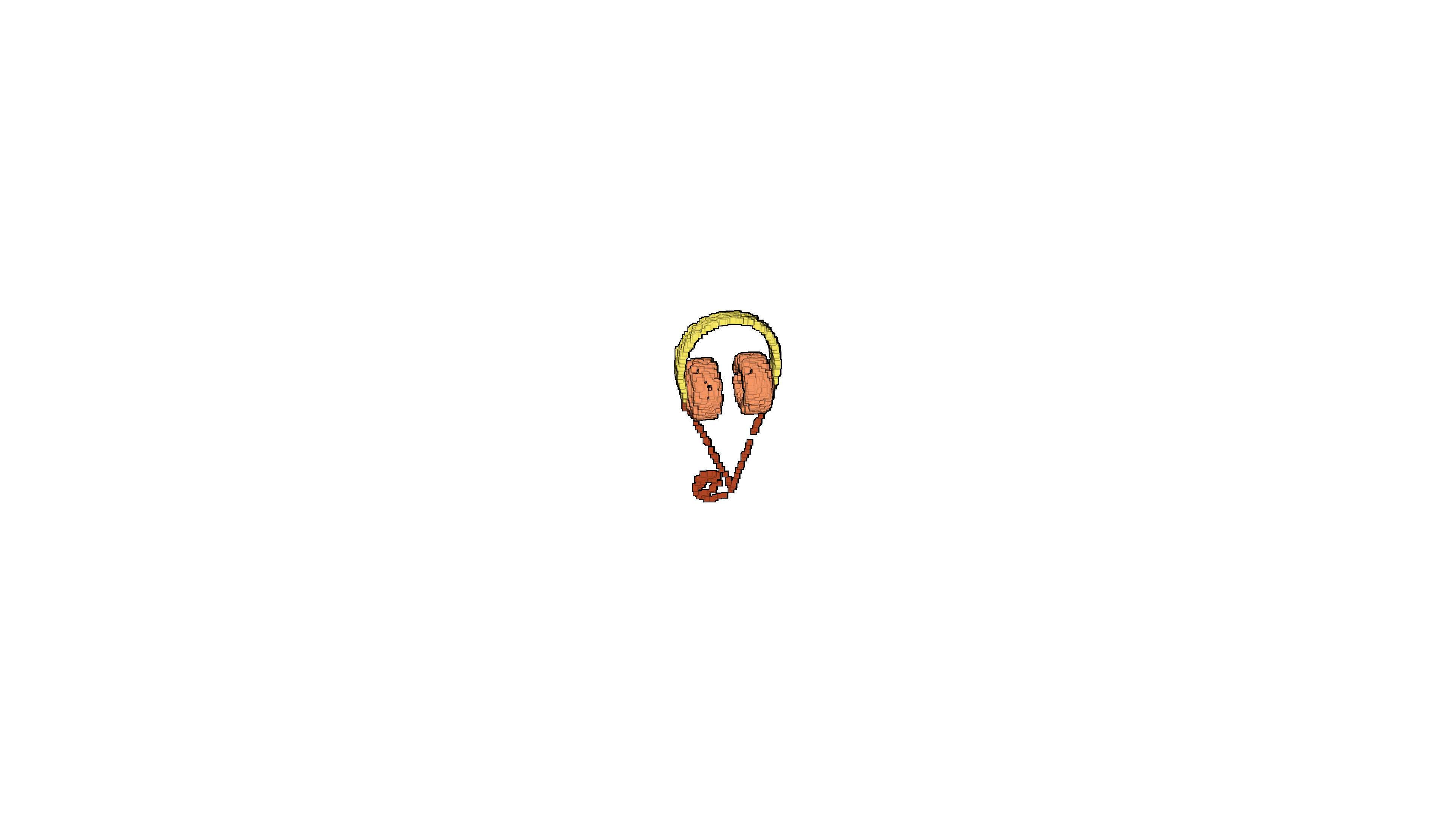}}
	\settoheight{\HTwo}{\includegraphics{\ImgTwo}}
	\def\ImgThree{./imgs/predictions/shapenet/chair_166_gt.png}
	\def\ImgThreePred{./imgs/predictions/shapenet/chair_166_pred.png}  
	\newlength{\WThree}
	\newlength{\HThree}
	\settowidth{\WThree}{\includegraphics{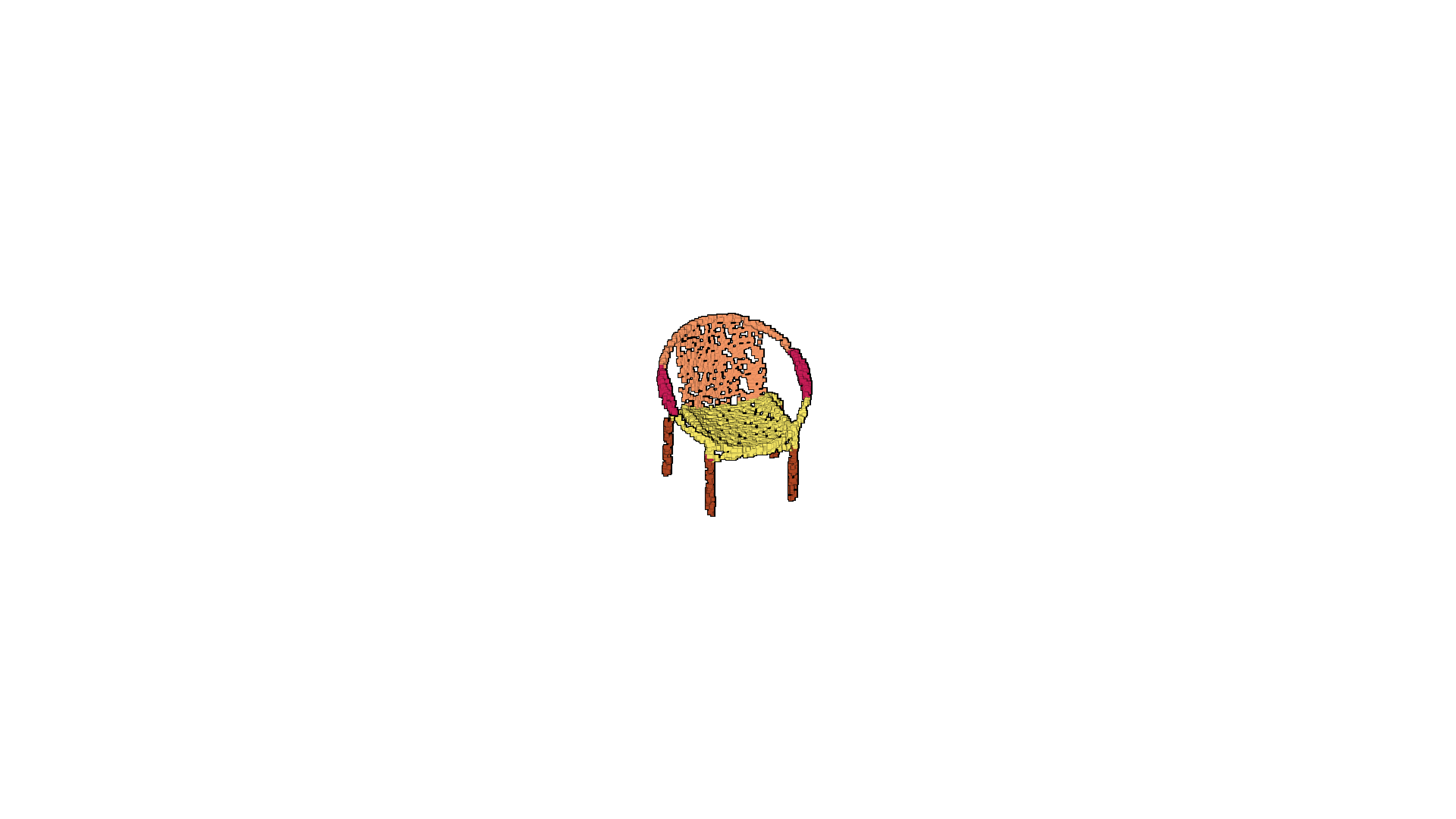}}
	\settoheight{\HThree}{\includegraphics{\ImgThree}}
	\begin{figure}
		\centering
		\begin{tabular}{cC{\W}C{\W}C{\W}}
			\centering
			\setlength{\tabcolsep}{1pt}
			\multirow{1}{*}[10pt]{\rotatebox{90}{Ground-truth}} &
			\includegraphics[align=c,trim=.43\WOne{} .4\HOne{} .445\WOne{} .33\HOne{},clip, width=.20\columnwidth]{\ImgOne}&
			\includegraphics[align=c,trim=.43\WTwo{} .38\HTwo{} .44\WTwo{} .38\HTwo{},clip, width=.22\columnwidth]{\ImgTwo}&
			\includegraphics[align=c,trim=.43\WTwo{} .36\HTwo{} .44\WTwo{} .35\HTwo{},clip, width=.22\columnwidth]{\ImgThree}\\
			\multirow{1}{*}[5pt]{\rotatebox{90}{Prediction}} &
			\includegraphics[align=c,trim=.43\WOne{} .4\HOne{} .445\WOne{} .33\HOne{},clip, width=.20\columnwidth]{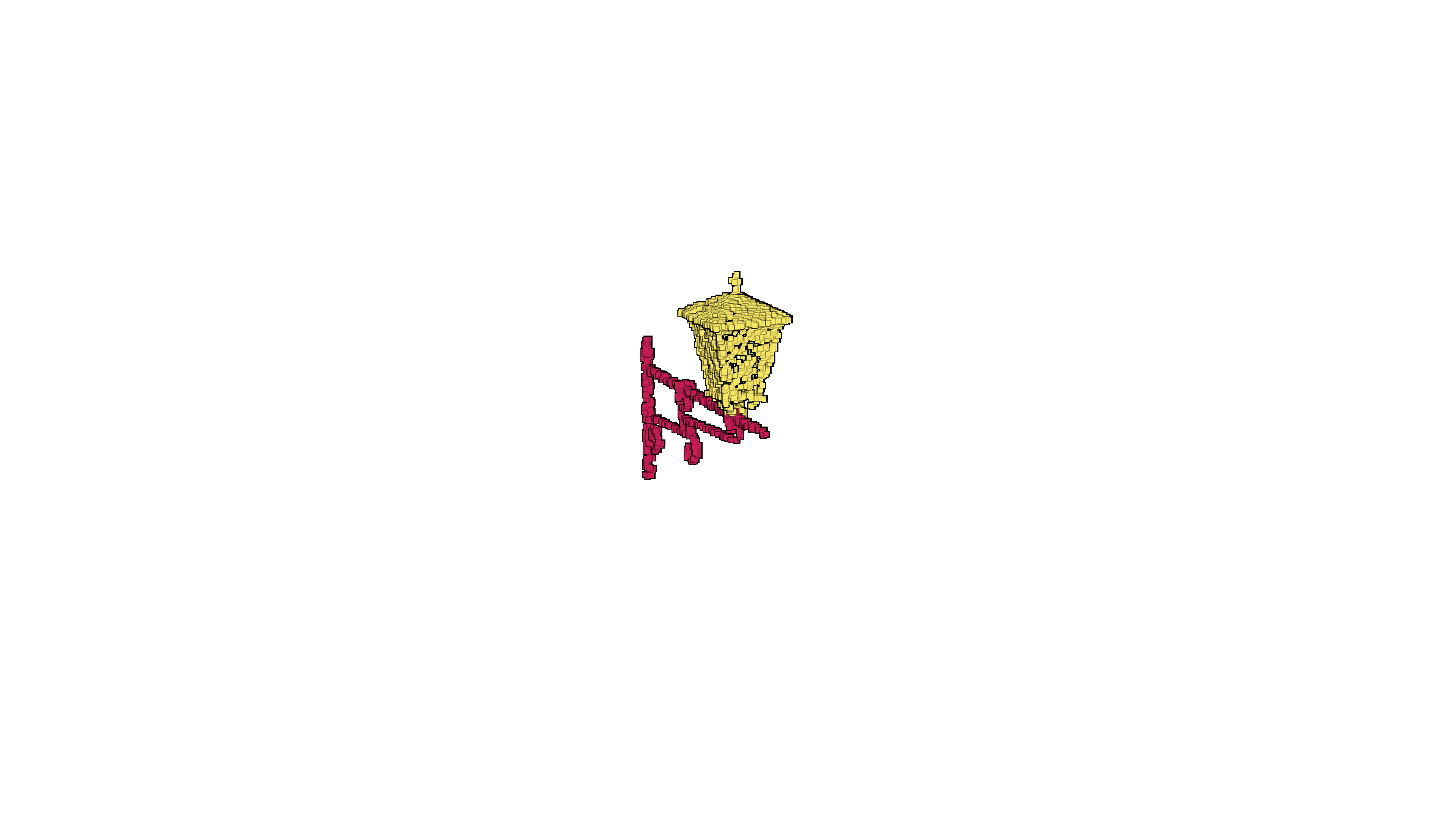}&
			\includegraphics[align=c,trim=.43\WTwo{} .38\HTwo{} .44\WTwo{} .38\HTwo{},clip, width=.22\columnwidth]{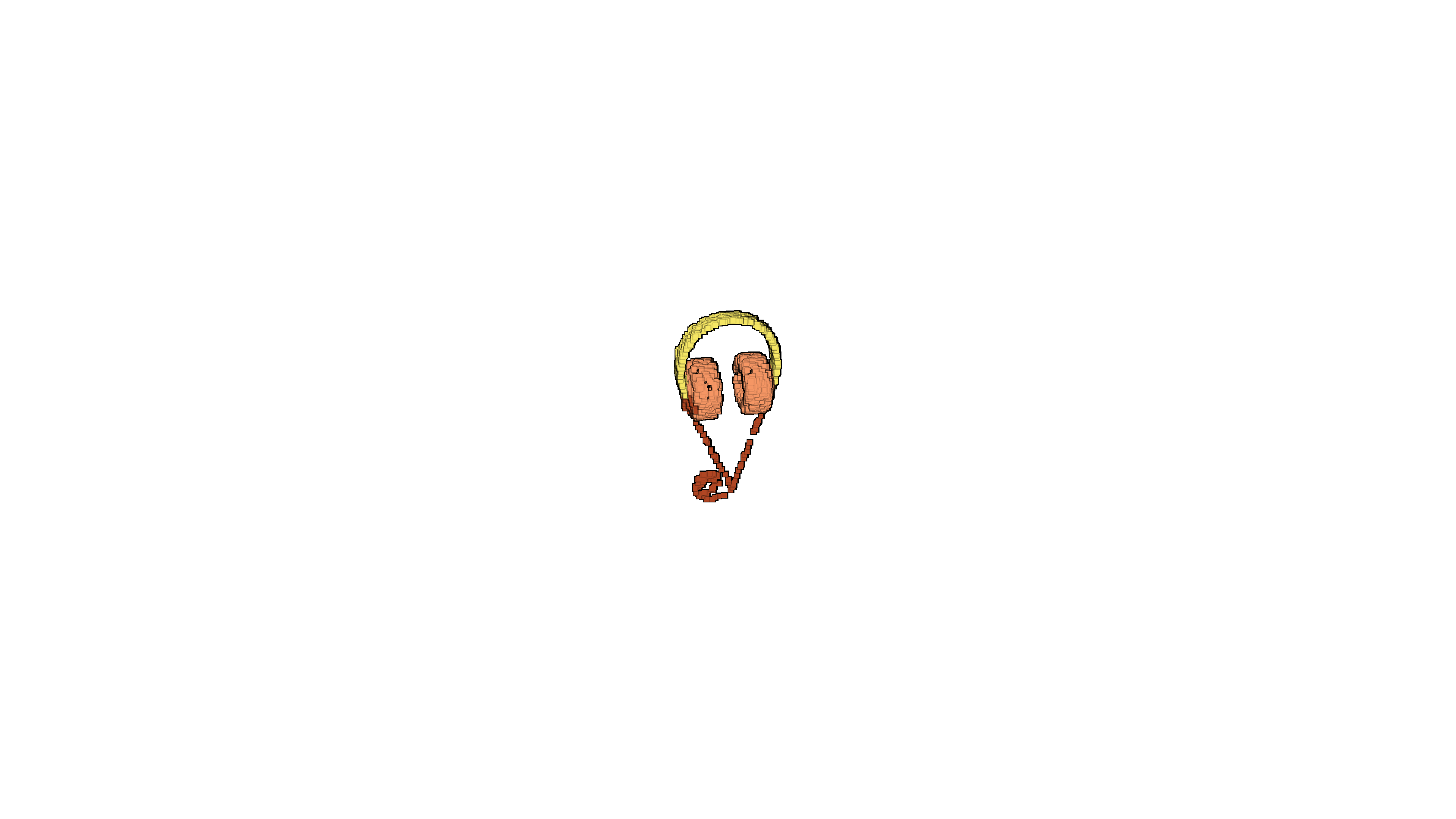}&
			\includegraphics[align=c,trim=.43\WTwo{} .36\HTwo{} .44\WTwo{} .35\HTwo{},clip, width=.22\columnwidth]{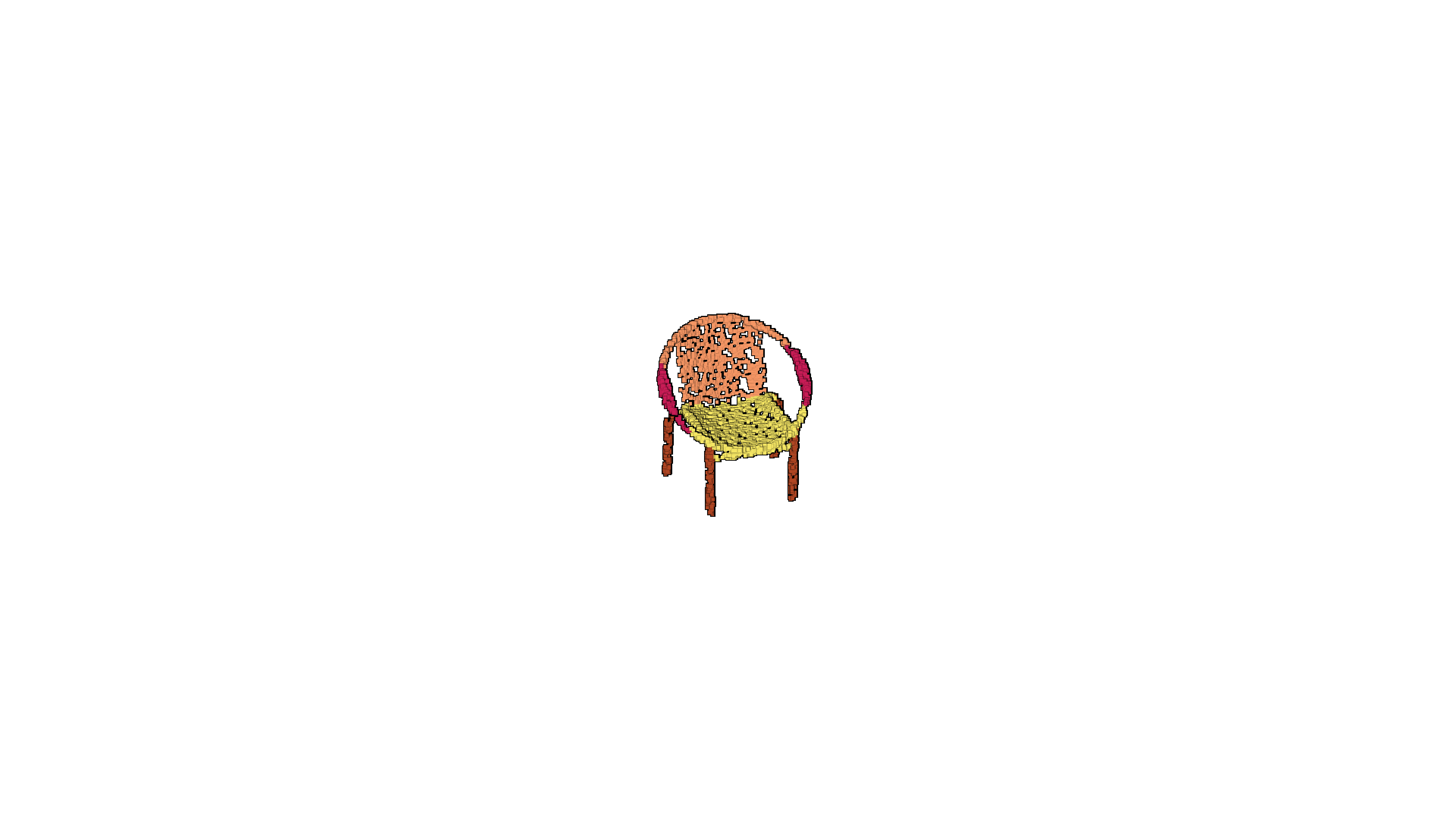}\\
		\end{tabular}
		\caption{ShapeNet~\cite{yi2016scalable} results of our method.}
		\label{fig:ShapeNetImages}
	\end{figure}
	\egroup

	\bgroup
	\def\W{2.6cm}
	\setlength{\tabcolsep}{1pt}
	\def\ImgMot{./imgs/motion_seg/val_004032_3_higher_saturation.png}
	\newlength{\WMot}
	\newlength{\HMot}
	\settowidth{\WMot}{\includegraphics{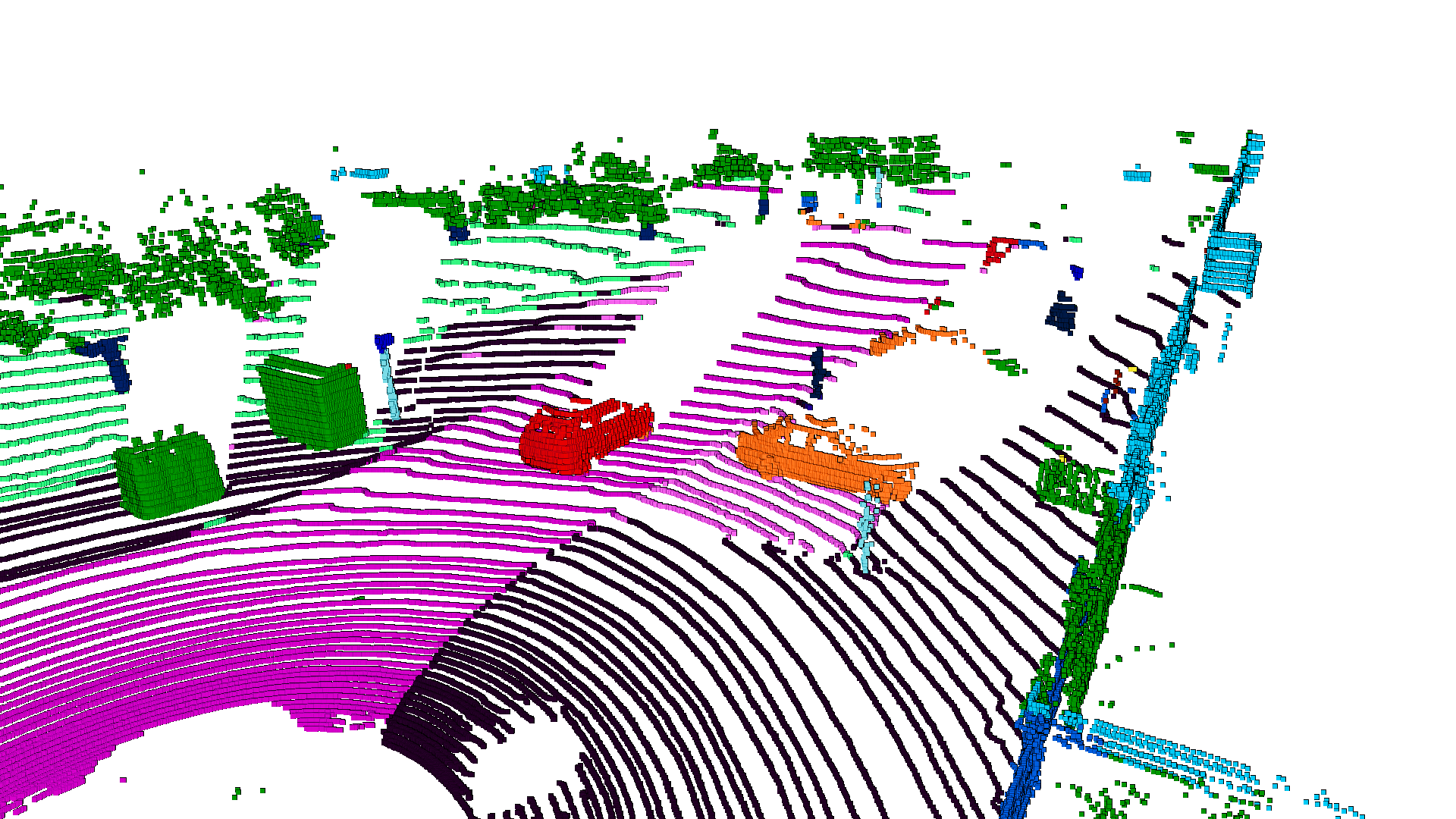}}
	\settoheight{\HMot}{\includegraphics{\ImgMot}}
	\begin{figure}
		\centering
		\includegraphics[align=c,trim=.25\WMot{} .25\HMot{} .25\WMot{} .25\HMot{},clip, width=.9\columnwidth]{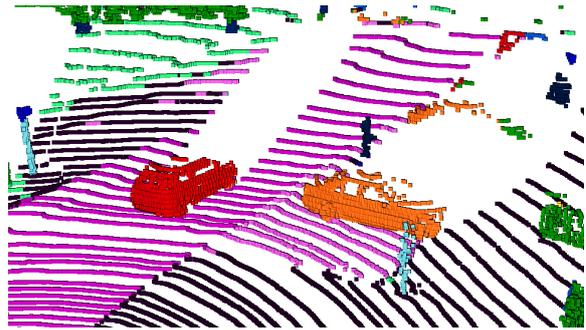} 
		\caption{Motion segmentation results on SemanticKITTI. The moving car on the road (red) is correctly distinguished from the parked car (orange). }
		\label{fig:motion_seg}
	\end{figure}
	\egroup

		\bgroup
		\def\W{\columnwidth}
		\setlength{\tabcolsep}{1pt}
		\def\ImgGt{./imgs/predictions/semantickitti/gt_3.png}
		\def\ImgMine{./imgs/predictions/semantickitti/mine_3.png}  
		\def\ImgTangentConv{./imgs/predictions/semantickitti/tangent_3.png}
		\def\ImgSplatNet{./imgs/predictions/semantickitti/splatnet_3.png}
		\newlength{\WSemK}
		\newlength{\HSemK}
		\settowidth{\WSemK}{\includegraphics{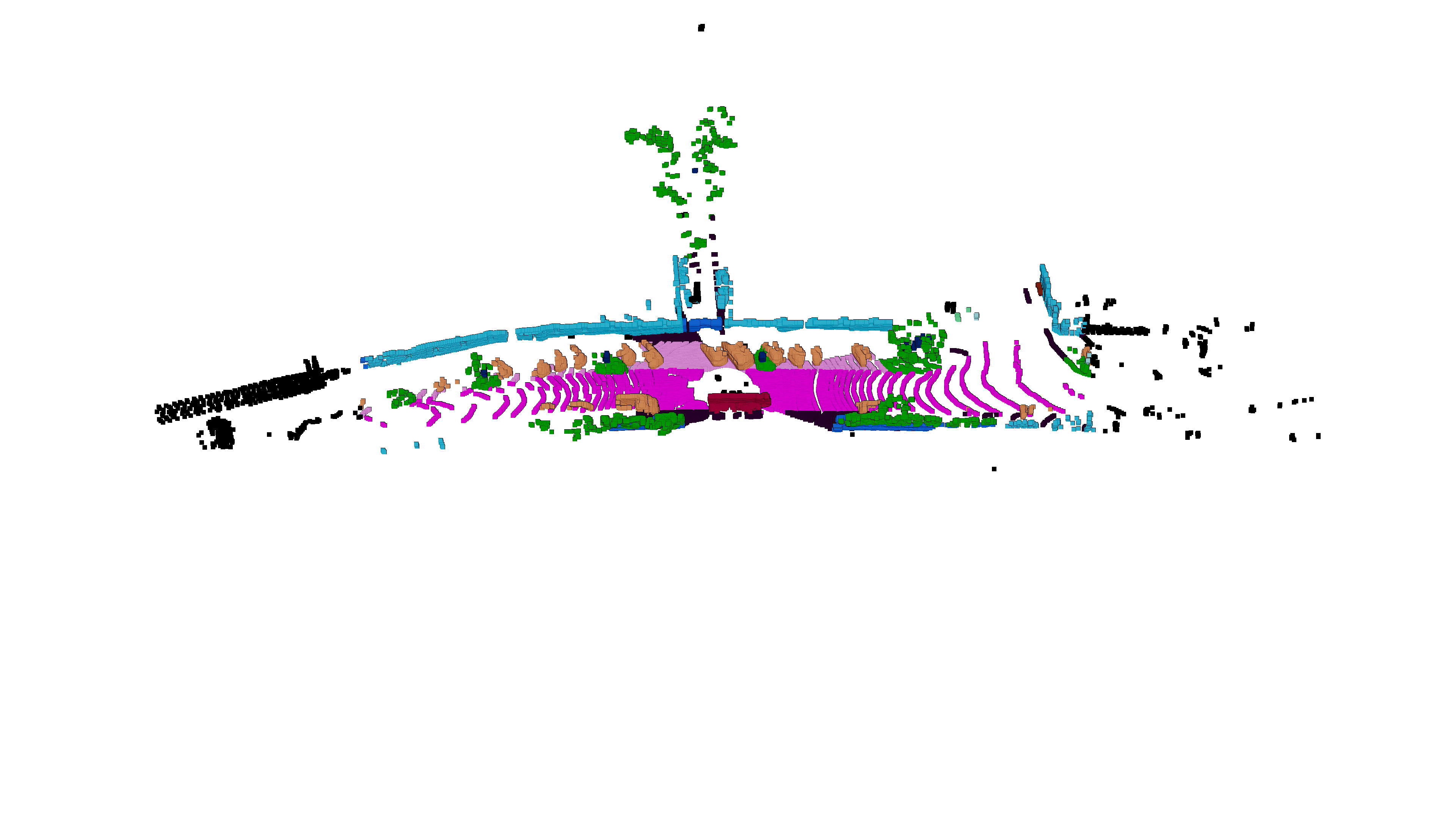}}
		\settoheight{\HSemK}{\includegraphics{\ImgGt}}
		\begin{figure*}
			\centering
			\hskip-6.5cm 
			\begin{tabular}{C{\W}}
				\centering
				\setlength{\tabcolsep}{1pt}
				\begin{tikzpicture}[spy using outlines={rectangle,yellow,magnification=2,size=1.5cm}]
				\node [minimum width={width("Ground truth")+7pt} ] at (-5.5,0) {Ground truth};
				\node {\includegraphics[align=c,trim=.24\WSemK{} .45\HSemK{} .31\WSemK{} .379\HSemK{},clip, width=.9\columnwidth]{\ImgGt}};
				\spy [blue!80!white, every spy on node/.append style={line width=0.7mm}, every spy in node/.append style={line width=1.0mm} ] on (-2.3, 0.1) in node [left] at (6, 0.1);
				\spy [orange, every spy on node/.append style={line width=0.7mm}, every spy in node/.append style={line width=1.0mm} ] on (0.05, 0.3) in node [left] at (7.8, 0.1);
				\spy [black!40!green, every spy on node/.append style={line width=0.7mm}, every spy in node/.append style={line width=1.0mm} ] on (0.8, -0.3) in node [left] at (9.6, 0.1);
				\node [] at (5.2, 1.2) {Tree trunk};
				\node [] at (7, 1.2) {Parking};
				\node [] at (8.8, 1.2) {Truck};
				\end{tikzpicture}\\

				\begin{tikzpicture}[spy using outlines={rectangle,yellow,magnification=2,size=1.5cm}]
				\node [minimum width={width("Ground truth")+7pt} ] at (-5.5,0) {LatticeNet};
				\node {\includegraphics[align=c,trim=.24\WSemK{} .45\HSemK{} .31\WSemK{} .379\HSemK{},clip, width=.9\columnwidth]{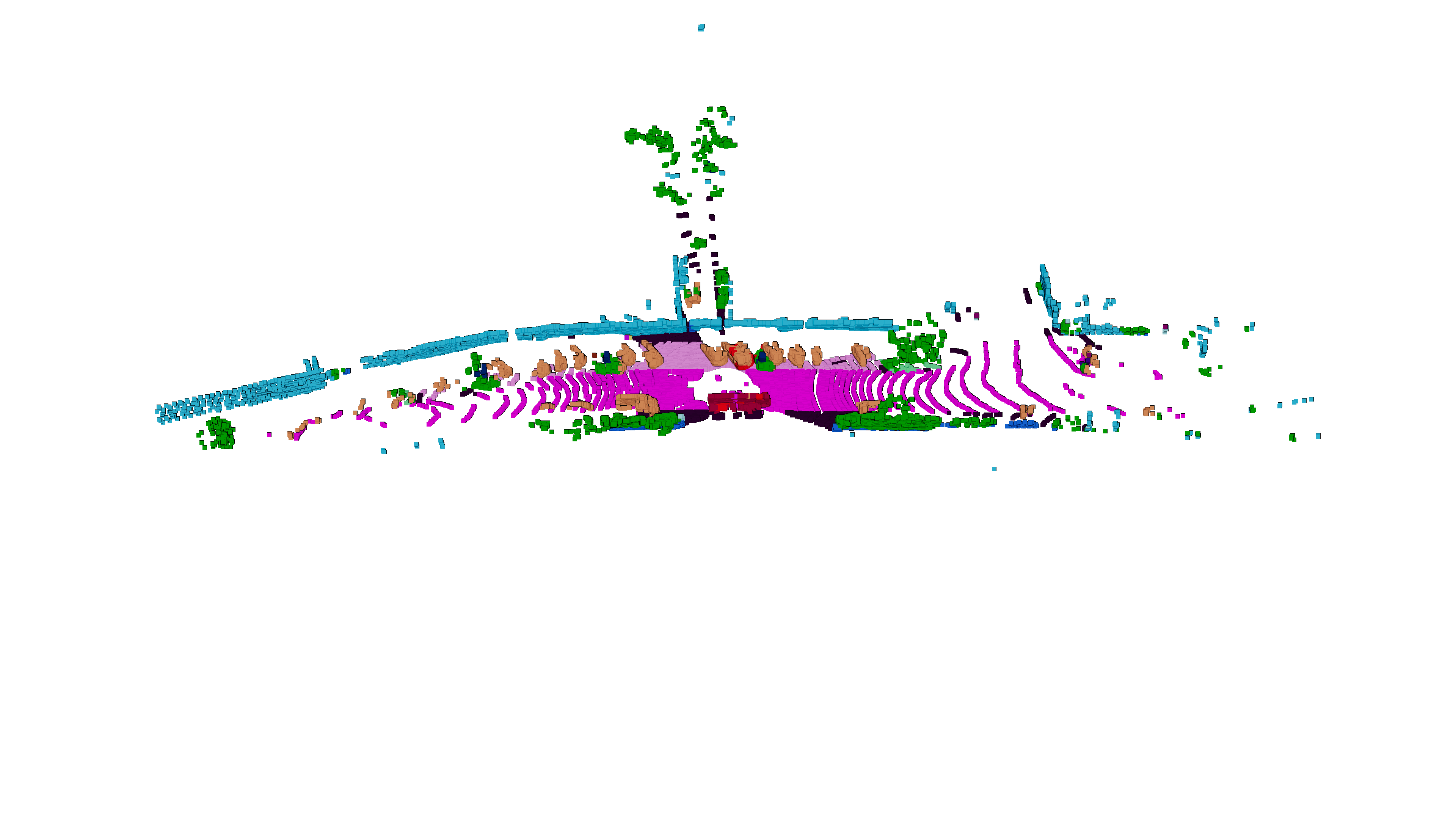}};
				\spy [blue!80!white, every spy on node/.append style={ultra thick, draw=none}, every spy in node/.append style={line width=1.0mm} ] on (-2.3, 0.1) in node [left] at (6, 0.1);
				\spy [orange, every spy on node/.append style={ultra thick, draw=none}, every spy in node/.append style={line width=1.0mm} ] on (0.05, 0.3) in node [left] at (7.8, 0.1);
				\spy [black!40!green, every spy on node/.append style={ultra thick, draw=none}, every spy in node/.append style={line width=1.0mm} ] on (0.8, -0.3) in node [left] at (9.6, 0.1);
				\end{tikzpicture}\\
				
				\begin{tikzpicture}[spy using outlines={rectangle,yellow,magnification=2,size=1.5cm}]
				\node [minimum width={width("Ground truth")+7pt} ] at (-5.5,0) {TangentConv};
				\node {\includegraphics[align=c,trim=.24\WSemK{} .45\HSemK{} .31\WSemK{} .379\HSemK{},clip, width=.9\columnwidth]{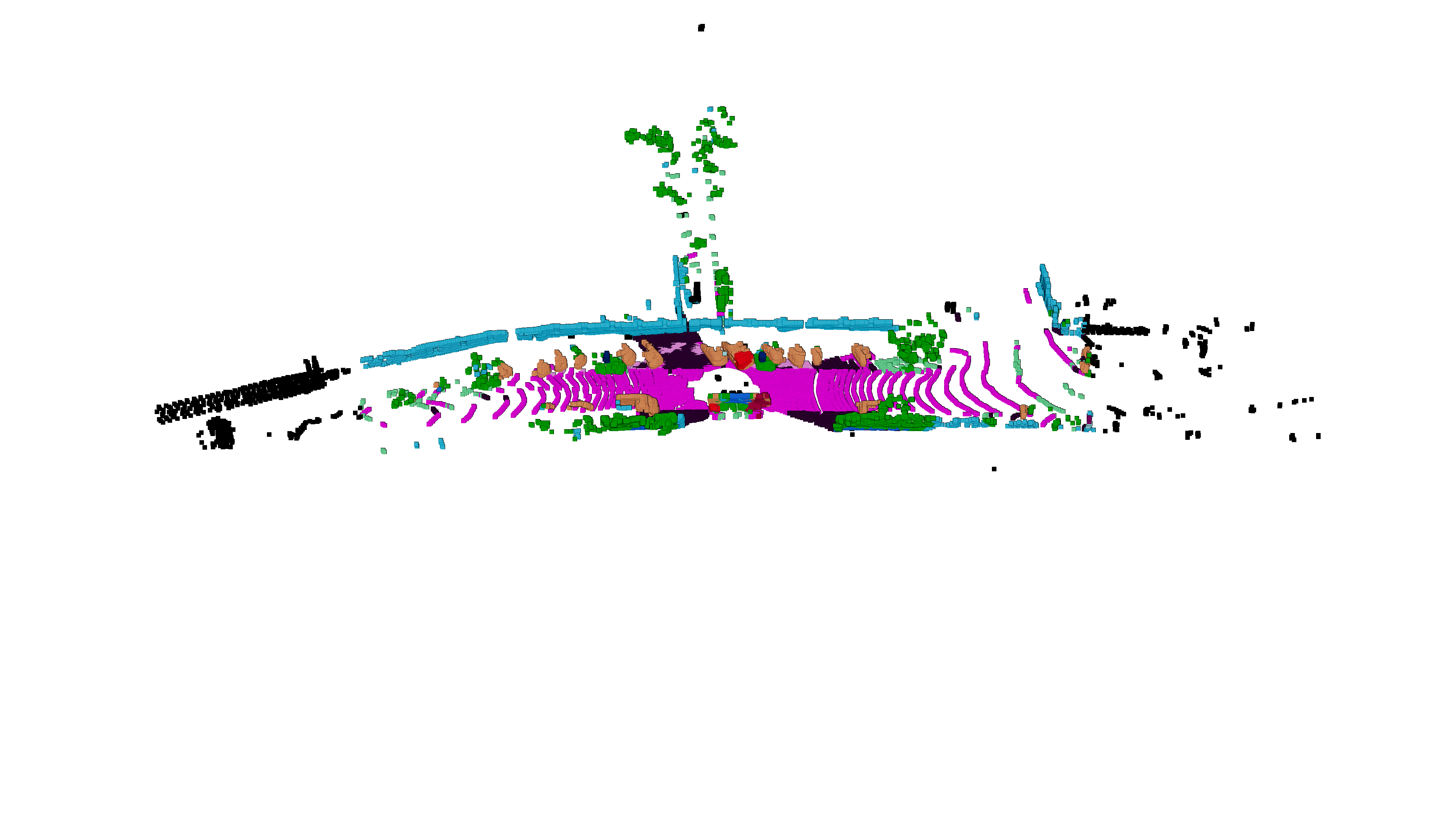}};
				\spy [blue!80!white, every spy on node/.append style={ultra thick, draw=none}, every spy in node/.append style={line width=1.0mm} ] on (-2.3, 0.1) in node [left] at (6, 0.1);
				\spy [orange, every spy on node/.append style={ultra thick, draw=none}, every spy in node/.append style={line width=1.0mm} ] on (0.05, 0.3) in node [left] at (7.8, 0.1);
				\spy [black!40!green, every spy on node/.append style={ultra thick, draw=none}, every spy in node/.append style={line width=1.0mm} ] on (0.8, -0.3) in node [left] at (9.6, 0.1);
				\end{tikzpicture}\\
				
				\begin{tikzpicture}[spy using outlines={rectangle,yellow,magnification=2,size=1.5cm}]
				\node [minimum width={width("Ground truth")+7pt} ] at (-5.5,0) {SplatNet};
				\node {\includegraphics[align=c,trim=.24\WSemK{} .45\HSemK{} .31\WSemK{} .379\HSemK{},clip, width=.9\columnwidth]{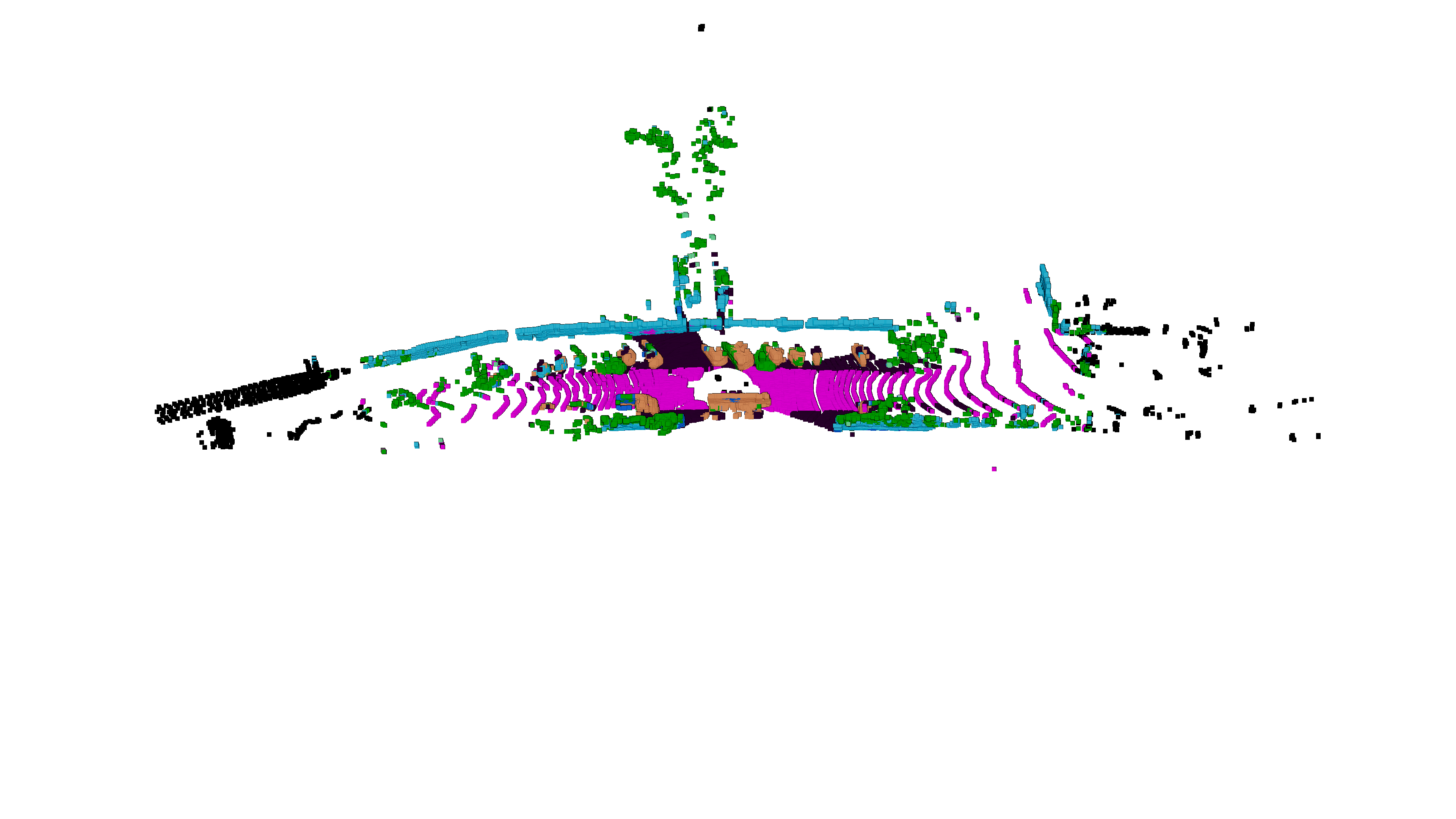}};
				\spy [blue!80!white, every spy on node/.append style={ultra thick, draw=none}, every spy in node/.append style={line width=1.0mm} ] on (-2.3, 0.1) in node [left] at (6, 0.1);
				\spy [orange, every spy on node/.append style={ultra thick, draw=none}, every spy in node/.append style={line width=1.0mm} ] on (0.05, 0.3) in node [left] at (7.8, 0.1);
				\spy [black!40!green, every spy on node/.append style={ultra thick, draw=none}, every spy in node/.append style={line width=1.0mm} ] on (0.8, -0.3) in node [left] at (9.6, 0.1);
				\end{tikzpicture}\\

			\end{tabular}
			\caption{SemanticKITTI results. We compare the prediction from our LatticeNet with the results from TangentConv~\cite{tatarchenko2018tangent} and SplatNet~\cite{su2018splatnet}. We can observe that our approach can better learn small objects like tree trunks, despite their relatively small number of points. Additionally, the network also effectively makes use of contextual information in order to correctly predict the parking place due to the existence of nearby cars.}
			\label{fig:SemanticKittiImages}
		\end{figure*}
		\egroup

	\bgroup
	\def\W{0.5\columnwidth}
	\setlength{\tabcolsep}{1pt}
	\def\ImgOne{./imgs/predictions/scannet/16_gt.png}
	\def\ImgOnePred{./imgs/predictions/scannet/16_pred.png}  
	\newlength{\WScan}
	\newlength{\HScan}
	\settowidth{\WScan}{\includegraphics{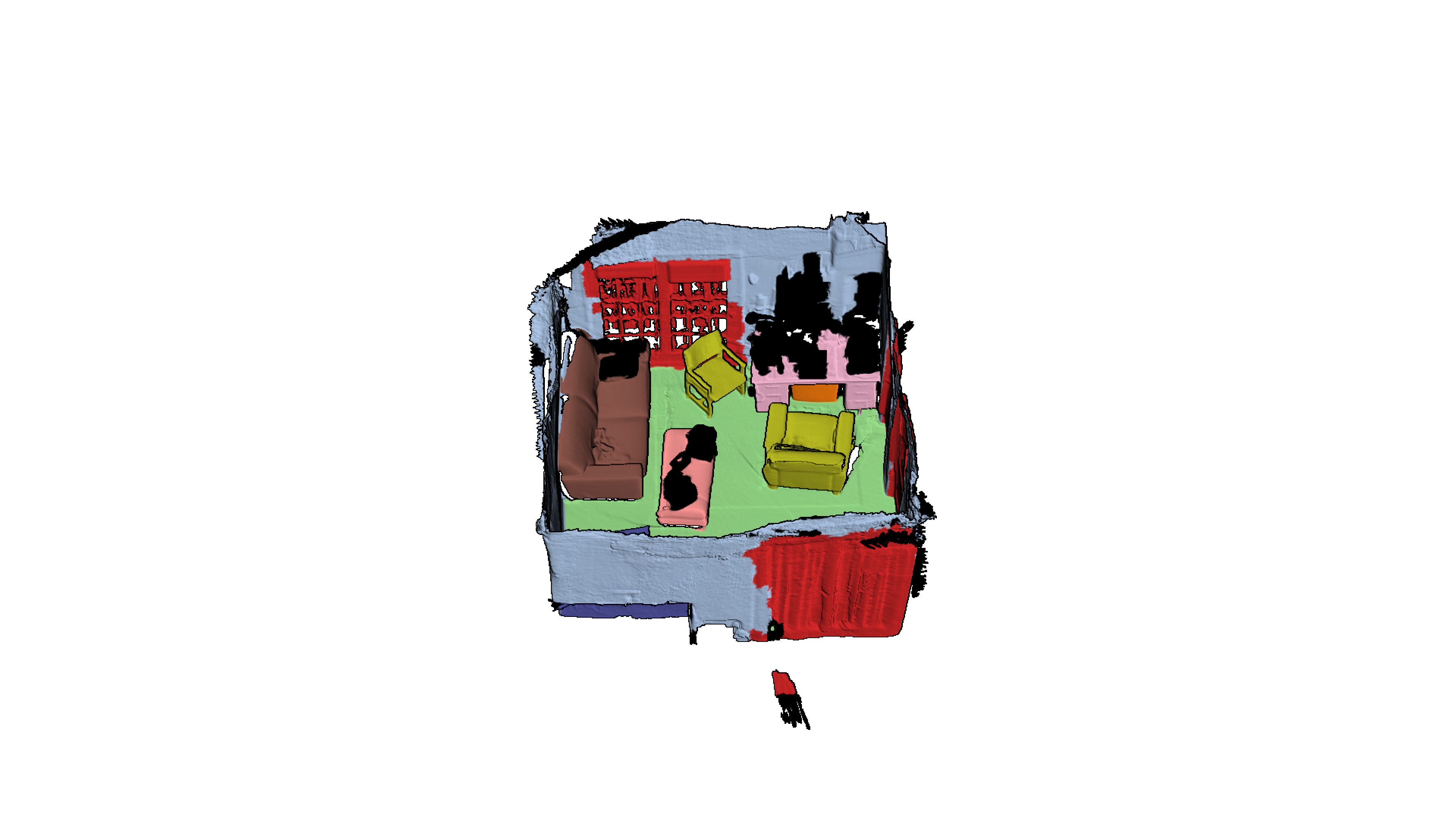}}
	\settoheight{\HScan}{\includegraphics{\ImgOne}}
	\begin{figure}[h!]
		\centering
		\begin{tabular}{C{\W}C{\W}}
			\centering
			\setlength{\tabcolsep}{1pt}
			\includegraphics[align=c,trim=.35\WScan{} .19\HScan{} .35\WScan{} .24\HScan{},clip, width=.35\columnwidth]{\ImgOne} &
			\includegraphics[align=c,trim=.35\WScan{} .19\HScan{} .35\WScan{} .24\HScan{},clip, width=.35\columnwidth]{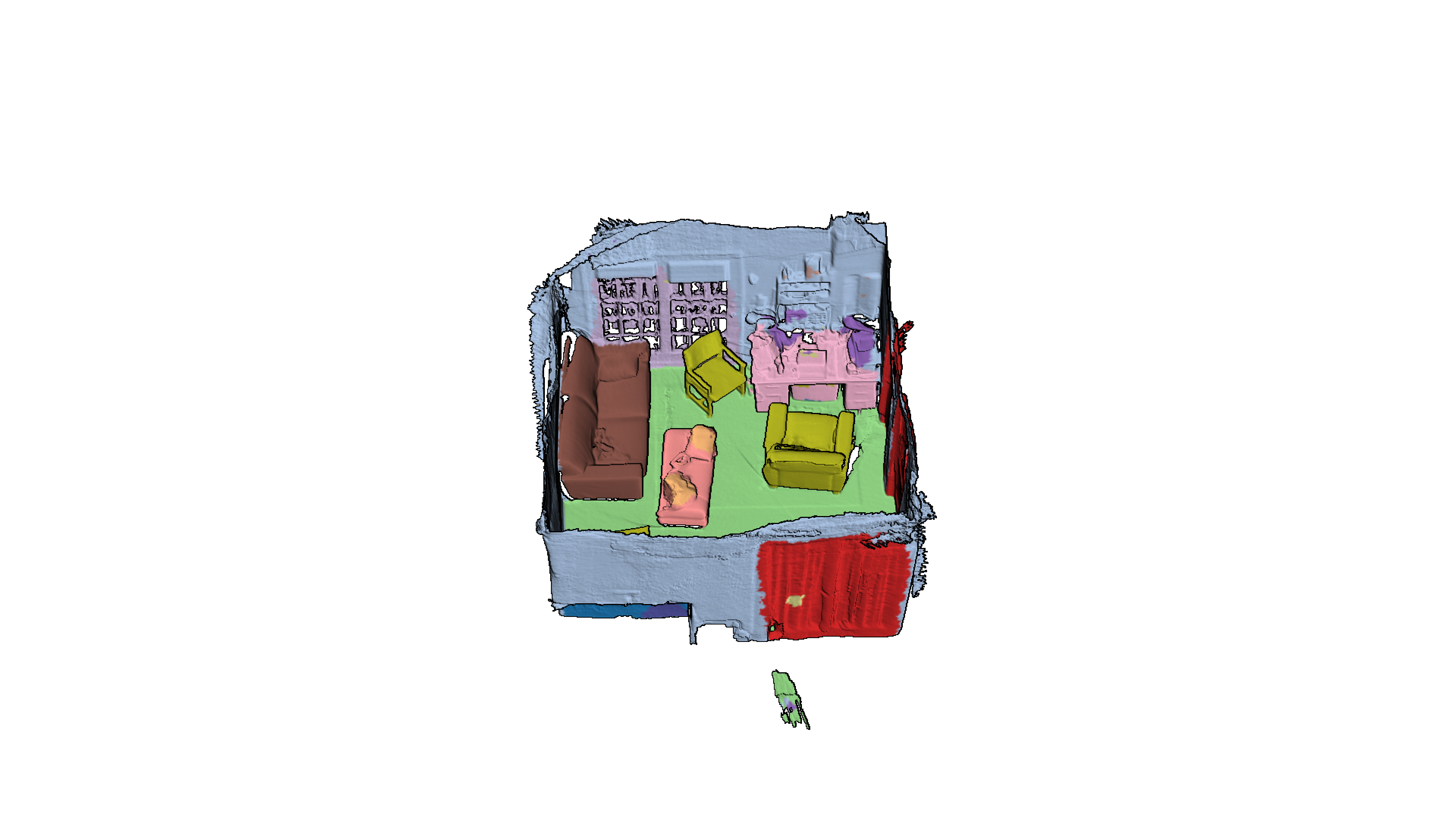}\\
		\end{tabular}
		\caption{ScanNet results. The left image shows the ground truth and the right one our prediction.}
		\label{fig:ScanNetImages}
	\end{figure}
	\egroup

	\bgroup
	\def\MethodW{3.1cm}
	\def\W{2cm}
	\small
	\begin{table*}[]
		\caption{Results on ShapeNet part segmentation~\cite{yi2016scalable}.}
		\centering
		\setlength{\tabcolsep}{2.4pt}
		\begin{tabularx}{\textwidth}{C{\MethodW}ccccccccccccccccc}
			\toprule
			\#instances                         &                                                                     & 2690      & 76   & 55   & 898  & 3758  & 69       & 787    & 392   & 1547 & 451    & 202        & 184  & 283    & 66     & 152        & 5271  \\ \hline
			\multicolumn{1}{l|}{}     & \multicolumn{1}{c|}{instance} & \small{air-} & \small{bag}  & \small{cap}  & \small{car}  & \small{chair} & \small{ear-} & \small{guitar} & \small{knife} & \small{lamp} & \small{laptop} & \small{motor-} & \small{mug}  & \small{pistol} & \small{rocket} & \small{skate-} & \small{table} \\
			\multicolumn{1}{l|}{}               
			& \multicolumn{1}{c|}{avg.} & \small{plane} & & & & & \small{phone} & & & & & \small{bike} & & & & \small{board} & \\
			\midrule
			\multicolumn{1}{l|}{PointNet{~\cite{qi2017pointnet}}}   
			& \multicolumn{1}{c|}{83.7}         & 83.4      & 78.7 & 82.5 & 74.9 & 89.6  & 73.0     & 91.5   & 85.9  & 80.8 & 95.3   & 65.2       & 93.0 & 81.2   & 57.9   & 72.8       & 80.6  \\
			\multicolumn{1}{l|}{PointNet++{~\cite{qi2017pointnet++}}}      
			 & \multicolumn{1}{c|}{85.1}         & 82.4      & 79.0 & 87.7 & 77.3 & \textbf{90.8}  & 71.8     & 91.0   & 85.9  & 83.7 & 95.3   & 71.6       & 94.1 & 81.3   & 58.7   & 76.4       & 82.6   \\
			\multicolumn{1}{l|}{SplatNet 3D{~\cite{su2018splatnet}}}       
			& \multicolumn{1}{c|}{84.6}         & 81.9 & 83.9 & 88.6 & 79.5 & 90.1 & 73.5 & 91.3 & 84.7 & 84.5 & 96.3 & 69.7 & 95.0 & 81.7 & 59.2 & 70.4 & 81.3  \\
			\multicolumn{1}{l|}{SplatNet 2D-3D{~\cite{su2018splatnet}}}      
			 & \multicolumn{1}{c|}{\textbf{85.4}}         & 83.2 & 84.3 & \textbf{89.1} & 80.3 & 90.7 & \textbf{75.5} & 92.1 & 87.1 & 83.9 & 96.3 & 75.6 & 95.8 & 83.8 & 64.0 & 75.5 & 81.8  \\
			\multicolumn{1}{l|}{FCPN{~\cite{rethage2018fully}}}       
			& \multicolumn{1}{c|}{84.0}         & \textbf{84.0}       & 82.8 & 86.4 & \textbf{88.3} & 83.3  & 73.6     & \textbf{93.4}   & 87.4  & 77.4 & \textbf{97.7}   & \textbf{81.4}       & 95.8 & \textbf{87.7}   & 68.4   & \textbf{83.6}       & 73.4  \\
			\midrule
			\multicolumn{1}{l|}{Ours}       
			& \multicolumn{1}{c|}{83.9}         & 82.3      & \textbf{84.8} & 79.1 & 81.0 & 86.9  & 71.0     & 91.9   & \textbf{89.4}  & \textbf{84.7} & 96.6   & 77.2       & \textbf{95.8} & 86.0   & \textbf{70.5}   & 79.3       & \textbf{87.0}  \\
			\bottomrule
		\end{tabularx}
		\label{tab:shapenet}
	\end{table*}
	\egroup
	
	\bgroup
	\newcolumntype{Q}{R{0.315cm}}
	\begin{table*}[]
		\scriptsize
		\caption{Results on SemanticKITTI~\cite{behley2019semantickitti}.}
		\begin{tabularx}{\textwidth}{L{2.1cm}c|QQQQQQQQQQQQQQQQQQQ}
			\toprule 
			Approach & \begin{sideways}mIoU\end{sideways} & \begin{sideways}road\end{sideways} & \begin{sideways}sidewalk\end{sideways} & \begin{sideways}parking\end{sideways} & \begin{sideways}other-ground\end{sideways} & \begin{sideways}building\end{sideways} & \begin{sideways}car\end{sideways} & \begin{sideways}truck\end{sideways} & \begin{sideways}bicycle\end{sideways} & \begin{sideways}motorcycle\end{sideways} & \begin{sideways}other-vehicle\end{sideways} & \begin{sideways}vegetation\end{sideways} & \begin{sideways}trunk\end{sideways} & \begin{sideways}terrain\end{sideways} & \begin{sideways}person\end{sideways} & \begin{sideways}bicyclist\end{sideways} & \begin{sideways}motorcyclist\end{sideways} & \begin{sideways}fence\end{sideways} & \begin{sideways}pole\end{sideways} & \begin{sideways}traffic sign\end{sideways}\\
			\midrule
			PointNet \cite{qi2017pointnet}  & 14.6 & 61.6 & 35.7 & 15.8 & \ph1.4 & 41.4 & 46.3 & \ph0.1 & \ph1.3 & \ph0.3 & \ph0.8 & 31.0 & \ph4.6 & 17.6 & \ph0.2 & \ph0.2 & \ph0.0 & 12.9 & \ph2.4 & \ph3.7\\
			SplatNet \cite{su2018splatnet}  & 18.4 & 64.6 & 39.1 & \ph0.4 & \ph0.0 & 58.3 & 58.2 & \ph0.0 & \ph0.0 & \ph0.0 & \ph0.0 & 71.1 & \ph9.9 & 19.3 & \ph0.0 & \ph0.0 & \ph0.0 & 23.1 & \ph5.6 & \ph0.0\\
			PointNet++ \cite{qi2017pointnet++}  & 20.1 & 72.0 & 41.8 & 18.7 & \ph5.6 & 62.3 & 53.7 & \ph0.9 & \ph1.9 & \ph0.2 & \ph0.2 & 46.5 & 13.8 & 30.0 & \ph0.9 & \ph1.0 & \ph0.0 & 16.9 & \ph6.0 & \ph8.9\\
			Minkowski34(25cm)~\cite{choy20194d}  & 33.0 & 80.8 & 43.0 & 36.9 & \ph0.5 & 73.5 & 83.0 & \textbf{42.9} & \ph2.0 & \ph2.9 & \ph7.8 & 74.4 & 42.9 & 36.7 & 11.2 & 22.8 & \ph4.4 & 37.2 & 35.4& 28.6\\
			SqueezeSegV2 \cite{wu2018squeezesegv2}  & 39.7 & 88.6 & 67.6 & 45.8 & 17.7 & 73.7 & 81.8 & 13.4 & 18.5 & 17.9 & 14.0 & 71.8 & 35.8 & 60.2 & 20.1 & 25.1 & \ph3.9 & 41.1 & 20.2 & 36.3\\
			TangentConv \cite{tatarchenko2018tangent}  & 40.9 & 83.9 & 63.9 & 33.4 & 15.4 & 83.4 & 90.8 & 15.2 & \ph2.7 & 16.5 & 12.1 & 79.5 & 49.3 & 58.1 & 23.0 & 28.4 & \ph8.1 & 49.0 & 35.8 & 28.5\\
			DarkNet21Seg \cite{behley2019semantickitti}  & 47.4 & 91.4 & 74.0 & 57.0 & 26.4 & 81.9 & 85.4 & 18.6 & \textbf{26.2} & 26.5 & 15.6 & 77.6 & 48.4 & 63.6 & 31.8 & 33.6 & \ph4.0 & 52.3 & 36.0 & 50.0\\
			DarkNet53Seg \cite{behley2019semantickitti} & 49.9 & \textbf{91.8} & \textbf{74.6} & \textbf{64.8} & \textbf{27.9} & 84.1 & 86.4 & 25.5 & 24.5 & \textbf{32.7} & \textbf{22.6} & 78.3 & 50.1 & \textbf{64.0} & \textbf{36.2} & 33.6 & \ph4.7 & 55.0 & 38.9 & \textbf{52.2}\\
			\midrule

			Ours  & \textbf{52.9} & 90.0 & 74.1 & 59.4 & 22.0 & \textbf{88.2} & \textbf{92.9} & 26.6 & 16.6 & 22.2 & 21.4 & \textbf{81.7} & \textbf{63.6} & 63.1 & 35.6 & \textbf{43.0} & \textbf{46.0} & \textbf{58.8} & \textbf{51.9} & 48.4\\
			\bottomrule
		\end{tabularx}
		\label{tab:semanticKitti}
	\end{table*}
	\egroup

	\setlength{\belowcaptionskip}{+4pt}
	\bgroup
	\newcolumntype{Q}{L{0.33cm}}
		\tabcolsep= 1mm
		\begin{table}[]
			\centering
			\small
			\setlength{\tabcolsep}{2pt}
			\caption{Results on ScanNet~\cite{dai2017scannet}}
			\resizebox{0.5\columnwidth}{!}{
				\begin{tabular}{r|c}
					\toprule
					\small{Method}  & \small{mIOU} \\
					\midrule
					PointNet++~\cite{qi2017pointnet++} & 33.9\\
					SplatNet~\cite{su2018splatnet} & 39.3 \\
					TangetConv~\cite{tatarchenko2018tangent} & 43.8 \\
					3DMV${}^\ddagger$~\cite{dai20183dmv} & 48.4 \\
					MinkowskiNet42 (5cm)~\cite{choy20194d} & 67.9 \\
					SparseConvNet~\cite{graham20183d}${}^\dagger$ & 72.5 \\
					MinkowskiNet42 (2cm)~\cite{choy20194d}${}^\dagger$ & \textbf{73.4} \\
					\midrule
					Ours & 64.0 \\
					\bottomrule
			\end{tabular}}
			\label{tab:scannet}
			\hspace{4pt}
			\caption*{\small{${}^\dagger$: post-CVPR submissions. ${}^\ddagger$: uses 2D images additionally.}}
	\end{table}
	\bgroup

	\bgroup
	\setlength\tabcolsep{4.5pt}\begin{table}[t]
			\centering
			\footnotesize
			\setlength{\tabcolsep}{3.0pt}
			\caption{Motion segmentation IoU results on SemanticKITTI~\cite{behley2019semantickitti} using a sequence of multiple past scans (in \%). Shaded cells correspond to the IoU of the moving classes, while unshaded entries are the non-moving classes.}
			{
			\begin{tabular}{lcccccc|c}
				\toprule 
				Approach & \begin{sideways}car\end{sideways} & \begin{sideways}truck\end{sideways} & \begin{sideways}other-vehicle\end{sideways} & \begin{sideways}person\end{sideways} & \begin{sideways}bicyclist\end{sideways} & \begin{sideways}motorcyclist\end{sideways} & \begin{sideways}mIoU\end{sideways}\\
				\midrule
				\multirow{2}{*}{TangentConv \cite{tatarchenko2018tangent}}  & 84.9 & 21.1 & 18.5 & 1.6 & 0.0 & 0.0 & \multirow{2}{*}{ 34.1}\\
				& \cellcolor{Gainsboro!90} 40.3 & \cellcolor{Gainsboro!90} 42.2 & \cellcolor{Gainsboro!90} 30.1 & \cellcolor{Gainsboro!90} 6.4 & \cellcolor{Gainsboro!90} 1.1 & \cellcolor{Gainsboro!90} 1.9 & \\
				
				\rule{0pt}{0.5cm}
				\multirow{2}{*}{DarkNet53Seg \cite{behley2019semantickitti}}  & 84.1 & 20.0 & 20.7 & 7.5 & 0.0 & 0.0 & \multirow{2}{*}{ 41.6}\\
				& \cellcolor{Gainsboro!90} 61.5 & \cellcolor{Gainsboro!90} 37.8 & \cellcolor{Gainsboro!90} 28.9 & \cellcolor{Gainsboro!90} 15.2 & \cellcolor{Gainsboro!90} 14.1 & \cellcolor{Gainsboro!90} 0.2 & \\
				
				\rule{0pt}{0.5cm}
				\multirow{2}{*}{SpSequenceNet \cite{shi2020spsequencenet}}  & 88.5 & 29.2 & 22.7 & 6.3 & 0.0 & 0.0 & \multirow{2}{*}{ 43.1}\\
				& \cellcolor{Gainsboro!90} 53.2 & \cellcolor{Gainsboro!90} 0.1 & \cellcolor{Gainsboro!90} 2.3 & \cellcolor{Gainsboro!90} 26.2 & \cellcolor{Gainsboro!90} 41.2 & \cellcolor{Gainsboro!90} 36.2 & \\
				
				\rule{0pt}{0.5cm}
				\multirow{2}{*}{KPConv \cite{thomas2019kpconv}}  & 93.7 & 70.3 & 38.6 & 21.6 & 0.0 & 0.0 & \multirow{2}{*}{ 51.2}\\
				& \cellcolor{Gainsboro!90} 69.4 & \cellcolor{Gainsboro!90} 5.8 & \cellcolor{Gainsboro!90} 4.7 & \cellcolor{Gainsboro!90} 67.5 & \cellcolor{Gainsboro!90} 67.4 & \cellcolor{Gainsboro!90} 47.2 & \\
				
				\midrule
				
				
				\multirow{2}{*}{ Ours }  & 91.1 & 65.4 & 23.1 & 6.8 & 0.0 & 0.0 & \multirow{2}{*}{ 45.2}\\
				& \cellcolor{Gainsboro!90} 54.8 & \cellcolor{Gainsboro!90} 3.5 & \cellcolor{Gainsboro!90} 0.6 & \cellcolor{Gainsboro!90} 49.9 & \cellcolor{Gainsboro!90} 44.6 & \cellcolor{Gainsboro!90} 64.3 & \\
				
				\bottomrule
		\end{tabular}}
		\setlength\tabcolsep{6.0pt}
		\label{tab:results_multiscan}\vspace{-0.2cm}
	\end{table}
	\egroup

	\begin{table}[t]
		\centering
		\caption{Instance segmentation performance on the maize and tomato plants of the Pheno4D dataset.}
		\begin{tabular}{ ccc }
			& \multicolumn{2}{c}{SBD}  \\ 
			& Maize & Tomato \\
			\toprule
			PointNet\cite{qi2017pointnet} & 69.7  & 47.3\\  
			PointNet++\cite{qi2017pointnet++} & 74.8 & 56.1\\   
			LatticeNet (ours) & \textbf{80.6} &  \textbf{74.2}\\
			\midrule
		\end{tabular}
		\label{tab:sbd}
	\end{table}

	\subsection{Ablation Studies}
	We perform various ablations regarding our contribution to judge how much they affect the network's performance. 
	 
	\noindent\textbf{DeformSlice}: We assess the impact that DeformSlice has on the network by comparing it with the Slice operator which does not use learned barycentric interpolation. We evaluate this on SemanticKITTI, the largest dataset that we are using.
	
	We also evaluate a version of DeformSlice which ensures that the new barycentric coordinates still sum up to one by adding an additional loss term:
	\begin{align}
	L= \frac{1}{\abs{P}} \sum_{p\in P} \left( \sum_{v\in I_p} \Delta b_{pv} \right)^2.
	\end{align}
	However, we observe little change after adding this regularization term and hence, use the default version of DeformSlice for the rest of the experiments.
	The results are gathered in \reftab{tab:AblationTalbe}.
	
	
	\noindent\textbf{Distribute and PointNet}: Another contribution of our work is the usage of a Distribute operator to provide values to the lattice vertices which are later embedded in a higher-dimensional space by a PointNet-like architecture. The positions and features of the point cloud are treated separately where the features (normals, color) are distributed directly. From the positions, we substract the locally averaged position as we assume that the local point distribution is more important than the coordinates in the global reference frame.
	We evaluate the impact of elevating the point features to a higher-dimensional space and subtracting the local mean against a simple splatting operator which just averages the features of the points around each corresponding vertex.  
	

	
	We observe that not subtracting the local mean, and just using the $xyz$ coordinates as features, heavily degrades the performance, causing the mIoU to drop from \num{52.9} to \num{43.0}. This further reinforces the idea that the local point distribution is a good local feature to use in the first layers of the network.
	
	Not elevating the point cloud features to a higher-dimensional space before applying the max-pool operation also hurts performance but not as severely. In our experiments, we elevate the features to \num{64} dimensions by using a series of fully connected layers.
	
	Finally, naive application of the splat operation performs worst with a mere \num{37.8}~mIoU.

	\subsection{Performance}
	We report the time taken for a forward pass and the maximum memory used in our shallow and deep network on the first three evaluated datasets. The performance was measured on a NVIDIA Titan X Pascal and the results are gathered in \reftab{tab:perf-results}. 
	
	In the case of motion segmentation, the inference times and memory used are the same as in the case of a single scan, as we use the same backbone network to extract features and the computational cost of fusing the temporal information is minimum. However for training, the network requires more memory with increasing time window due to the back-propagation through time. This scales linearly with the time window size and the amount of points in the cloud.
	
	Despite the reduced memory usage compared to SplatNet and increased speed of execution, there are still memory savings possible by fusing the Distribute and PointNet operators into one GPU operation. This is similar to fusing our DeformSlice and the classification layer. Additionally, we expect the network to become even faster as further advances on highly optimized kernels for convolution on sparse lattices become available. At the moment, the convolutions are performed by our custom CUDA kernels. Tighter integration however with highly optimized libraries like cuDNN~\cite{chetlur2014cudnn} could be beneficial.
	\bgroup
	\newcolumntype{M}{L{1.3cm}} 
	\newcolumntype{Q}{R{1.0cm}}
	\newcolumntype{P}{L{1.0cm}}
		\begin{table}[]
		\scriptsize
		\caption{Average time used by the forward pass and the maximum memory used during training. An X indicates a method that failed to process the whole cloud due to memory limitations. }
		\centering
		\begin{tabularx}{\columnwidth}{M|QP|QP|QP}
			\toprule
			         & \multicolumn{2}{c|}{ShapeNet} & \multicolumn{2}{c|}{ScanNet} & \multicolumn{2}{c}{SemanticKITTI}  \\ 
			         & \multicolumn{1}{r}{$\left[\SI{}{\milli\second}\right]$} & \multicolumn{1}{l|}{$\left[\SI{}{\giga\byte}\right]$}
			         & \multicolumn{1}{r}{$\left[\SI{}{\milli\second}\right]$} & \multicolumn{1}{l|}{$\left[\SI{}{\giga\byte}\right]$}
			         & \multicolumn{1}{r}{$\left[\SI{}{\milli\second}\right]$} & \multicolumn{1}{l}{$\left[\SI{}{\giga\byte}\right]$}\\
			         \midrule
			SplatNet & 129\ph&\ph0.6 & X\ph&\ph X & 2931\ph&\ph8.9\\ 
			\midrule 
			Ours & 49\ph&\ph0.5 & 180\ph&\ph6.5  & 143\ph&\ph3.5\\
			
			\bottomrule
 		\end{tabularx}
		\label{tab:perf-results}
	\end{table}
	\egroup

\bgroup
\definecolor{abl-green}{RGB}{49, 158, 78}
\definecolor{abl-red}{RGB}{204, 27, 24}
\newcolumntype{Q}{C{0.5cm}}
\begin{table}[]
	\centering
	\caption{Ablation study of the various components of LatticeNet. Various features are disabled (indicated in red) and the impact to the IoU is evaluated.}
	\begin{tabularx}{0.76\columnwidth}{L{3.0cm}c|Q|Q|Q|Q|Q}
		\hhline{|-------|}
		& \multicolumn{1}{c}{\begin{sideways}mIoU\end{sideways}} & \multicolumn{1}{Q}{\begin{sideways} LocalAvg\end{sideways}} & \multicolumn{1}{Q}{\begin{sideways} PointNet elevate\end{sideways}} & \multicolumn{1}{Q}{\begin{sideways} DeformSlice\end{sideways}} & \multicolumn{1}{Q}{\begin{sideways} Offsets zero sum\end{sideways}} \\
		
		\hhline{|-------|}
		
		LN splat & 37.8  & \cellcolor{abl-red} &  \cellcolor{abl-red} & \cellcolor{abl-green}  & \cellcolor{abl-red} \\
		\hhline{|~~-----|}
		LN no local avg & 43.0  & \cellcolor{abl-red} &  \cellcolor{abl-green} & \cellcolor{abl-green}  & \cellcolor{abl-red}\\
		\hhline{|~~-----|}
		LN no elevate & 46.8  & \cellcolor{abl-green} &  \cellcolor{abl-red} & \cellcolor{abl-green}  & \cellcolor{abl-red}\\
		\hhline{|~~-----|}
		LN slice & 50.4  & \cellcolor{abl-green} &  \cellcolor{abl-green} & \cellcolor{abl-red}  & \cellcolor{abl-red}\\
		\hhline{|~~-----|}
		LN reg & 52.7  & \cellcolor{abl-green} &  \cellcolor{abl-green} & \cellcolor{abl-green}  & \cellcolor{abl-green} \\
		\hhline{|-------|}
		LatticeNet & \textbf{52.9}  & \cellcolor{abl-green} &  \cellcolor{abl-green} & \cellcolor{abl-green}  & \cellcolor{abl-red} \\
		\hhline{|-------|}
	\end{tabularx}
	\label{tab:AblationTalbe}
\end{table}
\egroup
 
\section{Conclusion}
We presented LatticeNet, a novel method for point cloud segmentation. A sparse permutohedral lattice allows us to efficiently process large point clouds. The usage of PointNet together with a data-dependent interpolation alleviates the quantization issues of other methods. Experiments on four datasets show state-of-the-art results, at a reduced time and memory budget.

\bibliographystyle{plainnat}
\interlinepenalty=10000 
\bibliography{references}

\begin{thebibliography}{53}
\providecommand{\natexlab}[1]{#1}
\providecommand{\url}[1]{\texttt{#1}}
\expandafter\ifx\csname urlstyle\endcsname\relax
  \providecommand{\doi}[1]{doi: #1}\else
  \providecommand{\doi}{doi: \begingroup \urlstyle{rm}\Url}\fi

\bibitem[phe()]{pheno4d}
A large scale spatio-temporal dataset of point clouds of maize and tomato
  plants.
\newblock \url{https://www.ipb.uni-bonn.de/data/pheno4d/}.
\newblock Accessed: 2021-01-1.

\bibitem[Baek and Adams(2009)]{baek2009some}
Jongmin Baek and Andrew Adams.
\newblock
  \href{https://graphics.stanford.edu/papers/permutohedral/permutohedral_techreport.pdf}{Some
  useful properties of the permutohedral lattice for {Gaussian} filtering}.
\newblock \emph{In other words}, 10\penalty0 (1):\penalty0 0, 2009.

\bibitem[Barron et~al.(2015)Barron, Adams, YiChang, and
  Hern{\'a}ndez]{barron2015fast}
Jonathan~T Barron, Andrew Adams, S~YiChang, and Carlos Hern{\'a}ndez.
\newblock \href{https://andrew.adams.pub/BarronCVPR2015_supp.pdf} {Fast
  bilateral-space stereo for synthetic defocus — Supplemental material}.
\newblock In \emph{Proc.~of the IEEE Conference on Computer Vision and Pattern
  Recognition (CVPR)}, pages 1--15, 2015.

\bibitem[Behley et~al.(2019)Behley, Garbade, Milioto, Quenzel, Behnke,
  Stachniss, and Gall]{behley2019semantickitti}
Jens Behley, Martin Garbade, Andres Milioto, Jan Quenzel, Sven Behnke, Cyrill
  Stachniss, and Juergen Gall.
\newblock \href{https://arxiv.org/abs/1904.01416} {{SemanticKITTI}: A Dataset
  for Semantic Scene Understanding of LiDAR Sequences}.
\newblock In \emph{Proc.~of the IEEE Int.~Conference on Computer Vision
  (ICCV)}, 2019.

\bibitem[Berman et~al.(2018)Berman, Triki, and Blaschko]{berman2018lovasz}
Maxim Berman, Amal~Rannen Triki, and Matthew~B Blaschko.
\newblock \href{https://arxiv.org/abs/1705.08790}{The Lov{\'a}sz-softmax loss:
  A tractable surrogate for the optimization of the intersection-over-union
  measure in neural networks}.
\newblock In \emph{Proc.~of the IEEE Conference on Computer Vision and Pattern
  Recognition (CVPR)}, pages 4413--4421, 2018.

\bibitem[Chen et~al.(2017)Chen, Papandreou, Schroff, and
  Adam]{chen2017rethinking}
Liang-Chieh Chen, George Papandreou, Florian Schroff, and Hartwig Adam.
\newblock \href{https://arxiv.org/abs/1706.05587}{Rethinking atrous convolution
  for semantic image segmentation}.
\newblock \emph{arXiv preprint arXiv:1706.05587}, 2017.

\bibitem[Chen et~al.(2018)Chen, Han, Li, Chen, Xing, Zhao, and
  Li]{chen2018deep}
Weikai Chen, Xiaoguang Han, Guanbin Li, Chao Chen, Jun Xing, Yajie Zhao, and
  Hao Li.
\newblock \href{https://arxiv.org/abs/1812.04302} {Deep {RBFNet}: Point cloud
  feature learning using radial basis functions}.
\newblock \emph{arXiv preprint arXiv:1812.04302}, 2018.

\bibitem[Chetlur et~al.(2014)Chetlur, Woolley, Vandermersch, Cohen, Tran,
  Catanzaro, and Shelhamer]{chetlur2014cudnn}
Sharan Chetlur, Cliff Woolley, Philippe Vandermersch, Jonathan Cohen, John
  Tran, Bryan Catanzaro, and Evan Shelhamer.
\newblock {cuDNN}: Efficient primitives for deep learning.
\newblock \emph{arXiv preprint arXiv:1410.0759}, 2014.

\bibitem[Choy et~al.(2019)Choy, Gwak, and Savarese]{choy20194d}
Christopher Choy, JunYoung Gwak, and Silvio Savarese.
\newblock \href{https://arxiv.org/abs/1904.08755}{4D Spatio-Temporal ConvNets:
  Minkowski Convolutional Neural Networks}.
\newblock \emph{arXiv preprint arXiv:1904.08755}, 2019.

\bibitem[Dai and Nie{\ss}ner(2018)]{dai20183dmv}
Angela Dai and Matthias Nie{\ss}ner.
\newblock \href{https://arxiv.org/abs/1803.10409} {{3DMV}: Joint
  {3D}-multi-view prediction for {3D} semantic scene segmentation}.
\newblock In \emph{Proc.~of the European Conference on Computer Vision (ECCV)},
  pages 452--468, 2018.

\bibitem[Dai et~al.(2017)Dai, Chang, Savva, Halber, Funkhouser, and
  Nie{\ss}ner]{dai2017scannet}
Angela Dai, Angel~X Chang, Manolis Savva, Maciej Halber, Thomas Funkhouser, and
  Matthias Nie{\ss}ner.
\newblock \href{https://arxiv.org/abs/1702.04405} {{ScanNet}: Richly-annotated
  {3D} reconstructions of indoor scenes}.
\newblock In \emph{Proc.~of the IEEE Conference on Computer Vision and Pattern
  Recognition (CVPR)}, pages 5828--5839, 2017.

\bibitem[De~Brabandere et~al.(2017)De~Brabandere, Neven, and
  Van~Gool]{de2017semantic}
Bert De~Brabandere, Davy Neven, and Luc Van~Gool.
\newblock \href{https://arxiv.org/abs/1708.02551}{Semantic instance
  segmentation with a discriminative loss function}.
\newblock \emph{arXiv preprint arXiv:1708.02551}, 2017.

\bibitem[Defferrard et~al.(2016)Defferrard, Bresson, and
  Vandergheynst]{defferrard2016convolutional}
Micha{\"e}l Defferrard, Xavier Bresson, and Pierre Vandergheynst.
\newblock
  \href{https://papers.nips.cc/paper/6081-convolutional-neural-networks-on-graphs-with-fast-localized-spectral-filtering.pdf}
  {Convolutional neural networks on graphs with fast localized spectral
  filtering}.
\newblock In \emph{Proc.~of the Advances in Neural Information Processing
  Systems (NIPS)}, pages 3844--3852, 2016.

\bibitem[Graham et~al.(2018)Graham, Engelcke, and van~der Maaten]{graham20183d}
Benjamin Graham, Martin Engelcke, and Laurens van~der Maaten.
\newblock
  \href{http://openaccess.thecvf.com/content_cvpr_2018/papers/Graham_3D_Semantic_Segmentation_CVPR_2018_paper.pdf}{{3D}
  semantic segmentation with submanifold sparse convolutional networks}.
\newblock In \emph{Proc.~of the IEEE Conference on Computer Vision and Pattern
  Recognition (CVPR)}, pages 9224--9232, 2018.

\bibitem[Gu et~al.(2019)Gu, Wang, Wu, Lee, and Wang]{gu2019hplflownet}
Xiuye Gu, Yijie Wang, Chongruo Wu, Yong~Jae Lee, and Panqu Wang.
\newblock \href{https://arxiv.org/abs/1906.05332}{{HPLFlowNet}: Hierarchical
  Permutohedral Lattice {FlowNet} for Scene Flow Estimation on Large-scale
  Point Clouds}.
\newblock In \emph{Proc.~of the IEEE Conference on Computer Vision and Pattern
  Recognition (CVPR)}, pages 3254--3263, 2019.

\bibitem[He et~al.(2016{\natexlab{a}})He, Zhang, Ren, and Sun]{he2016deep}
Kaiming He, Xiangyu Zhang, Shaoqing Ren, and Jian Sun.
\newblock
  \href{https://www.cv-foundation.org/openaccess/content_cvpr_2016/papers/He_Deep_Residual_Learning_CVPR_2016_paper.pdf}{Deep
  residual learning for image recognition}.
\newblock In \emph{Proc.~of the IEEE Conference on Computer Vision and Pattern
  Recognition (CVPR)}, pages 770--778, 2016{\natexlab{a}}.

\bibitem[He et~al.(2016{\natexlab{b}})He, Zhang, Ren, and Sun]{he2016identity}
Kaiming He, Xiangyu Zhang, Shaoqing Ren, and Jian Sun.
\newblock \href{https://arxiv.org/abs/1603.05027}{Identity mappings in deep
  residual networks}.
\newblock In \emph{Proc.~of the European Conference on Computer Vision (ECCV)},
  pages 630--645, 2016{\natexlab{b}}.

\bibitem[Huang et~al.(2017)Huang, Liu, Van Der~Maaten, and
  Weinberger]{huang2017densely}
Gao Huang, Zhuang Liu, Laurens Van Der~Maaten, and Kilian~Q Weinberger.
\newblock
  \href{http://openaccess.thecvf.com/content_cvpr_2017/papers/Huang_Densely_Connected_Convolutional_CVPR_2017_paper.pdf}{Densely
  connected convolutional networks}.
\newblock In \emph{Proc.~of the IEEE Conference on Computer Vision and Pattern
  Recognition (CVPR)}, pages 4700--4708, 2017.

\bibitem[Huang et~al.(2019)Huang, Zhang, Yi, Funkhouser, Nie{\ss}ner, and
  Guibas]{huang2019texturenet}
Jingwei Huang, Haotian Zhang, Li~Yi, Thomas Funkhouser, Matthias Nie{\ss}ner,
  and Leonidas~J Guibas.
\newblock
  \href{http://openaccess.thecvf.com/content_CVPR_2019/papers/Huang_TextureNet_Consistent_Local_Parametrizations_for_Learning_From_High-Resolution_Signals_on_CVPR_2019_paper.pdf}
  {{TextureNet}: Consistent local parametrizations for learning from
  high-resolution signals on meshes}.
\newblock In \emph{Proc.~of the IEEE Conference on Computer Vision and Pattern
  Recognition (CVPR)}, pages 4440--4449, 2019.

\bibitem[Li et~al.(2018)Li, Bu, Sun, Wu, Di, and Chen]{li2018pointcnn}
Yangyan Li, Rui Bu, Mingchao Sun, Wei Wu, Xinhan Di, and Baoquan Chen.
\newblock
  \href{https://papers.nips.cc/paper/7362-pointcnn-convolution-on-x-transformed-points.pdf}
  {{PointCNN}: Convolution on x-transformed points}.
\newblock In \emph{Proc.~of the Advances in Neural Information Processing
  Systems (NIPS)}, pages 820--830, 2018.

\bibitem[Lin et~al.(2017)Lin, Milan, Shen, and Reid]{lin2017refinenet}
Guosheng Lin, Anton Milan, Chunhua Shen, and Ian Reid.
\newblock
  \href{http://openaccess.thecvf.com/content_cvpr_2017/papers/Lin_RefineNet_Multi-Path_Refinement_CVPR_2017_paper.pdf}{{RefineNet}:
  Multi-path refinement networks for high-resolution semantic segmentation}.
\newblock In \emph{Proc.~of the IEEE Conference on Computer Vision and Pattern
  Recognition (CVPR)}, pages 1925--1934, 2017.

\bibitem[Long et~al.(2015)Long, Shelhamer, and Darrell]{long2015fully}
Jonathan Long, Evan Shelhamer, and Trevor Darrell.
\newblock
  \href{https://people.eecs.berkeley.edu/~jonlong/long_shelhamer_fcn.pdf}{Fully
  convolutional networks for semantic segmentation}.
\newblock In \emph{Proc.~of the IEEE Conference on Computer Vision and Pattern
  Recognition (CVPR)}, pages 3431--3440, 2015.

\bibitem[Masci et~al.(2015)Masci, Boscaini, Bronstein, and
  Vandergheynst]{masci2015geodesic}
Jonathan Masci, Davide Boscaini, Michael Bronstein, and Pierre Vandergheynst.
\newblock
  \href{https://www.cv-foundation.org/openaccess/content_iccv_2015_workshops/w22/papers/Masci_Geodesic_Convolutional_Neural_ICCV_2015_paper.pdf}
  {Geodesic convolutional neural networks on {Riemannian} manifolds}.
\newblock In \emph{Workshop Proc.~of the IEEE Int.~Conference on Computer
  Vision (ICCV Workshops)}, pages 37--45, 2015.

\bibitem[Monti et~al.(2017)Monti, Boscaini, Masci, Rodola, Svoboda, and
  Bronstein]{monti2017geometric}
Federico Monti, Davide Boscaini, Jonathan Masci, Emanuele Rodola, Jan Svoboda,
  and Michael~M Bronstein.
\newblock \href{https://arxiv.org/abs/1611.08402} {Geometric deep learning on
  graphs and manifolds using mixture model {CNNs}}.
\newblock In \emph{Proc.~of the IEEE Conference on Computer Vision and Pattern
  Recognition (CVPR)}, pages 5115--5124, 2017.

\bibitem[Neven et~al.(2019)Neven, Brabandere, Proesmans, and
  Gool]{neven2019instance}
Davy Neven, Bert~De Brabandere, Marc Proesmans, and Luc~Van Gool.
\newblock \href{https://arxiv.org/abs/1906.11109}{Instance segmentation by
  jointly optimizing spatial embeddings and clustering bandwidth}.
\newblock In \emph{Proc.~of the IEEE Conference on Computer Vision and Pattern
  Recognition (CVPR)}, pages 8837--8845, 2019.

\bibitem[Nie{\ss}ner et~al.(2013)Nie{\ss}ner, Zollh{\"o}fer, Izadi, and
  Stamminger]{niessner2013real}
Matthias Nie{\ss}ner, Michael Zollh{\"o}fer, Shahram Izadi, and Marc
  Stamminger.
\newblock
  \href{https://niessnerlab.org/papers/2013/4hashing/niessner2013hashing.pdf}{Real-time
  {3D} reconstruction at scale using voxel hashing}.
\newblock \emph{ACM Transactions on Graphics (ToG)}, 32\penalty0 (6):\penalty0
  1--11, 2013.

\bibitem[Paszke et~al.(2017)Paszke, Gross, Chintala, Chanan, Yang, DeVito, Lin,
  Desmaison, Antiga, and Lerer]{paszke2017automatic}
Adam Paszke, Sam Gross, Soumith Chintala, Gregory Chanan, Edward Yang, Zachary
  DeVito, Zeming Lin, Alban Desmaison, Luca Antiga, and Adam Lerer.
\newblock
  \href{https://pdfs.semanticscholar.org/b36a/5bb1707bb9c70025294b3a310138aae8327a.pdf}
  {Automatic Differentiation in {PyTorch}}.
\newblock In \emph{NIPS Autodiff Workshop}, 2017.

\bibitem[Pham et~al.(2019{\natexlab{a}})Pham, Hua, Nguyen, and
  Yeung]{pham2019real}
Quang-Hieu Pham, Binh-Son Hua, Thanh Nguyen, and Sai-Kit Yeung.
\newblock \href{https://arxiv.org/abs/1804.00257}{Real-time progressive {3D}
  semantic segmentation for indoor scenes}.
\newblock In \emph{Proc.~of the IEEE Workshop on Applications of Computer
  Vision}, pages 1089--1098, 2019{\natexlab{a}}.

\bibitem[Pham et~al.(2019{\natexlab{b}})Pham, Nguyen, Hua, Roig, and
  Yeung]{pham2019jsis3d}
Quang-Hieu Pham, Thanh Nguyen, Binh-Son Hua, Gemma Roig, and Sai-Kit Yeung.
\newblock \href{https://arxiv.org/abs/1904.00699}{{JSIS3D}: Joint
  semantic-instance segmentation of {3D} point clouds with multi-task pointwise
  networks and multi-value conditional random fields}.
\newblock In \emph{Proc.~of the IEEE Conference on Computer Vision and Pattern
  Recognition (CVPR)}, pages 8827--8836, 2019{\natexlab{b}}.

\bibitem[Qi et~al.(2017{\natexlab{a}})Qi, Su, Mo, and Guibas]{qi2017pointnet}
Charles~R Qi, Hao Su, Kaichun Mo, and Leonidas~J Guibas.
\newblock
  \href{http://openaccess.thecvf.com/content_cvpr_2017/papers/Qi_PointNet_Deep_Learning_CVPR_2017_paper.pdf}
  {{PointNet}: Deep learning on point sets for {3D} classification and
  segmentation}.
\newblock In \emph{Proc.~of the IEEE Conference on Computer Vision and Pattern
  Recognition (CVPR)}, pages 652--660, 2017{\natexlab{a}}.

\bibitem[Qi et~al.(2019)Qi, Litany, He, and Guibas]{qi2019deep}
Charles~R Qi, Or~Litany, Kaiming He, and Leonidas~J Guibas.
\newblock
  \href{https://openaccess.thecvf.com/content_ICCV_2019/papers/Qi_Deep_Hough_Voting_for_3D_Object_Detection_in_Point_Clouds_ICCV_2019_paper.pdf}{Deep
  {Hough} voting for {3D} object detection in point clouds}.
\newblock In \emph{Proc.~of the IEEE Int.~Conference on Computer Vision
  (ICCV)}, pages 9277--9286, 2019.

\bibitem[Qi et~al.(2017{\natexlab{b}})Qi, Yi, Su, and Guibas]{qi2017pointnet++}
Charles~Ruizhongtai Qi, Li~Yi, Hao Su, and Leonidas~J Guibas.
\newblock
  \href{https://papers.nips.cc/paper/7095-pointnet-deep-hierarchical-feature-learning-on-point-sets-in-a-metric-space.pdf}
  {{PointNet++}: Deep hierarchical feature learning on point sets in a metric
  space}.
\newblock In \emph{Proc.~of the Advances in Neural Information Processing
  Systems (NIPS)}, pages 5099--5108, 2017{\natexlab{b}}.

\bibitem[Ravanbakhsh et~al.(2016)Ravanbakhsh, Schneider, and
  P{\'o}czos]{Ravanbakhsh2016DeepLW}
Siamak Ravanbakhsh, Jeff~G. Schneider, and Barnab{\'a}s P{\'o}czos.
\newblock \href{https://arxiv.org/abs/1611.04500}{Deep Learning with Sets and
  Point Clouds}.
\newblock \emph{arXiv preprint arXiv:1611.04500}, 2016.

\bibitem[Rethage et~al.(2018)Rethage, Wald, Sturm, Navab, and
  Tombari]{rethage2018fully}
Dario Rethage, Johanna Wald, Jurgen Sturm, Nassir Navab, and Federico Tombari.
\newblock \href{https://arxiv.org/abs/1808.06840}{Fully-convolutional point
  networks for large-scale point clouds}.
\newblock In \emph{Proc.~of the European Conference on Computer Vision (ECCV)},
  pages 596--611, 2018.

\bibitem[Ronneberger et~al.(2015)Ronneberger, Fischer, and
  Brox]{ronneberger2015u}
Olaf Ronneberger, Philipp Fischer, and Thomas Brox.
\newblock \href{https://arxiv.org/abs/1505.04597}{U-Net: Convolutional networks
  for biomedical image segmentation}.
\newblock In \emph{{International Conference on Medical Image Computing and
  Computer-assisted Intervention}}, pages 234--241, 2015.

\bibitem[Rosu et~al.(2020)Rosu, Sch{\"u}tt, Quenzel, and
  Behnke]{rosu2019latticenet}
Radu~Alexandru Rosu, Peer Sch{\"u}tt, Jan Quenzel, and Sven Behnke.
\newblock \href{https://arxiv.org/abs/1912.05905}{{LatticeNet}: Fast point
  cloud segmentation using permutohedral lattices}.
\newblock \emph{Proc.~of Robotics: Science and Systems}, 2020.

\bibitem[Shi et~al.(2020)Shi, Lin, Wang, Hung, and Wang]{shi2020spsequencenet}
Hanyu Shi, Guosheng Lin, Hao Wang, Tzu-Yi Hung, and Zhenhua Wang.
\newblock
  \href{https://openaccess.thecvf.com/content_CVPR_2020/papers/Shi_SpSequenceNet_Semantic_Segmentation_Network_on_4D_Point_Clouds_CVPR_2020_paper.pdf}
  {SpSequenceNet: Semantic Segmentation Network on {4D} Point Clouds}.
\newblock In \emph{Proc.~of the IEEE Conference on Computer Vision and Pattern
  Recognition (CVPR)}, pages 4574--4583, 2020.

\bibitem[Stotko et~al.(2019)Stotko, Krumpen, Weinmann, and
  Klein]{stotko2019efficient}
Patrick Stotko, Stefan Krumpen, Michael Weinmann, and Reinhard Klein.
\newblock \href{https://arxiv.org/abs/1908.03118}{Efficient {3D} Reconstruction
  and Streaming for Group-Scale Multi-Client Live Telepresence}.
\newblock In \emph{Proc.~of the IEEE Int.~Symposium on Mixed and Augmented
  Reality (ISMAR)}, pages 19--25, 2019.

\bibitem[Su et~al.(2018)Su, Jampani, Sun, Maji, Kalogerakis, Yang, and
  Kautz]{su2018splatnet}
Hang Su, Varun Jampani, Deqing Sun, Subhransu Maji, Evangelos Kalogerakis,
  Ming-Hsuan Yang, and Jan Kautz.
\newblock
  \href{http://openaccess.thecvf.com/content_cvpr_2018/papers/Su_SPLATNet_Sparse_Lattice_CVPR_2018_paper.pdf}
  {{SplatNet}: Sparse lattice networks for point cloud processing}.
\newblock In \emph{Proc.~of the IEEE Conference on Computer Vision and Pattern
  Recognition (CVPR)}, pages 2530--2539, 2018.

\bibitem[Tanke et~al.(2019)Tanke, Kwon, Stotko, Rosu, Weinmann, Errami, Behnke,
  Bennewitz, Klein, Weber, et~al.]{tanke2019bonn}
Julian Tanke, Oh-Hun Kwon, Patrick Stotko, Radu~Alexandru Rosu, Michael
  Weinmann, Hassan Errami, Sven Behnke, Maren Bennewitz, Reinhard Klein,
  Andreas Weber, et~al.
\newblock \href{https://arxiv.org/abs/1912.06354}{Bonn Activity Maps: Dataset
  Description}.
\newblock \emph{arXiv preprint arXiv:1912.06354}, 2019.

\bibitem[Tatarchenko et~al.(2018)Tatarchenko, Park, Koltun, and
  Zhou]{tatarchenko2018tangent}
Maxim Tatarchenko, Jaesik Park, Vladlen Koltun, and Qian-Yi Zhou.
\newblock
  \href{https://lmb.informatik.uni-freiburg.de/Publications/2018/Tat18/}{Tangent
  convolutions for dense prediction in {3D}}.
\newblock In \emph{Proc.~of the IEEE Conference on Computer Vision and Pattern
  Recognition (CVPR)}, pages 3887--3896, 2018.

\bibitem[Tchapmi et~al.(2017)Tchapmi, Choy, Armeni, Gwak, and
  Savarese]{tchapmi2017segcloud}
Lyne Tchapmi, Christopher Choy, Iro Armeni, JunYoung Gwak, and Silvio Savarese.
\newblock \href{https://arxiv.org/abs/1710.07563} {{SEGCloud}: Semantic
  segmentation of {3D} point clouds}.
\newblock In \emph{Intl.~Conf.~on 3D Vision (3DV)}, pages 537--547. IEEE, 2017.

\bibitem[Thomas et~al.(2019)Thomas, Qi, Deschaud, Marcotegui, Goulette, and
  Guibas]{thomas2019kpconv}
Hugues Thomas, Charles~R Qi, Jean-Emmanuel Deschaud, Beatriz Marcotegui,
  Fran{\c{c}}ois Goulette, and Leonidas~J Guibas.
\newblock \href{https://arxiv.org/abs/1904.08889} { {KPConv}: Flexible and
  deformable convolution for point clouds}.
\newblock In \emph{Proc.~of the IEEE Int.~Conference on Computer Vision
  (ICCV)}, pages 6411--6420, 2019.

\bibitem[Wang et~al.(2018{\natexlab{a}})Wang, Suo, Ma, Pokrovsky, and
  Urtasun]{wang2018deep}
Shenlong Wang, Simon Suo, Wei-Chiu Ma, Andrei Pokrovsky, and Raquel Urtasun.
\newblock
  \href{http://openaccess.thecvf.com/content_cvpr_2018/papers/Wang_Deep_Parametric_Continuous_CVPR_2018_paper.pdf}
  {Deep parametric continuous convolutional neural networks}.
\newblock In \emph{Proc.~of the IEEE Conference on Computer Vision and Pattern
  Recognition (CVPR)}, pages 2589--2597, 2018{\natexlab{a}}.

\bibitem[Wang et~al.(2018{\natexlab{b}})Wang, Yu, Huang, and
  Neumann]{wang2018sgpn}
Weiyue Wang, Ronald Yu, Qiangui Huang, and Ulrich Neumann.
\newblock \href{https://arxiv.org/abs/1711.08588}{{SGPN}: Similarity group
  proposal network for {3D} point cloud instance segmentation}.
\newblock In \emph{Proc.~of the IEEE Conference on Computer Vision and Pattern
  Recognition (CVPR)}, pages 2569--2578, 2018{\natexlab{b}}.

\bibitem[Wang et~al.(2019)Wang, Liu, Shen, Shen, and
  Jia]{wang2019associatively}
Xinlong Wang, Shu Liu, Xiaoyong Shen, Chunhua Shen, and Jiaya Jia.
\newblock
  \href{https://openaccess.thecvf.com/content_CVPR_2019/papers/Wang_Associatively_Segmenting_Instances_and_Semantics_in_Point_Clouds_CVPR_2019_paper.pdf}{Associatively
  segmenting instances and semantics in point clouds}.
\newblock In \emph{Proc.~of the IEEE Conference on Computer Vision and Pattern
  Recognition (CVPR)}, pages 4096--4105, 2019.

\bibitem[Wu et~al.(2018)Wu, Zhou, Zhao, Yue, and Keutzer]{wu2018squeezesegv2}
Bichen Wu, Xuanyu Zhou, Sicheng Zhao, Xiangyu Yue, and Kurt Keutzer.
\newblock \href{https://arxiv.org/abs/1809.08495}{SqueezeSegv2: Improved model
  structure and unsupervised domain adaptation for road-object segmentation
  from a lidar point cloud}.
\newblock \emph{arXiv preprint arXiv:1809.08495}, 2018.

\bibitem[Wu et~al.(2019)Wu, Qi, and Fuxin]{wu2019pointconv}
Wenxuan Wu, Zhongang Qi, and Li~Fuxin.
\newblock
  \href{http://openaccess.thecvf.com/content_CVPR_2019/papers/Wu_PointConv_Deep_Convolutional_Networks_on_3D_Point_Clouds_CVPR_2019_paper.pdf}
  {{PointConv}: Deep convolutional networks on {3D} point clouds}.
\newblock In \emph{Proc.~of the IEEE Conference on Computer Vision and Pattern
  Recognition (CVPR)}, pages 9621--9630, 2019.

\bibitem[Wu and He(2018)]{wu2018group}
Yuxin Wu and Kaiming He.
\newblock
  \href{https://eccv2018.org/openaccess/content_ECCV_2018/papers/Yuxin_Wu_Group_Normalization_ECCV_2018_paper.pdf}
  {Group normalization}.
\newblock In \emph{Proc.~of the European Conference on Computer Vision (ECCV)},
  pages 3--19, 2018.

\bibitem[Yang et~al.(2019)Yang, Wang, Clark, Hu, Wang, Markham, and
  Trigoni]{yang2019learning}
Bo~Yang, Jianan Wang, Ronald Clark, Qingyong Hu, Sen Wang, Andrew Markham, and
  Niki Trigoni.
\newblock \href{https://arxiv.org/abs/1906.01140} {Learning object bounding
  boxes for {3D} instance segmentation on point clouds}.
\newblock \emph{arXiv preprint arXiv:1906.01140}, 2019.

\bibitem[Yi et~al.(2016)Yi, Kim, Ceylan, Shen, Yan, Su, Lu, Huang, Sheffer,
  Guibas, et~al.]{yi2016scalable}
Li~Yi, Vladimir~G Kim, Duygu Ceylan, I~Shen, Mengyan Yan, Hao Su, Cewu Lu,
  Qixing Huang, Alla Sheffer, Leonidas Guibas, et~al.
\newblock
  \href{https://cs.stanford.edu/~ericyi/papers/part_annotation_16_small.pdf}{A
  scalable active framework for region annotation in {3D} shape collections}.
\newblock \emph{ACM Transactions on Graphics (ToG)}, 35\penalty0 (6):\penalty0
  210, 2016.

\bibitem[Yi et~al.(2019)Yi, Zhao, Wang, Sung, and Guibas]{yi2019gspn}
Li~Yi, Wang Zhao, He~Wang, Minhyuk Sung, and Leonidas~J Guibas.
\newblock \href{https://arxiv.org/abs/1812.03320}{{GSPN}: Generative shape
  proposal network for {3D} instance segmentation in point cloud}.
\newblock In \emph{Proc.~of the IEEE Conference on Computer Vision and Pattern
  Recognition (CVPR)}, pages 3947--3956, 2019.

\bibitem[Zaheer et~al.(2017)Zaheer, Kottur, Ravanbakhsh, Poczos, Salakhutdinov,
  and Smola]{zaheer2017deep}
Manzil Zaheer, Satwik Kottur, Siamak Ravanbakhsh, Barnabas Poczos, Russ~R
  Salakhutdinov, and Alexander~J Smola.
\newblock \href{https://papers.nips.cc/paper/6931-deep-sets.pdf}{Deep sets}.
\newblock In \emph{Proc.~of the Advances in Neural Information Processing
  Systems (NIPS)}, pages 3391--3401, 2017.

\end{thebibliography}
\end{document}